\documentclass[10pt]{article} %

\newcommand{\myparagraph}[1]{\vspace{0.25em}\noindent\textbf{#1}\hspace{0.5em}}

\usepackage{xspace}
\usepackage{tabularx}
\usepackage{siunitx} %
\usepackage{xfrac}
\newrobustcmd\B{\DeclareFontSeriesDefault[rm]{bf}{b}\bfseries}

\usepackage{makecell}
\usepackage{longtable}
\usepackage{wrapfig}

\usepackage{tikz}
\usepackage[outline]{contour}
    \contourlength{0.1em}
\usepackage{pgfplots}
\usepgfplotslibrary{groupplots}
\usepgfplotslibrary{colorbrewer}
\pgfplotsset{compat=1.18} 

\usepackage{tocloft}
\usetikzlibrary{positioning} %

\usepackage{tcolorbox}
\newtcolorbox{mybox}{colback=transformercolor!15!white,colframe=transformercolor!75!white, boxsep=-1pt}%
\newcommand{\answerbox}[1]{%
\begin{mybox}
\textbf{Conclusion:} #1
\end{mybox}
}

\newcommand \colorindicator[1]{%
	{\textcolor{#1}{$\blacksquare\!\!\!\!\!\blacksquare$}}%
}

\newcommand*\circled[1]{%
    \tikz[baseline={([yshift=-2.75pt] C.base)}] \node[draw=#1, circle, fill=#1, inner sep=2pt] (C) {};%
}

\newcommand*\squared[1]{%
    \tikz[baseline={([yshift=-2.5pt] C.base)}] \node[draw=#1, rectangle, fill=#1, inner sep=2.5pt] (C) {};%
}

\newcommand*\triangled[1]{%
    \tikz[baseline=(C.base)] 
        \filldraw[fill=#1, draw=#1] 
        (0,0) -- ++(3pt,5.5pt) -- ++(3pt,-5.5pt) -- cycle;%
}

\newcommand*\diamonded[1]{%
    \tikz[baseline=(C.base)] 
        \filldraw[fill=#1, draw=#1] 
        (0pt,0.3pt) -- ++(-2.2pt,3pt) -- ++(2.2pt,+3pt)  -- ++(2.2pt,-3pt) -- cycle;%
}

\newcommand*\pentagoned[1]{%
    \tikz[baseline=(C.base)] 
        \filldraw[fill=#1, draw=#1] 
        (0pt,0.3pt) -- ++(-1.5pt,3pt) -- ++(3pt,+2.275pt)  -- ++(3pt,-2.275pt)  -- ++(-1.5pt,-3pt) -- cycle;%
}

\newcommand*\tablearrowcenter[0]{}

\definecolor{mymagentadecent}{rgb}{0.678, 0.435, 0.678}
\definecolor{cnn_im1k}{rgb}{0.78, 0.949, 0.949}
\definecolor{cnn_im21k}{rgb}{0.384, 0.78, 0.78}
\definecolor{cnn_bd}{rgb}{0, 0.58, 0.58}
\definecolor{transformer_im1k}{rgb}{0.941, 0.808, 0.941}
\definecolor{transformer_im21k}{rgb}{0.902, 0.514, 0.902}
\definecolor{transformer_bd}{rgb}{0.82, 0, 0.82}
\definecolor{bcos_im1k}{rgb}{0.69, 0.733, 0.851}
\definecolor{bcos_im21k}{rgb}{0.455, 0.529, 0.722}
\definecolor{bcos_bd}{rgb}{0.227, 0.267, 0.365}
\definecolor{vlm_im1k}{rgb}{0.988, 0.925, 0.612}
\definecolor{vlm_im21k}{rgb}{0.98, 0.882, 0.38}
\definecolor{vlm_bd}{rgb}{0.831, 0.694, 0.016}

\definecolor{cnncolor}{HTML}{56B4E9}
\definecolor{transformercolor}{HTML}{E69F00}
\definecolor{bcoscolor}{HTML}{009E73}
\definecolor{vilcolor}{HTML}{F0E442}

\pgfplotsset{colormap={correlation_cm}{HTML=(8F6200) color=(transformercolor) color=(white) color=(cnncolor) HTML=(10567E)}}

\newcommand{\tablesetupcnnvstrans}[0]{(h)\xspace}
\newcommand{\tablesetupcnndataset}[0]{(a)\xspace}
\newcommand{\tablesetuptransdataset}[0]{(b)\xspace}
\newcommand{\tablesetupat}[0]{(c)\xspace}
\newcommand{\tablesetupselfsllp}[0]{(d)\xspace}
\newcommand{\tablesetupselfsletoe}[0]{(e)\xspace}
\newcommand{\tablesetupsemisl}[0]{(f)\xspace}
\newcommand{\tablesetuplong}[0]{(g)\xspace}

\newcommand{\tablesetupbcos}[0]{(i)\xspace}
\newcommand{\tablesetupzeroshot}[0]{(j)\xspace}

\newcommand{\nrmodels}[0]{326\xspace}

\newcommand{\updated}[1]{\textcolor{black}{#1}}
\usepackage[normalem]{ulem} %

\usepackage{subcaption}
\usepackage{amssymb}
\usepackage{booktabs}

\def\eg{\emph{e.g}\onedot} 
\def\ie{\emph{i.e}\onedot} 
\def\cf{\emph{cf}\onedot} 
\def\vsiccv{\emph{vs}\onedot}
\makeatletter
\DeclareRobustCommand\onedot{\futurelet\@let@token\@onedot}
\def\@onedot{\ifx\@let@token.\else.\null\fi\xspace}

\usepackage[accepted]{tmlr}

\usepackage{amsmath,amsfonts,bm}

\def\eqref#1{equation~\ref{#1}}

\def\1{\bm{1}}

\DeclareMathAlphabet{\mathsfit}{\encodingdefault}{\sfdefault}{m}{sl}
\SetMathAlphabet{\mathsfit}{bold}{\encodingdefault}{\sfdefault}{bx}{n}

\newcommand{\R}{\mathbb{R}}

\usepackage{hyperref}
\usepackage{url}

\usepackage[capitalize]{cleveref}
\crefname{section}{Sec.}{Secs.}
\Crefname{section}{Section}{Sections}
\Crefname{table}{Table}{Tables}
\crefname{table}{Tab.}{Tabs.}
\Crefname{figure}{Figure}{Figures}
\crefname{figure}{Fig.}{Figs.}

\title{Beyond Accuracy: What Matters in Designing \\ Well-Behaved \updated{Image Classification} Models?}

\author{\name Robin Hesse$^{*\text{\textdagger}}$ \email rhesse@mpi-inf.mpg.de \\
      \addr Max Planck Institute for Informatics, SIC
      \AND
      \name Do\u{g}ukan Ba\u{g}c\i{}$^{*}$ \email dogukan.bagci@stud.tu-darmstadt.de \\
      \addr Department of Computer Science\\
      Technical University of Darmstadt\\
      Zuse School ELIZA
      \AND
      \name Bernt Schiele \email schiele@mpi-inf.mpg.de\\
      \addr Max Planck Institute for Informatics, SIC\\
      Zuse School ELIZA
      \AND
      \name Simone Schaub-Meyer \email simone.schaub@visinf.tu-darmstadt.de \\
      \addr Department of Computer Science\\
      Technical University of Darmstadt\\
      hessian.AI
      \AND
      \name Stefan Roth \email stefan.roth@visinf.tu-darmstadt.de \\
      \addr Department of Computer Science\\
      Technical University of Darmstadt\\
      hessian.AI\\
      Zuse School ELIZA
      }

\def\month{01}  %
\def\year{2026} %
\def\openreview{\url{https://openreview.net/forum?id=E7HDtLCoT6}} %

\begin{document}

\maketitle

\begin{abstract}
  	
Deep learning has become an essential part of computer vision, with deep neural networks (DNNs) excelling in predictive performance. However, they often fall short in other critical quality dimensions, such as robustness, calibration, or fairness. While existing studies have focused on a subset of these quality dimensions, none have explored a more general form of ``well-behavedness'' of DNNs.
With this work, we address this gap by simultaneously studying nine different quality dimensions for image classification. Through a large-scale study, we provide a bird's-eye view by analyzing \nrmodels backbone models and how different training paradigms and model architectures affect these quality dimensions. We reveal various new insights such that \emph{(i)} vision-language models exhibit high class balance on ImageNet-1k classification and strong robustness against domain changes; 
\emph{(ii)} training models initialized with weights obtained through self-supervised learning is an effective strategy to improve most considered quality dimensions;
and \emph{(iii)} the training dataset size is a major driver for most of the quality dimensions. We conclude our study by introducing the QUBA score (\textbf{Q}uality \textbf{U}nderstanding \textbf{B}eyond \textbf{A}ccuracy), a novel metric that ranks models across multiple dimensions of quality, enabling tailored recommendations based on specific user needs.\footnote{Project page: \url{https://visinf.github.io/beyond-accuracy}; \textsuperscript{\normalfont{}*}equal contribution; \textsuperscript{\normalfont{}\textdagger}work done while at Technical University of Darmstadt}%

\end{abstract}

\section{Introduction}
\label{sec_introduction}

Today's computer vision research is heavily shaped by advances in the field of deep learning. While deep neural networks (DNNs) excel at predictive performance, often measured via the accuracy, it has been shown that they are flawed across various other quality dimensions, such as robustness~\citep{Hendrycks:2019:BNN, Goodfellow:2015:EHA}, calibration~\citep{Guo:2017:OCM}, and fairness~\citep{Du:FDL:2021}. To address these challenges, the scientific community started various parallel streams of research focusing on \textit{individual} aspects of DNN quality, developing mostly orthogonally. We argue that this orthogonal development is somewhat surprising, as the overarching goal for most applications should be the implementation of \emph{more well-behaved} networks that excel in \emph{many} %
quality dimensions \citep{Liu:2023:TTA}. While some works study the relationship between a subset of quality dimensions, \eg, how accuracy and calibration relate~\citep{Minderer:2021:RCM,Guo:2017:OCM}, or how adversarial training~\citep{Goodfellow:2015:EHA} improves calibration~\citep{Grabinski:2022:RBA}, we are not aware of any work that studies a broad range of quality dimensions \textit{simultaneously}. Consequently, it is largely unknown how model improvements in one direction affect other quality dimensions. Here, we close this gap by studying how \nrmodels backbone models perform along nine different quality dimensions for ImageNet-1k~\citep{Russakovsky:2015:ILS} image classification. %
By doing so, we analyze how different training paradigms and model architectures can be used to improve these quality dimensions, uncover unknown connections between quality dimensions, and give recommendations on what models to use based on specific user needs. %
We expect our findings to be highly relevant for advancing the development of classification models that not only excel in accuracy but also across a wide range of DNN quality dimensions. %

\begin{figure}%
\centering
\centering
        \begin{subfigure}{0.45\linewidth}
            \centering
            \newcommand{\D}{9} %
\newcommand{\U}{4} %

\newdimen\R %
\R=1.2cm 
\newdimen\L %
\L=1.5cm
\newdimen\maxLabelDist %
\maxLabelDist=1.15cm %

\newcommand{\myA}{360/\D} %

\begin{tikzpicture}[scale=1,every node/.style={font=\sffamily}]
\sffamily
  \path (0:0cm) coordinate (O); %

  \foreach \X in {1,...,\D}{
    \draw [opacity=0.2](\X*\myA + 50:0) -- (\X*\myA + 50:\R); %
  } 

  \foreach \Y in {0,...,\U}{
    \foreach \X in {1,...,\D}{
      \path (\X*\myA + 50:\Y*\R/\U) coordinate (D\X-\Y); %
    }
    \draw [opacity=0.2] (90:\Y*\R/\U) \foreach \X in {1,...,\D}{
        -- (\X*\myA + 50:\Y*\R/\U) %
    } -- cycle;
  }

   \path (1*\myA + 50:\L) node (L1) {\tiny Accuracy};
  \path (2*\myA + 50:\L) node (L2) {\tiny {\makecell{Adv. \\ Rob.}}};
  \path (3*\myA + 50:\L) node (L3) {\tiny C-Rob.};
  \path (4*\myA + 50:\L) node (L4) {\tiny {\makecell{OOD \\ Rob.}}};
  \path (5*\myA + 50:\L) node (L5) {\tiny Cal. Error};
  \path (6*\myA + 50:\L) node (L6) {\tiny {\makecell{Class \\ Balance}}};
  \path (7*\myA + 50:\L) node (L7) {\tiny {\makecell{Obj. \\ Focus}}};
  \path (8*\myA + 50:\L) node (L8) {\tiny {\makecell{Shape \\ Bias}}};
  \path (9*\myA + 50:\L) node (L9) {\tiny Params.};

  \path (1*\myA + 50:0.5\R) node[anchor=center, xshift=0pt, yshift=0pt] {\tiny 0};
  \path (1*\myA + 50:0.75\R) node[anchor=center, xshift=0pt, yshift=0pt] {\tiny 1};
  \path (1*\myA + 50:1.0\R) node[anchor=center, xshift=0pt, yshift=0pt] {\tiny 2};
  \path (1*\myA + 50:0.25\R) node[anchor=center, xshift=0pt, yshift=0pt] {\tiny -1};
  \path (1*\myA + 50:0.0\R) node[anchor=center, xshift=0pt, yshift=0pt] {\tiny -2};

  \draw [color=transformercolor,line width=0.5pt,opacity=0.5, fill=transformercolor](1*\myA + 50:0.8333333333333327\R) -- (2*\myA + 50:0.6136363636363636\R) -- (3*\myA + 50:0.717391304347826\R) -- (4*\myA + 50:0.7666666666666667\R) -- (5*\myA + 50:0.5\R) -- (6*\myA + 50:0.8125000000000002\R) -- (7*\myA + 50:0.75\R) -- (8*\myA + 50:0.7325581395348837\R) -- (9*\myA + 50:0.6891891891891893\R) -- cycle;

  \draw [color=cnncolor,line width=0.5pt,opacity=0.7, fill=cnncolor](1*\myA + 50:0.5833333333333334\R) -- (2*\myA + 50:0.4318181818181818\R) -- (3*\myA + 50:0.6521739130434783\R) -- (4*\myA + 50:0.5166666666666667\R) -- (5*\myA + 50:0.39814814814814814\R) -- (6*\myA + 50:0.6041666666666667\R) -- (7*\myA + 50:0.5\R) -- (8*\myA + 50:0.71875\R) -- (9*\myA + 50:0.7267441860465116\R) -- cycle;
  
  \path (-90:\L+0.5cm) node (L99) {\scriptsize \textbf{SL~\colorindicator{cnncolor} \vsiccv self-SL \& fine-tuning~\colorindicator{transformercolor}}\strut};

\end{tikzpicture}
        \end{subfigure}
        \begin{subfigure}{0.45\linewidth}
           \centering
           \newcommand{\D}{9} %
\newcommand{\U}{4} %

\newdimen\R %
\R=1.2cm 
\newdimen\L %
\L=1.5cm
\newdimen\maxLabelDist %
\maxLabelDist=1.15cm %

\newcommand{\myA}{360/\D} %

\begin{tikzpicture}[scale=1,every node/.style={font=\sffamily}]
\sffamily
  \path (0:0cm) coordinate (O); %

  \foreach \X in {1,...,\D}{
    \draw [opacity=0.2](\X*\myA + 50:0) -- (\X*\myA + 50:\R); %
  } 

  \foreach \Y in {0,...,\U}{
    \foreach \X in {1,...,\D}{
      \path (\X*\myA + 50:\Y*\R/\U) coordinate (D\X-\Y); %
    }
    \draw [opacity=0.2] (90:\Y*\R/\U) \foreach \X in {1,...,\D}{
        -- (\X*\myA + 50:\Y*\R/\U) %
    } -- cycle;
  }

   \path (1*\myA + 50:\L) node (L1) {\tiny Accuracy};
  \path (2*\myA + 50:\L) node (L2) {\tiny {\makecell{Adv. \\ Rob.}}};
  \path (3*\myA + 50:\L) node (L3) {\tiny C-Rob.};
  \path (4*\myA + 50:\L) node (L4) {\tiny {\makecell{OOD \\ Rob.}}};
  \path (5*\myA + 50:\L) node (L5) {\tiny Cal. Error};
  \path (6*\myA + 50:\L) node (L6) {\tiny {\makecell{Class \\ Balance}}};
  \path (7*\myA + 50:\L) node (L7) {\tiny {\makecell{Obj. \\ Focus}}};
  \path (8*\myA + 50:\L) node (L8) {\tiny {\makecell{Shape \\ Bias}}};
  \path (9*\myA + 50:\L) node (L9) {\tiny Params.};

  \path (1*\myA + 50:0.5\R) node[anchor=center, xshift=0pt, yshift=0pt] {\tiny 0};
  \path (1*\myA + 50:0.75\R) node[anchor=center, xshift=0pt, yshift=0pt] {\tiny 6};
  \path (1*\myA + 50:1.0\R) node[anchor=center, xshift=0pt, yshift=0pt] {\tiny 12};
  \path (1*\myA + 50:0.25\R) node[anchor=center, xshift=0pt, yshift=0pt] {\tiny -6};
  \path (1*\myA + 50:0.0\R) node[anchor=center, xshift=0pt, yshift=0pt] {\tiny -12};

  \draw [color=vilcolor,line width=0.5pt,opacity=0.6, fill=vilcolor](1*\myA + 50:0.4166666666666666\R) -- (2*\myA + 50:0.46590909090909094\R) -- (3*\myA + 50:0.5126811594202898\R) -- (4*\myA + 50: 0.6194444444444445\R) -- (5*\myA + 50:0.9506172839506173\R) -- (6*\myA + 50:0.7083333333333334\R) -- (7*\myA + 50: 0.5\R) -- (8*\myA + 50:0.6302083333333334\R) -- (9*\myA + 50:0.7131782945736435\R) -- cycle;

  \draw [color=cnncolor,line width=0.5pt,opacity=0.7, fill=cnncolor](1*\myA + 50:0.513888888888889\R) -- (2*\myA + 50: 0.4962121212121212\R) -- (3*\myA + 50:0.5163043478260869\R) -- (4*\myA + 50:0.49722222222222223\R) -- (5*\myA + 50:0.4984567901234568\R) -- (6*\myA + 50:0.5173611111111112\R) -- (7*\myA + 50:0.5\R) -- (8*\myA + 50: 0.5208333333333334\R) -- (9*\myA + 50:0.5939922480620154\R) -- cycle;
  
  \path (-90:\L+0.5cm) node (L99) {\scriptsize \textbf{Standard~\colorindicator{cnncolor} \vsiccv ViL~\colorindicator{vilcolor} models}\strut};

\end{tikzpicture}
        \end{subfigure}

\caption{\textit{Visualization of two of our main results.} We compare nine different quality dimensions for popular backbone models trained with standard supervised learning~\colorindicator{cnncolor} (SL) against the corresponding backbones trained %
after initialization with weights obtained through self-supervised learning~\colorindicator{transformercolor} \emph{(left)} and when utilized in a vision-language (ViL) model~\colorindicator{vilcolor} \emph{(right)}.
Axis units indicate the distance (in standard deviations) to the mean (0 line) of each quality dimension; see \cref{eq:quba} and its explanation for details. Please refer to \cref{tab:comparisons} \tablesetupselfsletoe and \tablesetupzeroshot for raw values and \cref{sec:experiments_whatmakesbetter} for an interpretation of the results.}
    \label{fig:teaser}
\end{figure}

Our contributions can be summarized as:
\textit{(1)} We introduce a novel benchmark to measure a broad range of quality dimensions simultaneously, which is compatible with \textit{any} DNN/backbone that performs ImageNet-1k~\citep{Russakovsky:2015:ILS} classification.
\textit{(2)} In a large-scale study, we evaluate how \nrmodels backbone models from prior work perform along nine considered quality dimensions.
\textit{(3)} We use this to analyze how different training paradigms and architectural changes can be utilized to improve the different quality dimensions. Among other things, we find that self-supervised pre-training followed by fine-tuning improves most quality dimensions and that vision-language models achieve high fairness \updated{(here measured as the class balance)} on ImageNet-1k classification while being fairly robust against domain changes (see \cref{fig:teaser}).
\textit{(4)} Building on trends in related work that examine relationships between individual quality dimensions~\citep[\eg,][]{Guo:2017:OCM, Minderer:2021:RCM, Miller:2021:AOT}, we analyze the relationships among \textit{all} considered quality dimensions.
\textit{(5)} We conclude our study by introducing a novel \textit{QUBA score} (\textbf{Q}uality \textbf{U}nderstanding \textbf{B}eyond \textbf{A}ccuracy) that ranks models across multiple dimensions of quality. We use this score to recommend top-performing models tailored to diverse user needs. %

\section{Evaluating quality beyond accuracy}
\label{sec:quality_dims}

We go beyond accuracy by exploring the general ``well-behavedness'' of DNNs. While this term is inherently ill-defined and task-dependent, we use it \textit{informally} as the performance across the nine quality dimensions chosen in this study. %
Specifically, we consider \textit{(1)}~accuracy; three robustness metrics: \textit{(2)}~adversarial robustness, \textit{(3)}~corruption robustness, and \textit{(4)}~out-of-domain robustness; \textit{(5)}~calibration error; \textit{(6)}~\updated{fairness measured via class balance}; two dimensions concerned with shortcut-learning: \textit{(7)}~object focus and \textit{(8)}~shape bias; and \textit{(9)}~computational cost measured via the number of parameters. These dimensions encompass a wide range of DNN qualities/properties and attracted considerable attention in related work, which is why we regard them as particularly important. 
We evaluate these quality dimensions simultaneously by merging corresponding evaluation protocols into a single, comprehensive benchmark. To enable a large-scale study with a feasible computational load, we only select protocols that require ImageNet-1k~\citep{Russakovsky:2015:ILS} classification without model fine-tuning.
While there are other important dimensions, such as explainability and out-of-distribution detection capabilities, they are challenging to evaluate as they are coupled to specific explanation/out-of-distribution detection methods that might vary across backbones~\citep{Hesse:2023:SVD} -- thus, we excluded them. \updated{Also, we note that all evaluation protocols are merely proxies for the targeted dimension; they do not necessarily reflect true performance in that dimension, nor do they capture the full complexity of the underlying behavior.} Consequently, different evaluation protocols may lead to different conclusions, which is worth keeping in mind when interpreting our findings.

We now outline related work, the considered quality dimensions, and how we measure them in this work; see \cref{tab:protocols} for a summary and \cref{appendix:sec:quality_dimensions} for more details.

\begin{table*}
  \centering
  \scriptsize
      \caption{\textit{Overview of our considered DNN quality dimensions for image classification.} Arrows indicate if higher ($\uparrow$) or lower ($\downarrow$) is better. If a quality dimension is computed over multiple datasets or metrics, we use the geometric mean.}

  \begin{tabularx}{\textwidth}{@{}lX@{}}
    \toprule
    \textbf{Quality Dimension} & \textbf{Description} (visual illustrations are provided in \cref{appendix:sec:quality_dimensions})\\
    \midrule
    Accuracy $(\uparrow)$ & Fraction of correctly classified (clean) images\\
    Adversarial Robustness $(\uparrow)$ & Fraction of correctly classified images after an FGSM or PGD attack (normalized by clean accuracy)\\
    C-Robustness $(\uparrow)$ & Fraction of correctly classified images after corrupting images (normalized by clean accuracy)\\
    OOD Robustness $(\uparrow)$ & Fraction of correctly classified images from different domains (normalized by clean accuracy)\\
    Calibration Error $(\downarrow)$& Misalignment of the output confidence and the true probability of a correct classification\\
    Class Balance $(\uparrow)$ & Standard deviation of the accuracies and average confidences across all individual classes\\
    Object Focus $(\uparrow)$& Fraction of decisions that are based on foreground and not on background\\
    Shape Bias $(\uparrow)$& Fraction of decisions that are based on shape and not on texture\\
    Parameters $(\downarrow)$ & Number of parameters\\
    \bottomrule
  \end{tabularx}
  \label{tab:protocols}
\end{table*}

\myparagraph{Accuracy.} The success of DNNs is largely driven by their superior accuracy, first showcased in 2012 by AlexNet~\citep{Krizhevsky:2012:INC} on ImageNet-1k~\citep{Russakovsky:2015:ILS}.
This marked the beginning of the \textit{``deep learning era,''} leading to increasingly powerful models~\citep{Simonyan:2015:VDC,He:2016:DRL,Dosovitskiy:2021:IWW,Liu:2022:ACF}.
To measure a model's accuracy, we report the top-1 accuracy on the ImageNet-1k evaluation split.

\myparagraph{Adversarial robustness.} 
DNNs are vulnerable to adversarial attacks, \ie, small perturbations in the input space~\citep{Szegedy:2014:IPN}. This vulnerability can be reduced by training with adversarial examples~\citep{Goodfellow:2015:EHA,Madry:2018:TDL} 
or by defensive distillation~\citep{Papernot:2016:DDA}.
To assess adversarial robustness, we measure the geometric mean\footnote{The geometric mean ensures that metrics of different scales contribute equally without one overshadowing the others.} of the accuracies after applying two popular attacks, FGSM~\citep{Goodfellow:2015:EHA} and PGD~\citep{Madry:2018:TDL}. %
To reduce the dependence on the clean accuracy of the model, we report adversarial robustness \textit{relative} to the clean ImageNet-1k accuracy. In \cref{appendix:sec:autoattack}, we compare our adversarial robustness measure to \emph{AutoAttack}~\citep{Croce:2021:RBS}, yielding very similar conclusions.%

\myparagraph{Corruption robustness.} DNNs are susceptible to common image corruptions such as JPEG compression and contrast changes~\citep{Hendrycks:2019:BNN}, which can be reduced with special kinds of data augmentations~\citep{Hendrycks:2020:SDP} or self-supervised learning~\citep{Hendrycks:2019:USS}.
To assess a model's robustness to common corruptions (C-robustness), we measure the mean accuracy on \mbox{ImageNet-C}~\citep{Hendrycks:2019:BNN}, \ie, the ImageNet evaluation split with different corruption types of increasing strength. To normalize C-robustness and to be consistent with our other robustness metrics, we again report the top-1 accuracy on the corrupted data \textit{relative} to the clean ImageNet-1k accuracy.

\myparagraph{OOD robustness.}
Out-of-domain (OOD) robustness is concerned with the generalizability of a model to OOD data \citep[\eg,][]{Hendrycks:2021:MFR,Wang:2019:LRG,Geirhos:2019:ITC}. Contrary to adversarial and corruption robustness, datasets to assess the OOD robustness exhibit stronger visual domain shifts and contain new data samples.
We assess OOD robustness by reporting the geometric mean of the relative accuracy on five established OOD datasets: ImageNet-R~\citep{Hendrycks:2021:MFR}, ImageNet-Sketch~\citep{Wang:2019:LRG}, as well as Stylized-ImageNet, Edge, and Silhouette from \citet{Geirhos:2019:ITC}. %

\myparagraph{Calibration error.} 
Calibration measures how well a model's output confidence reflects the probability of a correct prediction. \citet{Guo:2017:OCM} found DNNs to be poorly calibrated, spurring the developments of approaches like deep ensembles~\citep{Lakshminarayanan:2017:SSP} and label smoothing~\citep{Mueller:2019:WDL}. 
We report the geometric mean of the expected calibration error (ECE)~\citep{Nixon:2019:MCD, Guo:2017:OCM} and the adaptive calibration error (ACE)~\citep{Nixon:2019:MCD}.

\myparagraph{\updated{Class balance.}} 
\updated{A well-behaved model should behave fairly. Fairness can, \eg, be improved by the class-wise weighting of the cross-entropy loss~\citep{Benz:2020:RMB}.
As fairness has multiple facets~\citep{Verma:2018:FDE} and there is no standardized fairness metric, we rely on a simplified notion of ``class balance'': no class should be particularly favored or disadvantaged~\citep{Benz:2020:RMB, Kuzucu:2024:UFM}. 
More specifically, we evaluate the class balance of a model in two ways: 
\textit{(1)} the standard deviation of ImageNet-1k class accuracies, similar to \citet{Croce:2021:RBS}, and \textit{(2)} the standard deviation of average class confidences, similar to \citet{Kuzucu:2024:UFM}. 
To align with other metrics, we subtract these values from 1, so higher scores indicate greater class balance, and we aggregate both measures using the geometric mean.} While aiming to approximate fairness, we note that ImageNet-1k itself exhibits dataset biases, and thus our conclusions about fairness may not fully reflect real-world concerns.

\myparagraph{Object focus.} 
ImageNet-trained DNNs rely on background features to the extent that they can be fooled by changing the background~\citep{Xiao:2021:NSR, Zhu:2017:ORO}. This can be avoided by training on images where the background signals are decorrelated from the class labels~\citep{Xiao:2021:NSR}.
Similar to \citet{Xiao:2021:NSR}, we measure object focus by assessing accuracy drops when replacing backgrounds with those from other classes.

\myparagraph{Shape bias.} 
ImageNet-trained CNNs exhibit a texture bias rather than a shape bias~\citep{Geirhos:2019:ITC}. Since this can hurt generalizability and robustness, models with increased shape bias are preferred and have been developed~\citep{Nam:2021:RDG,Shi:2020:IDR}.
The shape bias is measured using images with a conflict between the shape and texture cues -- \eg, an image of a cat (shape) with the skin of an elephant (texture). These images are fed into the model to determine whether it predicts based on texture (texture bias) or shape (shape bias)~\citep{Geirhos:2019:ITC}. %

\begin{wrapfigure}{r}{0.3\textwidth}
  \centering
    \vspace{-2.5em}
    \begin{tikzpicture}[every node/.style={font=\sffamily}]
\sffamily

\scriptsize

\def\yticklabeloffset{-0.55};
\def\xticklabeloffset{-0.6};
\def\xticklabeloffsetx{-0.1};
\def\xticklabelrotation{60};

\begin{axis}[%
    axis equal image, %
    width=3.5cm,
    axis line style = {line width=.5pt,draw=gray!50},
    scatter, %
    colormap name=correlation_cm, %
    colorbar, %
    point meta min=-1,
    point meta max=1,
    clip=false,
    grid=minor, %
    minor grid style={line width=.5pt,draw=gray!50},
    minor tick num=1, %
    minor tick length=0pt, 
    tickwidth=0pt, %
    ticks=none, %
    try min ticks=4, %
    y dir=reverse, %
    colorbar style={
        at={(1.05,0.5)},anchor=west,width=0.25cm,
        ytick={-1,0,1},
        yticklabels={-1, 0, 1},%
    },
    xticklabel pos=right, %
    enlargelimits={abs=0.5}, %
    scatter/@pre marker code/.append code={%
      \pgfplotstransformcoordinatex{sqrt(abs(\pgfplotspointmeta))}%
      \scope[mark size=\pgfplotsunitxlength*\pgfmathresult/3.7 + \pgfplotsunitxlength*20.7/2, fill=mapped color]
    },
    scatter/@post marker code/.append code={%
      \endscope%
    }
    ]

\addplot +[
    point meta=explicit, %
    only marks, %
    every node near coord/.append style={font=\small, color=white,anchor=center},
    visualization depends on={value \thisrow{label} \as \Label}, %
    visualization depends on={value \thisrow{size} \as \Size}, %
    ] table [
    x expr={int(mod(\coordindex+0.01,3))}, %
    y expr={int((\coordindex+0.01)/3))},
    meta=value,
] {
X   Y   value label size
0 0 1.0 $+$ 0.1em
0 0 0 $$ 0.1em
0 0 0 $$ 0.1em
 
0 0 0.8567 $+$ 0.1em
0 0 1.0 $+$ 0.1em
0 0 0 $$ 0.1em
 
0 0 0.9574 $+$ 0.1em
0 0 0.8948 $+$ 0.1em
0 0 1.0 $+$ 0.1em

};

\node[anchor=east] at (axis cs:0+\yticklabeloffset,0) {Parameters\strut};
\node[anchor=east] at (axis cs:0+\yticklabeloffset,1) {FLOPs \strut};
\node[anchor=east] at (axis cs:0+\yticklabeloffset,2) {Memory\strut};

\node[anchor=east] at (axis cs:0+\yticklabeloffset,11.3) {\strut};

\node[anchor=west, rotate=\xticklabelrotation] at (axis cs:0+\xticklabeloffsetx,0+\xticklabeloffset) {Parameters\strut};
\node[anchor=west, rotate=\xticklabelrotation] at (axis cs:1+\xticklabeloffsetx,0+\xticklabeloffset) {FLOPs\strut};
\node[anchor=west, rotate=\xticklabelrotation] at (axis cs:2+\xticklabeloffsetx,0+\xticklabeloffset) {Memory\strut};

\end{axis}

\end{tikzpicture}
    \vspace{-12.5em}
    \caption{\updated{\textit{Rank correlation matrix for the considered metrics on computational cost for our full model zoo of \nrmodels models.} All entries have a $p$-value below 0.05, indicating statistical significance.}}
    \vspace{-1em}
  \label{fig:correlation_matrix_cost}
\end{wrapfigure}
\updated{Unlike the other quality dimensions, a higher shape bias is not inherently better, as some applications may benefit from a stronger focus on texture. Nonetheless, we include shape bias as a dimension and, for consistency with related work and since it is more in line with human vision~\citep{Geirhos:2019:ITC,Wang:2020:HFC}, assume that higher values are preferable. This assumption does not affect the core of our analysis. 
Further, in \cref{sec:experiments_rankingdetails}, we introduce the QUBA score, where the shape bias weight can be adjusted or even inverted to reflect specific preferences.}

\myparagraph{Parameters.}
\updated{Deep neural networks (DNNs) should be memory-efficient and fast to reduce resource consumption and operate sustainably. Since actual memory and runtime performance depend heavily on specific implementations and hardware, we follow established practice and use the number of model parameters as a hardware- and implementation-independent proxy for computational cost \citep{Tay:SEI:2021, Kaplan:2020:SLF}. To validate its suitability as a proxy, we visualize the Spearman’s rank correlation for \nrmodels models (see \cref{sec:model_zoo}) between the number of parameters, required memory, and theoretical FLOPs in \cref{fig:correlation_matrix_cost}. All three measures are strongly correlated (correlation coefficients $>0.85$), confirming that the number of parameters is a reliable proxy. Thus, for practical purposes and ease of comparison in future studies (memory consumption is implementation dependent, and the computation for the theoretical number of FLOPs needs to be adjusted for novel model components), we adopt the number of parameters as our primary metric. However, we note that parameter count alone cannot fully capture the efficiency of a model and should be interpreted with caution~\citep{Dehghani:2022:TEM}.}
Various studies are concerned with reducing the number of parameters or the computational cost of DNNs while keeping the accuracy as high as possible~\citep{Hesse:2023:CAD,Li:2022:EVT,Tan:2019:ERM,Tan:2021:ESM}. %

\section{What makes a model more well-behaved?}
\label{sec:experiments_whatmakesbetter}

Equipped with the above quality metrics, we now study how different design choices affect DNN quality. %
This not only sheds light on \emph{quality beyond accuracy} of the current state of the art for image classification, but also facilitates the development of more well-behaved DNNs in the future.
In \cref{tab:comparisons}, we compare the average of each quality dimension for different setups. 
Since not all backbones can be considered in each configuration (\eg, not all have adversarially trained variants), we ensure fair comparisons by selecting the subset of models that is consistently available for all configurations of the respective setup (with no or minor modifications; see \cref{appendix:sec:details_comparison}).
In \cref{fig:scatter_dimensions_acc}, 
we plot the different quality dimensions for selected models against accuracy, distinguishing between model groups.

\myparagraph{Experimental setup.} 
We use \nrmodels publicly available models in our large-scale study. See \cref{sec:model_zoo} for an overview, including numerical results and implementation details for each quality dimension.

\begin{figure*}[t]
    \centering
        \hspace{-1.5em}
        \begin{subfigure}{0.275\linewidth}
            \centering
            \begin{tikzpicture}[every node/.style={font=\sffamily}]
\sffamily
\scriptsize

\begin{axis}[
            title={Adv. Rob.\strut},
            title style={yshift=-1em},
            width=1\linewidth,
            minor y tick num=0,
            xtick={0.6,0.7, 0.8, 0.9},
            xticklabels={0.6,0.7, 0.8, 0.9},
            ytick={0,0.2, 0.4, 0.6},
            yticklabels={0,0.2, 0.4, 0.6},
            label style={font=\scriptsize},
            tick label style={font=\scriptsize},
            x label style={at={(axis description cs:0.5,+0.05)},anchor=north},
            y label style={at={(axis description cs:+0.08,.5)},anchor=center},
            ]
    \addplot [
        scatter,
        only marks,
        point meta=explicit symbolic,
        scatter/classes={
            a={mark=square*,fill=cnncolor,draw=cnncolor,mark size=2pt, opacity=0.5, draw opacity=0.7},
            b={mark=*,fill=cnncolor,draw=cnncolor, mark size=2.2pt, opacity=0.5, draw opacity=0.7},
            c={mark=triangle*,fill=cnncolor,draw=cnncolor, mark size=3pt, opacity=0.5, draw opacity=0.7},
            d={mark=pentagon*,fill=cnncolor,draw=cnncolor, mark size=2.5pt, opacity=0.5, draw opacity=0.7},
            e={mark=diamond*,fill=cnncolor,draw=cnncolor, mark size=3pt, opacity=0.5, draw opacity=0.7},
            f={mark=square*,fill=transformercolor,draw=transformercolor,mark size=2pt, opacity=0.5, draw opacity=0.7},
            g={mark=*,fill=transformercolor,draw=transformercolor, mark size=2.2pt, opacity=0.5, draw opacity=0.7},
            h={mark=triangle*,fill=transformercolor,draw=transformercolor, mark size=3pt, opacity=0.5, draw opacity=0.7},
            i={mark=pentagon*,fill=transformercolor,draw=transformercolor, mark size=2.5pt, opacity=0.5, draw opacity=0.7},
            j={mark=diamond*,fill=transformercolor,draw=transformercolor, mark size=3pt, opacity=0.5, draw opacity=0.7},
            k={mark=square*,fill=bcoscolor,draw=bcoscolor,mark size=2pt, opacity=0.5, draw opacity=0.7},
            l={mark=triangle*,fill=vilcolor,draw=vilcolor, mark size=3pt, opacity=0.5, draw opacity=0.7}
        },
        visualization depends on={value \thisrow{symbol} \as \symbol},
        visualization depends on={value \thisrow{centeroffset} \as \centeroffset},
        nodes near coords*={\symbol},
        every node near coord/.append style={anchor=center, yshift=-\centeroffset, opacity=0.5},
    ] table [meta=label] {
x	y	label	symbol	centeroffset
0.56516	0.03520598132175203	a	{$ $}	0.0pt
0.69784	0.08336962037138924	a	{$ $}	0.0pt
0.74216	0.02313139060183982	a	{$ $}	0.0pt
0.7832	0.02573364378214033	a	{$ $}	0.0pt
0.78844	0.04918779490692883	a	{$ $}	0.0pt
0.58182	0.0144598221593309	a	{$ $}	0.0pt
0.80132	0.155207586500866	a	{$ $}	0.0pt
0.80446	0.151272772915873	a	{$ $}	0.0pt
0.83246	0.0411825259627465	a	{$ $}	0.0pt
0.77114	0.0555962242053594	a	{$ $}	0.0pt
0.7904	0.05413577443827669	a	{$ $}	0.0pt
0.7187	0.0220085708395447	a	{$ $}	0.0pt
0.76226	0.0108437488530857	a	{$ $}	0.0pt
0.82628	0.1890455467287331	a	{$ $}	0.0pt
0.74046	0.0272302833552138	a	{$ $}	0.0pt
0.779	0.002587214888990194	a	{$\cdot$}	0.3pt
0.64526	0.00349900321206046	a	{$ $}	0.0pt
0.76514	0.0238379895773606	a	{$ $}	0.0pt
0.84124	0.072245988237705	a	{$ $}	0.0pt
0.86828	0.109800125669087	e	{\fontsize{4pt}{4pt}$\star$}	0.0pt
0.68404	0.3763219697519731	b	{$ $}	0.0pt
0.63862	0.3625998080497007	b	{$ $}	0.0pt
0.84508	0.01214271747202183	a	{$\cdot$}	0.3pt
0.80874	0.0246602015177562	a	{$ $}	0.0pt
0.81066	0.1759295120004755	f	{$ $}	0.0pt
0.8358	0.04744995216257621	f	{$ $}	0.0pt
0.83162	0.0885281033178427	a	{$ $}	0.0pt
0.83332	0.06018549519867554	a	{$\cdot$}	0.3pt
0.84556	0.1532253970095634	a	{$ $}	0.0pt
0.84772	0.1529627624929557	a	{$\cdot$}	0.3pt
0.852	0.17252672388622	a	{$ $}	0.0pt
0.85822	0.1688902164715952	a	{$\cdot$}	0.3pt
0.8198	0.13424151251287	f	{$ $}	0.0pt
0.8229	0.14920035122047	f	{$ $}	0.0pt
0.85044	0.164956148951956	f	{$ $}	0.0pt
0.82502	0.1538050613999951	f	{$ $}	0.0pt
0.82896	0.1989064082365947	f	{$ $}	0.0pt
0.82588	0.1689597032114894	f	{$ $}	0.0pt
0.82446	0.1337498193750414	f	{$ $}	0.0pt
0.836	0.1237194462337648	f	{$ $}	0.0pt
0.8493	0.256140769130309	f	{$ $}	0.0pt
0.81358	0.151587035168662	f	{$ $}	0.0pt
0.83068	0.275058774443752	f	{$\cdot$}	0.3pt
0.83094	0.265402219091004	f	{$ $}	0.0pt
0.84558	0.242816104049227	f	{$\cdot$}	0.3pt
0.83786	0.26646502529849	f	{$ $}	0.0pt
0.85708	0.25991072524207	f	{$\cdot$}	0.3pt
0.84778	0.318426937419631	f	{$ $}	0.0pt
0.86976	0.348769965618503	f	{$\cdot$}	0.3pt
0.85252	0.275412455280522	f	{$ $}	0.0pt
0.84598	0.04456484533200876	f	{$ $}	0.0pt
0.86942	0.264706665020486	f	{$\cdot$}	0.3pt
0.79172	0.1479157116610288	f	{$ $}	0.0pt
0.80882	0.08541640477462238	f	{$\cdot$}	0.3pt
0.81502	0.208793323198474	f	{$ $}	0.0pt
0.8324	0.113806408985984	f	{$\cdot$}	0.3pt
0.83248	0.236072697020133	f	{$ $}	0.0pt
0.85084	0.128596380206269	f	{$\cdot$}	0.3pt
0.78836	0.07400043358587835	f	{$ $}	0.0pt
0.85838	0.02805364847953247	f	{$\cdot$}	0.3pt
0.82516	0.09331344628825539	a	{$ $}	0.0pt
0.829	0.158091097944856	a	{$\cdot$}	0.3pt
0.83634	0.1115025623490814	a	{$ $}	0.0pt
0.8459	0.20104843267703	a	{$\cdot$}	0.3pt
0.84062	0.1159079148005859	a	{$ $}	0.0pt
0.85812	0.204962083109558	a	{$\cdot$}	0.3pt
0.84414	0.1222078966335481	a	{$ $}	0.0pt
0.86606	0.223626288548999	a	{$\cdot$}	0.3pt
0.8522	0.074401727686739	h	{$\cdot$}	0.0pt
0.83374	0.1402055533506747	f	{$ $}	0.0pt
0.8465	0.1869906530125084	f	{$ $}	0.0pt
0.78744	0.5880353818649724	g	{$ $}	0.0pt
0.78118	0.5678054219936125	b	{$ $}	0.0pt
0.76074	0.567349268099113	b	{$ $}	0.0pt
0.7743	0.5946513954994364	b	{$ $}	0.0pt
0.76718	0.5452933630390857	g	{$ $}	0.0pt
0.81864	0.1842018825073163	c	{$ $}	0.0pt
0.8203	0.1182954474445436	c	{$\cdot$}	0.0pt
0.82938	0.2006447959931005	c	{$ $}	0.0pt
0.83892	0.1633251951269493	c	{$\cdot$}	0.0pt
0.84876	0.280803050674841	c	{$ $}	0.0pt
0.86744	0.2802004231189423	c	{$\cdot$}	0.0pt
0.8576	0.3508179481476644	c	{$ $}	0.0pt
0.87264	0.3182331935592866	c	{$\cdot$}	0.0pt
0.86068	0.187783598799804	h	{$ $}	0.0pt
0.88264	0.0985588767091609	h	{$\cdot$}	0.0pt
0.84088	0.210440125744022	f	{$ $}	0.0pt
0.835	0.1374881077795623	f	{$ $}	0.0pt
0.85602	0.3313009934879599	h	{$ $}	0.0pt
0.81226	0.3582591143181425	a	{$ $}	0.0pt
0.79078	0.2803478311815448	a	{$ $}	0.0pt
0.80314	0.1982447007644184	e	{\fontsize{4pt}{4pt}$\star$}	0.0pt
0.81138	0.1568247214601345	e	{\fontsize{4pt}{4pt}$\star$}	0.0pt
0.84278	0.3335326701427104	e	{\fontsize{4pt}{4pt}$\star$}	0.0pt
0.73286	0.1140210600073006	e	{\fontsize{4pt}{4pt}$\star$}	0.0pt
0.83742	0.2529049039242492	h	{$ $}	0.0pt
0.66754	0.03610289889888413	h	{$ $}	0.0pt
0.65178	0.05722717503935306	h	{$ $}	0.0pt
0.78064	0.1344297996748871	h	{$ $}	0.0pt
0.82858	0.2888684816696597	h	{$ $}	0.0pt
0.75292	0.1746620389767571	c	{$ $}	0.0pt
0.78456	0.451594856895194	c	{$ $}	0.0pt
0.86398	0.1941234630952666	h	{\fontsize{4pt}{4pt}$\star$}	0.0pt
0.75466	0.007883	f	{$\cdot$}	0.3pt
0.7607	0.09851414368351413	l	{\fontsize{4pt}{4pt}$\star$}	0.0pt
0.68364	0.2872113240572833	l	{\fontsize{4pt}{4pt}$\star$}	0.0pt
0.82038	0.2893567324722511	l	{\fontsize{4pt}{4pt}$\star$}	0.0pt
0.63376	0.2631005458482035	l	{\fontsize{4pt}{4pt}$\star$}	0.0pt
0.76764	0.00253821693905284	l	{\fontsize{4pt}{4pt}$\star$}	0.0pt
0.81632	0.2057587623084416	d	{$ $}	0.0pt
0.82414	0.5633575693370966	d	{$ $}	0.0pt
0.81958	0.605272	d	{$ $}	0.0pt
0.80532	0.588034	d	{$ $}	0.0pt
0.81516	0.6936239508867137	d	{$ $}	0.0pt
0.82388	0.180625	d	{$ $}	0.0pt
0.8212	0.5495333436537474	d	{$ $}	0.0pt
0.81772	0.4914365213775378	d	{$ $}	0.0pt
0.80308	0.563239	d	{$ $}	0.0pt
0.81678	0.4568789310403994	d	{$ $}	0.0pt
0.8229	0.1695925720940079	d	{$ $}	0.0pt
0.8229	0.4080955492472749	d	{$ $}	0.0pt
0.80546	0.3923412739191968	d	{$ $}	0.0pt
0.78838	0.5016481013470249	d	{$ $}	0.0pt
0.8159	0.4996793373143888	d	{$ $}	0.0pt
0.79482	0.02786626693840694	k	{$ $}	0.0pt
0.77104	0.02119710333948711	k	{$ $}	0.0pt
0.76626	0.007643205402844954	k	{$ $}	0.0pt
0.76492	0.08808505793949326	k	{$ $}	0.0pt
0.74408	0.008844768967272263	k	{$ $}	0.0pt
0.79228	0.1687286485479796	a	{$ $}	0.0pt
0.7725	0.01431221936917593	l	{\fontsize{4pt}{4pt}$\star$}	0.0pt
0.76902	0.09119571062283105	h	{$ $}	0.0pt
0.80462	0.2292961183915035	l	{\fontsize{4pt}{4pt}$\star$}	0.0pt
0.76526	0.001746189511636016	k	{$ $}	0.0pt
0.7213	0.03137655161598994	l	{\fontsize{4pt}{4pt}$\star$}	0.0pt
0.75898	0.08037031642046331	l	{\fontsize{4pt}{4pt}$\star$}	0.0pt
0.79168	0.08920776629163492	l	{\fontsize{4pt}{4pt}$\star$}	0.0pt
0.72786	0.4875557784196847	g	{$ $}	0.0pt
    };
\end{axis}
\end{tikzpicture}
        \end{subfigure}
        \hspace{-2.5em}
        \begin{subfigure}{0.275\linewidth}
           \centering
           \begin{tikzpicture}[every node/.style={font=\sffamily}]
\sffamily
\scriptsize

\begin{axis}[
            title={C-Rob.\strut},
            title style={yshift=-1em},
            width=1\linewidth,
            minor y tick num=0,
            xtick={0.6,0.7, 0.8, 0.9},
            xticklabels={0.6,0.7, 0.8, 0.9},
            ytick={0, 0.4, 0.8},
            yticklabels={0, 0.4, 0.8},
            label style={font=\scriptsize},
            tick label style={font=\scriptsize},
            x label style={at={(axis description cs:0.5,+0.05)},anchor=north},
            y label style={at={(axis description cs:+0.08,.5)},anchor=center},
            ]
    \addplot [
        scatter,
        only marks,
        point meta=explicit symbolic,
        scatter/classes={
            a={mark=square*,fill=cnncolor,draw=cnncolor,mark size=2pt, opacity=0.5, draw opacity=0.7},
            b={mark=*,fill=cnncolor,draw=cnncolor, mark size=2.2pt, opacity=0.5, draw opacity=0.7},
            c={mark=triangle*,fill=cnncolor,draw=cnncolor, mark size=3pt, opacity=0.5, draw opacity=0.7},
            d={mark=pentagon*,fill=cnncolor,draw=cnncolor, mark size=2.5pt, opacity=0.5, draw opacity=0.7},
            e={mark=diamond*,fill=cnncolor,draw=cnncolor, mark size=3pt, opacity=0.5, draw opacity=0.7},
            f={mark=square*,fill=transformercolor,draw=transformercolor,mark size=2pt, opacity=0.5, draw opacity=0.7},
            g={mark=*,fill=transformercolor,draw=transformercolor, mark size=2.2pt, opacity=0.5, draw opacity=0.7},
            h={mark=triangle*,fill=transformercolor,draw=transformercolor, mark size=3pt, opacity=0.5, draw opacity=0.7},
            i={mark=pentagon*,fill=transformercolor,draw=transformercolor, mark size=2.5pt, opacity=0.5, draw opacity=0.7},
            j={mark=diamond*,fill=transformercolor,draw=transformercolor, mark size=3pt, opacity=0.5, draw opacity=0.7},
            k={mark=square*,fill=bcoscolor,draw=bcoscolor,mark size=2pt, opacity=0.5, draw opacity=0.7},
            l={mark=triangle*,fill=vilcolor,draw=vilcolor, mark size=3pt, opacity=0.5, draw opacity=0.7}
        },
        visualization depends on={value \thisrow{symbol} \as \symbol},
        visualization depends on={value \thisrow{centeroffset} \as \centeroffset},
        nodes near coords*={\symbol},
        every node near coord/.append style={anchor=center, yshift=-\centeroffset, opacity=0.5},
    ] table [meta=label] {
x	y	label	symbol	centeroffset
0.56516	0.344160407781601	a	{$ $}	0.0pt
0.69784	0.5012197494506836	a	{$ $}	0.0pt
0.74216	0.377986341714859	a	{$ $}	0.0pt
0.7832	0.4719710648059845	a	{$ $}	0.0pt
0.78844	0.5385048389434814	a	{$ $}	0.0pt
0.58182	0.3152553	a	{$ $}	0.0pt
0.80132	0.5576058	a	{$ $}	0.0pt
0.80446	0.6316438317298889	a	{$ $}	0.0pt
0.83246	0.5413721	a	{$ $}	0.0pt
0.77114	0.5363963	a	{$ $}	0.0pt
0.7904	0.5736517310142517	a	{$ $}	0.0pt
0.7187	0.43511006	a	{$ $}	0.0pt
0.76226	0.3578168451786041	a	{$ $}	0.0pt
0.82628	0.6756982803344727	a	{$ $}	0.0pt
0.74046	0.48685446	a	{$ $}	0.0pt
0.779	0.5557306408882141	a	{$\cdot$}	0.3pt
0.64526	0.15471885	a	{$ $}	0.0pt
0.76514	0.40012038	a	{$ $}	0.0pt
0.84124	0.7705923318862915	a	{$ $}	0.0pt
0.86828	0.5695245	e	{\fontsize{4pt}{4pt}$\star$}	0.0pt
0.68404	0.5158775448799133	b	{$ $}	0.0pt
0.63862	0.5016200542449951	b	{$ $}	0.0pt
0.84508	0.5526469349861145	a	{$\cdot$}	0.3pt
0.80874	0.4797781109809875	a	{$ $}	0.0pt
0.81066	0.6636431813240051	f	{$ $}	0.0pt
0.8358	0.664406418800354	f	{$ $}	0.0pt
0.83162	0.6621919870376587	a	{$ $}	0.0pt
0.83332	0.6780634522438049	a	{$\cdot$}	0.3pt
0.84556	0.7071669101715088	a	{$ $}	0.0pt
0.84772	0.7111818194389343	a	{$\cdot$}	0.3pt
0.852	0.7391367554664612	a	{$ $}	0.0pt
0.85822	0.7268839478492737	a	{$\cdot$}	0.3pt
0.8198	0.60640067	f	{$ $}	0.0pt
0.8229	0.62019706	f	{$ $}	0.0pt
0.85044	0.70287114	f	{$ $}	0.0pt
0.82502	0.6393834	f	{$ $}	0.0pt
0.82896	0.73923934	f	{$ $}	0.0pt
0.82588	0.63565344	f	{$ $}	0.0pt
0.82446	0.6045531630516052	f	{$ $}	0.0pt
0.836	0.597086	f	{$ $}	0.0pt
0.8493	0.7448872	f	{$ $}	0.0pt
0.81358	0.73801124	f	{$ $}	0.0pt
0.83068	0.66778564	f	{$\cdot$}	0.3pt
0.83094	0.7592914	f	{$ $}	0.0pt
0.84558	0.689123	f	{$\cdot$}	0.3pt
0.83786	0.7870896	f	{$ $}	0.0pt
0.85708	0.7272292	f	{$\cdot$}	0.3pt
0.84778	0.82069916	f	{$ $}	0.0pt
0.86976	0.79811984	f	{$\cdot$}	0.3pt
0.85252	0.71979344	f	{$ $}	0.0pt
0.84598	0.6580666303634644	f	{$ $}	0.0pt
0.86942	0.8129622	f	{$\cdot$}	0.3pt
0.79172	0.640385	f	{$ $}	0.0pt
0.80882	0.6733151078224182	f	{$\cdot$}	0.3pt
0.81502	0.69402796	f	{$ $}	0.0pt
0.8324	0.7039885	f	{$\cdot$}	0.3pt
0.83248	0.7309962	f	{$ $}	0.0pt
0.85084	0.7534791	f	{$\cdot$}	0.3pt
0.78836	0.6732136011123657	f	{$ $}	0.0pt
0.85838	0.7127097249031067	f	{$\cdot$}	0.3pt
0.82516	0.6456781029701233	a	{$ $}	0.0pt
0.829	0.7100374102592468	a	{$\cdot$}	0.3pt
0.83634	0.6821578741073608	a	{$ $}	0.0pt
0.8459	0.7580978274345398	a	{$\cdot$}	0.3pt
0.84062	0.7006479501724243	a	{$ $}	0.0pt
0.85812	0.7815569043159485	a	{$\cdot$}	0.3pt
0.84414	0.7193056344985962	a	{$ $}	0.0pt
0.86606	0.8065952658653259	a	{$\cdot$}	0.3pt
0.8522	0.7924337387084961	h	{$\cdot$}	0.0pt
0.83374	0.6320197582244873	f	{$ $}	0.0pt
0.8465	0.7188307642936707	f	{$ $}	0.0pt
0.78744	0.6510159373283386	g	{$ $}	0.0pt
0.78118	0.6356788873672485	b	{$ $}	0.0pt
0.76074	0.6374897360801697	b	{$ $}	0.0pt
0.7743	0.6552168130874634	b	{$ $}	0.0pt
0.76718	0.6877485513687134	g	{$ $}	0.0pt
0.81864	0.6483466	c	{$ $}	0.0pt
0.8203	0.68491954	c	{$\cdot$}	0.0pt
0.82938	0.6942982	c	{$ $}	0.0pt
0.83892	0.721432626247406	c	{$\cdot$}	0.0pt
0.84876	0.766647	c	{$ $}	0.0pt
0.86744	0.7923387289047241	c	{$\cdot$}	0.0pt
0.8576	0.788891	c	{$ $}	0.0pt
0.87264	0.7979338765144348	c	{$\cdot$}	0.0pt
0.86068	0.6962693333625793	h	{$ $}	0.0pt
0.88264	0.6468083	h	{$\cdot$}	0.0pt
0.84088	0.6778085231781006	f	{$ $}	0.0pt
0.835	0.6256858	f	{$ $}	0.0pt
0.85602	0.793698251247406	h	{$ $}	0.0pt
0.81226	0.08967660367488861	a	{$ $}	0.0pt
0.79078	0.04598647728562355	a	{$ $}	0.0pt
0.80314	0.04679727926850319	e	{\fontsize{4pt}{4pt}$\star$}	0.0pt
0.81138	0.07402011752128601	e	{\fontsize{4pt}{4pt}$\star$}	0.0pt
0.84278	0.05995470285415649	e	{\fontsize{4pt}{4pt}$\star$}	0.0pt
0.73286	0.06386150419712067	e	{\fontsize{4pt}{4pt}$\star$}	0.0pt
0.83742	0.7053480744361877	h	{$ $}	0.0pt
0.66754	0.4341807961463928	h	{$ $}	0.0pt
0.65178	0.5844069719314575	h	{$ $}	0.0pt
0.78064	0.6561568975448608	h	{$ $}	0.0pt
0.82858	0.7550220489501953	h	{$ $}	0.0pt
0.75292	0.0256922859698534	c	{$ $}	0.0pt
0.78456	0.066361435	c	{$ $}	0.0pt
0.86398	0.8617190718650818	h	{\fontsize{4pt}{4pt}$\star$}	0.0pt
0.75466	0.563729	f	{$\cdot$}	0.3pt
0.7607	0.62633926	l	{\fontsize{4pt}{4pt}$\star$}	0.0pt
0.68364	0.4458472430706024	l	{\fontsize{4pt}{4pt}$\star$}	0.0pt
0.82038	0.7692824602127075	l	{\fontsize{4pt}{4pt}$\star$}	0.0pt
0.63376	0.04110763594508171	l	{\fontsize{4pt}{4pt}$\star$}	0.0pt
0.76764	0.1504968106746674	l	{\fontsize{4pt}{4pt}$\star$}	0.0pt
0.81632	0.02400410547852516	d	{$ $}	0.0pt
0.82414	0.0411037877202034	d	{$ $}	0.0pt
0.81958	0.067712	d	{$ $}	0.0pt
0.80532	0.046333	d	{$ $}	0.0pt
0.81516	0.08021477609872818	d	{$ $}	0.0pt
0.82388	0.020531	d	{$ $}	0.0pt
0.8212	0.03784825652837753	d	{$ $}	0.0pt
0.81772	0.072100929915905	d	{$ $}	0.0pt
0.80308	0.05446	d	{$ $}	0.0pt
0.81678	0.07166226208209991	d	{$ $}	0.0pt
0.8229	0.01607382483780384	d	{$ $}	0.0pt
0.8229	0.03075164183974266	d	{$ $}	0.0pt
0.80546	0.06084002554416656	d	{$ $}	0.0pt
0.78838	0.03728524222970009	d	{$ $}	0.0pt
0.8159	0.05504607036709785	d	{$ $}	0.0pt
0.79482	0.5690929293632507	k	{$ $}	0.0pt
0.77104	0.5202130079269409	k	{$ $}	0.0pt
0.76626	0.1704872697591782	k	{$ $}	0.0pt
0.76492	0.1446615010499954	k	{$ $}	0.0pt
0.74408	0.6645836234092712	k	{$ $}	0.0pt
0.79228	0.02356681413948536	a	{$ $}	0.0pt
0.7725	0.78620684	l	{\fontsize{4pt}{4pt}$\star$}	0.0pt
0.76902	0.5977217555046082	h	{$ $}	0.0pt
0.80462	0.7206496	l	{\fontsize{4pt}{4pt}$\star$}	0.0pt
0.76526	0.1719221472740173	k	{$ $}	0.0pt
0.7213	0.6824942827224731	l	{\fontsize{4pt}{4pt}$\star$}	0.0pt
0.75898	0.7379972338676453	l	{\fontsize{4pt}{4pt}$\star$}	0.0pt
0.79168	0.8011206388473511	l	{\fontsize{4pt}{4pt}$\star$}	0.0pt
0.72786	0.6263083815574646	g	{$ $}	0.0pt
    };
\end{axis}
\end{tikzpicture}
        \end{subfigure}
        \hspace{-2.5em}
        \begin{subfigure}{0.275\linewidth}
           \centering
           \begin{tikzpicture}[every node/.style={font=\sffamily}]
\sffamily
\scriptsize

\begin{axis}[
            title={OOD Rob.\strut},
            title style={yshift=-1em},
            width=1\linewidth,
            minor y tick num=0,
            xtick={0.6,0.7, 0.8, 0.9},
            xticklabels={0.6,0.7, 0.8, 0.9},
            ytick={0.25,0.5, 0.75,1},
            yticklabels={0.25,0.5, 0.75,1},
            label style={font=\scriptsize},
            tick label style={font=\scriptsize},
            x label style={at={(axis description cs:0.5,+0.05)},anchor=north},
            y label style={at={(axis description cs:+0.08,.5)},anchor=center},
            ]
    \addplot [
        scatter,
        only marks,
        point meta=explicit symbolic,
        scatter/classes={
            a={mark=square*,fill=cnncolor,draw=cnncolor,mark size=2pt, opacity=0.5, draw opacity=0.7},
            b={mark=*,fill=cnncolor,draw=cnncolor, mark size=2.2pt, opacity=0.5, draw opacity=0.7},
            c={mark=triangle*,fill=cnncolor,draw=cnncolor, mark size=3pt, opacity=0.5, draw opacity=0.7},
            d={mark=pentagon*,fill=cnncolor,draw=cnncolor, mark size=2.5pt, opacity=0.5, draw opacity=0.7},
            e={mark=diamond*,fill=cnncolor,draw=cnncolor, mark size=3pt, opacity=0.5, draw opacity=0.7},
            f={mark=square*,fill=transformercolor,draw=transformercolor,mark size=2pt, opacity=0.5, draw opacity=0.7},
            g={mark=*,fill=transformercolor,draw=transformercolor, mark size=2.2pt, opacity=0.5, draw opacity=0.7},
            h={mark=triangle*,fill=transformercolor,draw=transformercolor, mark size=3pt, opacity=0.5, draw opacity=0.7},
            i={mark=pentagon*,fill=transformercolor,draw=transformercolor, mark size=2.5pt, opacity=0.5, draw opacity=0.7},
            j={mark=diamond*,fill=transformercolor,draw=transformercolor, mark size=3pt, opacity=0.5, draw opacity=0.7},
            k={mark=square*,fill=bcoscolor,draw=bcoscolor,mark size=2pt, opacity=0.5, draw opacity=0.7},
            l={mark=triangle*,fill=vilcolor,draw=vilcolor, mark size=3pt, opacity=0.5, draw opacity=0.7}
        },
        visualization depends on={value \thisrow{symbol} \as \symbol},
        visualization depends on={value \thisrow{centeroffset} \as \centeroffset},
        nodes near coords*={\symbol},
        every node near coord/.append style={anchor=center, yshift=-\centeroffset, opacity=0.5},
    ] table [meta=label] {
x	y	label	symbol	centeroffset
0.56516	0.5698545338145893	a	{$ $}	0.0pt
0.69784	0.5440479682661307	a	{$ $}	0.0pt
0.74216	0.461569793455964	a	{$ $}	0.0pt
0.7832	0.5589738080624422	a	{$ $}	0.0pt
0.78844	0.4669013474749842	a	{$ $}	0.0pt
0.58182	0.476549978816473	a	{$ $}	0.0pt
0.80132	0.5480603305654692	a	{$ $}	0.0pt
0.80446	0.556307996498741	a	{$ $}	0.0pt
0.83246	0.528030839977374	a	{$ $}	0.0pt
0.77114	0.513900734208324	a	{$ $}	0.0pt
0.7904	0.5218874517056283	a	{$ $}	0.0pt
0.7187	0.4967935076654995	a	{$ $}	0.0pt
0.76226	0.4764477090278845	a	{$ $}	0.0pt
0.82628	0.5470884119561313	a	{$ $}	0.0pt
0.74046	0.5553863102507673	a	{$ $}	0.0pt
0.779	0.540237657646377	a	{$\cdot$}	0.3pt
0.64526	0.247301165208127	a	{$ $}	0.0pt
0.76514	0.4792887926357755	a	{$ $}	0.0pt
0.84124	0.5960896664954716	a	{$ $}	0.0pt
0.86828	0.664506532334082	e	{\fontsize{4pt}{4pt}$\star$}	0.0pt
0.68404	0.4720466103484823	b	{$ $}	0.0pt
0.63862	0.4111456392163045	b	{$ $}	0.0pt
0.84508	0.6529914941266242	a	{$\cdot$}	0.3pt
0.80874	0.5412553264181874	a	{$ $}	0.0pt
0.81066	0.5561082515449453	f	{$ $}	0.0pt
0.8358	0.5864546422673911	f	{$ $}	0.0pt
0.83162	0.3165409979293317	a	{$ $}	0.0pt
0.83332	0.1309492408745278	a	{$\cdot$}	0.3pt
0.84556	0.528201663291512	a	{$ $}	0.0pt
0.84772	0.1606003395670111	a	{$\cdot$}	0.3pt
0.852	0.6402641412252392	a	{$ $}	0.0pt
0.85822	0.1288929307007369	a	{$\cdot$}	0.3pt
0.8198	0.549336632375421	f	{$ $}	0.0pt
0.8229	0.581097800030699	f	{$ $}	0.0pt
0.85044	0.571034119832928	f	{$ $}	0.0pt
0.82502	0.563753823362011	f	{$ $}	0.0pt
0.82896	0.670488379257203	f	{$ $}	0.0pt
0.82588	0.58978817790773	f	{$ $}	0.0pt
0.82446	0.590682708495121	f	{$ $}	0.0pt
0.836	0.5635734228250356	f	{$ $}	0.0pt
0.8493	0.6561470940158188	f	{$ $}	0.0pt
0.81358	0.588530586256191	f	{$ $}	0.0pt
0.83068	0.7019474013661994	f	{$\cdot$}	0.3pt
0.83094	0.6593021639189447	f	{$ $}	0.0pt
0.84558	0.7066759399738803	f	{$\cdot$}	0.3pt
0.83786	0.6954736705432643	f	{$ $}	0.0pt
0.85708	0.7603304277935553	f	{$\cdot$}	0.3pt
0.84778	0.7559898994530709	f	{$ $}	0.0pt
0.86976	0.8674976487519339	f	{$\cdot$}	0.3pt
0.85252	0.7180690574615327	f	{$ $}	0.0pt
0.84598	0.5331987914267325	f	{$ $}	0.0pt
0.86942	0.8117422082823446	f	{$\cdot$}	0.3pt
0.79172	0.5831426027523637	f	{$ $}	0.0pt
0.80882	0.655913288412389	f	{$\cdot$}	0.3pt
0.81502	0.6279271686299431	f	{$ $}	0.0pt
0.8324	0.7344521042718946	f	{$\cdot$}	0.3pt
0.83248	0.6615275673230121	f	{$ $}	0.0pt
0.85084	0.801169228509162	f	{$\cdot$}	0.3pt
0.78836	0.5550667564047156	f	{$ $}	0.0pt
0.85838	0.449764602372865	f	{$\cdot$}	0.3pt
0.82516	0.6002108754276722	a	{$ $}	0.0pt
0.829	0.6921675453127449	a	{$\cdot$}	0.3pt
0.83634	0.6508618198695629	a	{$ $}	0.0pt
0.8459	0.7567401786554702	a	{$\cdot$}	0.3pt
0.84062	0.6633555724304067	a	{$ $}	0.0pt
0.85812	0.8177887739085015	a	{$\cdot$}	0.3pt
0.84414	0.6661901313807128	a	{$ $}	0.0pt
0.86606	0.827390616451267	a	{$\cdot$}	0.3pt
0.8522	0.8009308247252808	h	{$\cdot$}	0.0pt
0.83374	0.5746206983887236	f	{$ $}	0.0pt
0.8465	0.6232435355703003	f	{$ $}	0.0pt
0.78744	0.7417890989412127	g	{$ $}	0.0pt
0.78118	0.6210023572524672	b	{$ $}	0.0pt
0.76074	0.8030474997974437	b	{$ $}	0.0pt
0.7743	0.4930107667551619	b	{$ $}	0.0pt
0.76718	0.5020858649440222	g	{$ $}	0.0pt
0.81864	0.649037228054508	c	{$ $}	0.0pt
0.8203	0.631704518741659	c	{$\cdot$}	0.0pt
0.82938	0.648352901165616	c	{$ $}	0.0pt
0.83892	0.7256398810893046	c	{$\cdot$}	0.0pt
0.84876	0.6973110723807968	c	{$ $}	0.0pt
0.86744	0.8184669029963325	c	{$\cdot$}	0.0pt
0.8576	0.7758417720745381	c	{$ $}	0.0pt
0.87264	0.8055260992486308	c	{$\cdot$}	0.0pt
0.86068	0.658425265758674	h	{$ $}	0.0pt
0.88264	0.578072802959305	h	{$\cdot$}	0.0pt
0.84088	0.639336721646642	f	{$ $}	0.0pt
0.835	0.56957831400866	f	{$ $}	0.0pt
0.85602	0.5452158793200601	h	{$ $}	0.0pt
0.81226	0.3470752029460658	a	{$ $}	0.0pt
0.79078	0.3378013817908559	a	{$ $}	0.0pt
0.80314	0.2391078669261527	e	{\fontsize{4pt}{4pt}$\star$}	0.0pt
0.81138	0.3261693472901511	e	{\fontsize{4pt}{4pt}$\star$}	0.0pt
0.84278	0.2979517618542906	e	{\fontsize{4pt}{4pt}$\star$}	0.0pt
0.73286	0.346694283903773	e	{\fontsize{4pt}{4pt}$\star$}	0.0pt
0.83742	0.5753629136581524	h	{$ $}	0.0pt
0.66754	0.4399568117935886	h	{$ $}	0.0pt
0.65178	0.6562305247126812	h	{$ $}	0.0pt
0.78064	0.4003424969211067	h	{$ $}	0.0pt
0.82858	0.6189507258426976	h	{$ $}	0.0pt
0.75292	0.1678140120470182	c	{$ $}	0.0pt
0.78456	0.319496132534405	c	{$ $}	0.0pt
0.86398	0.8667237993555099	h	{\fontsize{4pt}{4pt}$\star$}	0.0pt
0.75466	0.544066	f	{$\cdot$}	0.3pt
0.7607	1.056322627702333	l	{\fontsize{4pt}{4pt}$\star$}	0.0pt
0.68364	0.9957050934119935	l	{\fontsize{4pt}{4pt}$\star$}	0.0pt
0.82038	1.028026448095295	l	{\fontsize{4pt}{4pt}$\star$}	0.0pt
0.63376	0.5339230369569176	l	{\fontsize{4pt}{4pt}$\star$}	0.0pt
0.76764	0.8828160294581722	l	{\fontsize{4pt}{4pt}$\star$}	0.0pt
0.81632	0.2371766689731943	d	{$ $}	0.0pt
0.82414	0.2819503022510937	d	{$ $}	0.0pt
0.81958	0.345129	d	{$ $}	0.0pt
0.80532	0.307806	d	{$ $}	0.0pt
0.81516	0.3229998982860607	d	{$ $}	0.0pt
0.82388	0.231765	d	{$ $}	0.0pt
0.8212	0.2841195427388228	d	{$ $}	0.0pt
0.81772	0.315178238058445	d	{$ $}	0.0pt
0.80308	0.270826	d	{$ $}	0.0pt
0.81678	0.29160505220003	d	{$ $}	0.0pt
0.8229	0.2358938894606034	d	{$ $}	0.0pt
0.8229	0.2747252263212832	d	{$ $}	0.0pt
0.80546	0.2877464498404197	d	{$ $}	0.0pt
0.78838	0.2317562941401736	d	{$ $}	0.0pt
0.8159	0.2576141115597082	d	{$ $}	0.0pt
0.79482	0.5388254669088318	k	{$ $}	0.0pt
0.77104	0.5175636780999646	k	{$ $}	0.0pt
0.76626	0.4381439480973191	k	{$ $}	0.0pt
0.76492	0.4028595851615789	k	{$ $}	0.0pt
0.74408	0.54596994246795	k	{$ $}	0.0pt
0.79228	0.2750545208556722	a	{$ $}	0.0pt
0.7725	1.09619676160778	l	{\fontsize{4pt}{4pt}$\star$}	0.0pt
0.76902	0.3993347802857467	h	{$ $}	0.0pt
0.80462	1.036640651017411	l	{\fontsize{4pt}{4pt}$\star$}	0.0pt
0.76526	0.4918420614798035	k	{$ $}	0.0pt
0.7213	1.087412792958675	l	{\fontsize{4pt}{4pt}$\star$}	0.0pt
0.75898	1.062919339393656	l	{\fontsize{4pt}{4pt}$\star$}	0.0pt
0.79168	1.072730812259283	l	{\fontsize{4pt}{4pt}$\star$}	0.0pt
0.72786	0.7161306227940001	g	{$ $}	0.0pt
    };
\end{axis}
\end{tikzpicture}
        \end{subfigure}
        \hspace{-2.5em}
        \begin{subfigure}{0.275\linewidth}
           \centering
           \begin{tikzpicture}[every node/.style={font=\sffamily}]
\sffamily
\scriptsize

\pgfplotsset{scaled y ticks=false}

\begin{axis}[
            title={Cal. Error\strut},
            title style={yshift=-1em},
            width=1\linewidth,
            minor y tick num=0,
            xtick={0.6,0.7, 0.8, 0.9},
            xticklabels={0.6,0.7, 0.8, 0.9},
            ytick={0,0.02, 0.04},
            yticklabels={0,0.02, 0.04},
            label style={font=\scriptsize},
            tick label style={font=\scriptsize},
            x label style={at={(axis description cs:0.5,+0.05)},anchor=north},
            y label style={at={(axis description cs:+0.08,.5)},anchor=center},
            ]
    \addplot [
        scatter,
        only marks,
        point meta=explicit symbolic,
        scatter/classes={
            a={mark=square*,fill=cnncolor,draw=cnncolor,mark size=2pt, opacity=0.5, draw opacity=0.7},
            b={mark=*,fill=cnncolor,draw=cnncolor, mark size=2.2pt, opacity=0.5, draw opacity=0.7},
            c={mark=triangle*,fill=cnncolor,draw=cnncolor, mark size=3pt, opacity=0.5, draw opacity=0.7},
            d={mark=pentagon*,fill=cnncolor,draw=cnncolor, mark size=2.5pt, opacity=0.5, draw opacity=0.7},
            e={mark=diamond*,fill=cnncolor,draw=cnncolor, mark size=3pt, opacity=0.5, draw opacity=0.7},
            f={mark=square*,fill=transformercolor,draw=transformercolor,mark size=2pt, opacity=0.5, draw opacity=0.7},
            g={mark=*,fill=transformercolor,draw=transformercolor, mark size=2.2pt, opacity=0.5, draw opacity=0.7},
            h={mark=triangle*,fill=transformercolor,draw=transformercolor, mark size=3pt, opacity=0.5, draw opacity=0.7},
            i={mark=pentagon*,fill=transformercolor,draw=transformercolor, mark size=2.5pt, opacity=0.5, draw opacity=0.7},
            j={mark=diamond*,fill=transformercolor,draw=transformercolor, mark size=3pt, opacity=0.5, draw opacity=0.7},
            k={mark=square*,fill=bcoscolor,draw=bcoscolor,mark size=2pt, opacity=0.5, draw opacity=0.7},
            l={mark=triangle*,fill=vilcolor,draw=vilcolor, mark size=3pt, opacity=0.5, draw opacity=0.7}
        },
        visualization depends on={value \thisrow{symbol} \as \symbol},
        visualization depends on={value \thisrow{centeroffset} \as \centeroffset},
        nodes near coords*={\symbol},
        every node near coord/.append style={anchor=center, yshift=-\centeroffset, opacity=0.5},
    ] table [meta=label] {
x	y	label	symbol	centeroffset
0.56516	0.001676800452785996	a	{$ $}	0.0pt
0.69784	0.003222704405069498	a	{$ $}	0.0pt
0.74216	0.001440463081460051	a	{$ $}	0.0pt
0.7832	0.001841679824609006	a	{$ $}	0.0pt
0.78844	0.002300077635806823	a	{$ $}	0.0pt
0.58182	0.0012663300499381	a	{$ $}	0.0pt
0.80132	0.00172038227687827	a	{$ $}	0.0pt
0.80446	0.002158485668482414	a	{$ $}	0.0pt
0.83246	0.00256208316533758	a	{$ $}	0.0pt
0.77114	0.00170170913495647	a	{$ $}	0.0pt
0.7904	0.003097929366011172	a	{$ $}	0.0pt
0.7187	0.001753475941773012	a	{$ $}	0.0pt
0.76226	0.002753486470684412	a	{$ $}	0.0pt
0.82628	0.002884798421595587	a	{$ $}	0.0pt
0.74046	0.001694585472935624	a	{$ $}	0.0pt
0.779	0.002000442602931861	a	{$\cdot$}	0.3pt
0.64526	0.00522754898101289	a	{$ $}	0.0pt
0.76514	0.004137652941320087	a	{$ $}	0.0pt
0.84124	0.003369122043621742	a	{$ $}	0.0pt
0.86828	0.003707961721016822	e	{\fontsize{4pt}{4pt}$\star$}	0.0pt
0.68404	0.002707248327608148	b	{$ $}	0.0pt
0.63862	0.003697003046706335	b	{$ $}	0.0pt
0.84508	0.001212005874537776	a	{$\cdot$}	0.3pt
0.80874	0.001834866028560826	a	{$ $}	0.0pt
0.81066	0.003442199654096709	f	{$ $}	0.0pt
0.8358	0.003670161662086606	f	{$ $}	0.0pt
0.83162	0.002685053614947841	a	{$ $}	0.0pt
0.83332	0.004018660231148619	a	{$\cdot$}	0.3pt
0.84556	0.005564006160079564	a	{$ $}	0.0pt
0.84772	0.004309530943292663	a	{$\cdot$}	0.3pt
0.852	0.005230558199515285	a	{$ $}	0.0pt
0.85822	0.004591782317029873	a	{$\cdot$}	0.3pt
0.8198	0.005321863826514815	f	{$ $}	0.0pt
0.8229	0.00511879880623407	f	{$ $}	0.0pt
0.85044	0.0021753733330679	f	{$ $}	0.0pt
0.82502	0.00411585279740875	f	{$ $}	0.0pt
0.82896	0.00400821931169956	f	{$ $}	0.0pt
0.82588	0.00171668232701026	f	{$ $}	0.0pt
0.82446	0.005143173799261745	f	{$ $}	0.0pt
0.836	0.00528525797615645	f	{$ $}	0.0pt
0.8493	0.00369509693943328	f	{$ $}	0.0pt
0.81358	0.002287100798023914	f	{$ $}	0.0pt
0.83068	0.003294720025928679	f	{$\cdot$}	0.3pt
0.83094	0.004023074125932701	f	{$ $}	0.0pt
0.84558	0.003187583328114395	f	{$\cdot$}	0.3pt
0.83786	0.003645927998730159	f	{$ $}	0.0pt
0.85708	0.002696452231887299	f	{$\cdot$}	0.3pt
0.84778	0.00312428342209592	f	{$ $}	0.0pt
0.86976	0.002353577097000088	f	{$\cdot$}	0.3pt
0.85252	0.005365528968598612	f	{$ $}	0.0pt
0.84598	0.004394547804829193	f	{$ $}	0.0pt
0.86942	0.004044197725687283	f	{$\cdot$}	0.3pt
0.79172	0.01100868686457631	f	{$ $}	0.0pt
0.80882	0.004144470689899034	f	{$\cdot$}	0.3pt
0.81502	0.009567142966689111	f	{$ $}	0.0pt
0.8324	0.003882865751042412	f	{$\cdot$}	0.3pt
0.83248	0.007707742624320476	f	{$ $}	0.0pt
0.85084	0.003409082168447404	f	{$\cdot$}	0.3pt
0.78836	0.001775287098830543	f	{$ $}	0.0pt
0.85838	0.0009816403198247383	f	{$\cdot$}	0.3pt
0.82516	0.007673144979457452	a	{$ $}	0.0pt
0.829	0.003704597059291385	a	{$\cdot$}	0.3pt
0.83634	0.00832501725909798	a	{$ $}	0.0pt
0.8459	0.00223005704785121	a	{$\cdot$}	0.3pt
0.84062	0.008896685592050618	a	{$ $}	0.0pt
0.85812	0.001877221326254333	a	{$\cdot$}	0.3pt
0.84414	0.006389657511369636	a	{$ $}	0.0pt
0.86606	0.001601321535470665	a	{$\cdot$}	0.3pt
0.8522	0.003610626128976895	h	{$\cdot$}	0.0pt
0.83374	0.001433992079812562	f	{$ $}	0.0pt
0.8465	0.003827178526308601	f	{$ $}	0.0pt
0.78744	0.008266095215734771	g	{$ $}	0.0pt
0.78118	0.007474499614966849	b	{$ $}	0.0pt
0.76074	0.006869609805013112	b	{$ $}	0.0pt
0.7743	0.006936115762525458	b	{$ $}	0.0pt
0.76718	0.007167798318876728	g	{$ $}	0.0pt
0.81864	0.006862908025512898	c	{$ $}	0.0pt
0.8203	0.002741012341403353	c	{$\cdot$}	0.0pt
0.82938	0.002614266656903989	c	{$ $}	0.0pt
0.83892	0.002558327636642288	c	{$\cdot$}	0.0pt
0.84876	0.003255135515382089	c	{$ $}	0.0pt
0.86744	0.00232706916998017	c	{$\cdot$}	0.0pt
0.8576	0.002826376021991549	c	{$ $}	0.0pt
0.87264	0.002256391305272784	c	{$\cdot$}	0.0pt
0.86068	0.01140188384889795	h	{$ $}	0.0pt
0.88264	0.003146152716505817	h	{$\cdot$}	0.0pt
0.84088	0.008162420142154509	f	{$ $}	0.0pt
0.835	0.00645454978438165	f	{$ $}	0.0pt
0.85602	0.003497699757570695	h	{$ $}	0.0pt
0.81226	0.009097635020845719	a	{$ $}	0.0pt
0.79078	0.01237103016754353	a	{$ $}	0.0pt
0.80314	0.008810788980503647	e	{\fontsize{4pt}{4pt}$\star$}	0.0pt
0.81138	0.002158090672216784	e	{\fontsize{4pt}{4pt}$\star$}	0.0pt
0.84278	0.01210975082532946	e	{\fontsize{4pt}{4pt}$\star$}	0.0pt
0.73286	0.006020984986758126	e	{\fontsize{4pt}{4pt}$\star$}	0.0pt
0.83742	0.004863628381437634	h	{$ $}	0.0pt
0.66754	0.007316406713124659	h	{$ $}	0.0pt
0.65178	0.002966830602026058	h	{$ $}	0.0pt
0.78064	0.001177854496932932	h	{$ $}	0.0pt
0.82858	0.003938382337494273	h	{$ $}	0.0pt
0.75292	0.008448507543417947	c	{$ $}	0.0pt
0.78456	0.0201216923768158	c	{$ $}	0.0pt
0.86398	0.001014719154455633	h	{\fontsize{4pt}{4pt}$\star$}	0.0pt
0.75466	0.000736	f	{$\cdot$}	0.3pt
0.7607	0.03730235986818582	l	{\fontsize{4pt}{4pt}$\star$}	0.0pt
0.68364	0.01090004581598756	l	{\fontsize{4pt}{4pt}$\star$}	0.0pt
0.82038	0.03873310684679587	l	{\fontsize{4pt}{4pt}$\star$}	0.0pt
0.63376	0.02567287017950134	l	{\fontsize{4pt}{4pt}$\star$}	0.0pt
0.76764	0.0324753425554671	l	{\fontsize{4pt}{4pt}$\star$}	0.0pt
0.81632	0.002682468642875588	d	{$ $}	0.0pt
0.82414	0.002162777021152435	d	{$ $}	0.0pt
0.81958	0.002811	d	{$ $}	0.0pt
0.80532	0.00342	d	{$ $}	0.0pt
0.81516	0.003504045155600947	d	{$ $}	0.0pt
0.82388	0.002587	d	{$ $}	0.0pt
0.8212	0.002077206754690432	d	{$ $}	0.0pt
0.81772	0.002647412678832452	d	{$ $}	0.0pt
0.80308	0.002595	d	{$ $}	0.0pt
0.81678	0.003157430379554663	d	{$ $}	0.0pt
0.8229	0.005406788646685873	d	{$ $}	0.0pt
0.8229	0.005022813131673387	d	{$ $}	0.0pt
0.80546	0.006717421603401008	d	{$ $}	0.0pt
0.78838	0.007486399292044436	d	{$ $}	0.0pt
0.8159	0.002460236896126301	d	{$ $}	0.0pt
0.79482	0.01608333161189769	k	{$ $}	0.0pt
0.77104	0.01165835902470626	k	{$ $}	0.0pt
0.76626	0.009872575187920384	k	{$ $}	0.0pt
0.76492	0.02070844555269459	k	{$ $}	0.0pt
0.74408	0.004158495920770831	k	{$ $}	0.0pt
0.79228	0.001485428057691981	a	{$ $}	0.0pt
0.7725	0.03750750940296452	l	{\fontsize{4pt}{4pt}$\star$}	0.0pt
0.76902	0.00122302794122072	h	{$ $}	0.0pt
0.80462	0.03836975607093664	l	{\fontsize{4pt}{4pt}$\star$}	0.0pt
0.76526	0.004209439660289872	k	{$ $}	0.0pt
0.7213	0.03647192830114739	l	{\fontsize{4pt}{4pt}$\star$}	0.0pt
0.75898	0.03697529932997521	l	{\fontsize{4pt}{4pt}$\star$}	0.0pt
0.79168	0.0382271658739983	l	{\fontsize{4pt}{4pt}$\star$}	0.0pt
0.72786	0.008051309669806654	g	{$ $}	0.0pt
    };
\end{axis}
\end{tikzpicture}
        \end{subfigure}
        \\
        \hspace{-1.5em}
        \begin{subfigure}{0.275\linewidth}
            \centering
            \begin{tikzpicture}[every node/.style={font=\sffamily}]
\sffamily
\scriptsize

\begin{axis}[
            title={Class Balance\strut},
            title style={yshift=-1em},
            width=1\linewidth,
            minor y tick num=0,
            xtick={0.6,0.7, 0.8, 0.9},
            xticklabels={0.6,0.7, 0.8, 0.9},
            ytick={0.7,0.8, 0.9},
            yticklabels={0.7,0.8, 0.9},
            label style={font=\scriptsize},
            tick label style={font=\scriptsize},
            x label style={at={(axis description cs:0.5,+0.05)},anchor=north},
            y label style={at={(axis description cs:+0.08,.5)},anchor=center},
            ]
    \addplot [
        scatter,
        only marks,
        point meta=explicit symbolic,
        scatter/classes={
            a={mark=square*,fill=cnncolor,draw=cnncolor,mark size=2pt, opacity=0.5, draw opacity=0.7},
            b={mark=*,fill=cnncolor,draw=cnncolor, mark size=2.2pt, opacity=0.5, draw opacity=0.7},
            c={mark=triangle*,fill=cnncolor,draw=cnncolor, mark size=3pt, opacity=0.5, draw opacity=0.7},
            d={mark=pentagon*,fill=cnncolor,draw=cnncolor, mark size=2.5pt, opacity=0.5, draw opacity=0.7},
            e={mark=diamond*,fill=cnncolor,draw=cnncolor, mark size=3pt, opacity=0.5, draw opacity=0.7},
            f={mark=square*,fill=transformercolor,draw=transformercolor,mark size=2pt, opacity=0.5, draw opacity=0.7},
            g={mark=*,fill=transformercolor,draw=transformercolor, mark size=2.2pt, opacity=0.5, draw opacity=0.7},
            h={mark=triangle*,fill=transformercolor,draw=transformercolor, mark size=3pt, opacity=0.5, draw opacity=0.7},
            i={mark=pentagon*,fill=transformercolor,draw=transformercolor, mark size=2.5pt, opacity=0.5, draw opacity=0.7},
            j={mark=diamond*,fill=transformercolor,draw=transformercolor, mark size=3pt, opacity=0.5, draw opacity=0.7},
            k={mark=square*,fill=bcoscolor,draw=bcoscolor,mark size=2pt, opacity=0.5, draw opacity=0.7},
            l={mark=triangle*,fill=vilcolor,draw=vilcolor, mark size=3pt, opacity=0.5, draw opacity=0.7}
        },
        visualization depends on={value \thisrow{symbol} \as \symbol},
        visualization depends on={value \thisrow{centeroffset} \as \centeroffset},
        nodes near coords*={\symbol},
        every node near coord/.append style={anchor=center, yshift=-\centeroffset, opacity=0.5},
    ] table [meta=label] {
x	y	label	symbol	centeroffset
0.56516	0.7318701238437063	a	{$ $}	0.0pt
0.69784	0.7481539191780298	a	{$ $}	0.0pt
0.74216	0.7414578365374316	a	{$ $}	0.0pt
0.7832	0.7398189333844221	a	{$ $}	0.0pt
0.78844	0.7605427388509575	a	{$ $}	0.0pt
0.58182	0.735187307837025	a	{$ $}	0.0pt
0.80132	0.7602588513258134	a	{$ $}	0.0pt
0.80446	0.7799480506745313	a	{$ $}	0.0pt
0.83246	0.7555765007448868	a	{$ $}	0.0pt
0.77114	0.7526038217552088	a	{$ $}	0.0pt
0.7904	0.7681839718216212	a	{$ $}	0.0pt
0.7187	0.7453606605987513	a	{$ $}	0.0pt
0.76226	0.7216941615163738	a	{$ $}	0.0pt
0.82628	0.794761706329618	a	{$ $}	0.0pt
0.74046	0.7370219485657713	a	{$ $}	0.0pt
0.779	0.7604752997563357	a	{$\cdot$}	0.3pt
0.64526	0.7361979564816274	a	{$ $}	0.0pt
0.76514	0.7498386106449177	a	{$ $}	0.0pt
0.84124	0.7698248297938272	a	{$ $}	0.0pt
0.86828	0.7783885354884892	e	{\fontsize{4pt}{4pt}$\star$}	0.0pt
0.68404	0.7427190549550785	b	{$ $}	0.0pt
0.63862	0.7449245446377324	b	{$ $}	0.0pt
0.84508	0.7780541943365011	a	{$\cdot$}	0.3pt
0.80874	0.7515422145638877	a	{$ $}	0.0pt
0.81066	0.7912695477530669	f	{$ $}	0.0pt
0.8358	0.7900965366320821	f	{$ $}	0.0pt
0.83162	0.7983401759088454	a	{$ $}	0.0pt
0.83332	0.7997726432223973	a	{$\cdot$}	0.3pt
0.84556	0.8159477490527972	a	{$ $}	0.0pt
0.84772	0.8051427196559463	a	{$\cdot$}	0.3pt
0.852	0.8169970658688352	a	{$ $}	0.0pt
0.85822	0.8119439622852834	a	{$\cdot$}	0.3pt
0.8198	0.7707630740555306	f	{$ $}	0.0pt
0.8229	0.7723796178870863	f	{$ $}	0.0pt
0.85044	0.7843646822839092	f	{$ $}	0.0pt
0.82502	0.777810856195609	f	{$ $}	0.0pt
0.82896	0.8009601596568185	f	{$ $}	0.0pt
0.82588	0.764497475640624	f	{$ $}	0.0pt
0.82446	0.7710619290879465	f	{$ $}	0.0pt
0.836	0.7850204581424773	f	{$ $}	0.0pt
0.8493	0.8071227363314202	f	{$ $}	0.0pt
0.81358	0.7731137295758308	f	{$ $}	0.0pt
0.83068	0.8011080148019118	f	{$\cdot$}	0.3pt
0.83094	0.7995181821350237	f	{$ $}	0.0pt
0.84558	0.8070426690816361	f	{$\cdot$}	0.3pt
0.83786	0.8012764583039489	f	{$ $}	0.0pt
0.85708	0.8137108843611847	f	{$\cdot$}	0.3pt
0.84778	0.8054503148034311	f	{$ $}	0.0pt
0.86976	0.8184772603990181	f	{$\cdot$}	0.3pt
0.85252	0.7983395427783049	f	{$ $}	0.0pt
0.84598	0.7966635478038795	f	{$ $}	0.0pt
0.86942	0.8184384824921183	f	{$\cdot$}	0.3pt
0.79172	0.7966032282174313	f	{$ $}	0.0pt
0.80882	0.7860596514022897	f	{$\cdot$}	0.3pt
0.81502	0.8079836248915847	f	{$ $}	0.0pt
0.8324	0.7974984092914432	f	{$\cdot$}	0.3pt
0.83248	0.8126624678247932	f	{$ $}	0.0pt
0.85084	0.8070200055526061	f	{$\cdot$}	0.3pt
0.78836	0.764998587556287	f	{$ $}	0.0pt
0.85838	0.7825510305298493	f	{$\cdot$}	0.3pt
0.82516	0.8054382968377223	a	{$ $}	0.0pt
0.829	0.799162610899697	a	{$\cdot$}	0.3pt
0.83634	0.81467156508543	a	{$ $}	0.0pt
0.8459	0.8038234156692152	a	{$\cdot$}	0.3pt
0.84062	0.8191501782048141	a	{$ $}	0.0pt
0.85812	0.809315654274204	a	{$\cdot$}	0.3pt
0.84414	0.8135313704312792	a	{$ $}	0.0pt
0.86606	0.8126268785537341	a	{$\cdot$}	0.3pt
0.8522	0.8092593942444893	h	{$\cdot$}	0.0pt
0.83374	0.7720415578725591	f	{$ $}	0.0pt
0.8465	0.7976393112988289	f	{$ $}	0.0pt
0.78744	0.7860140289528961	g	{$ $}	0.0pt
0.78118	0.7907111113276568	b	{$ $}	0.0pt
0.76074	0.7835797769586369	b	{$ $}	0.0pt
0.7743	0.7874701011523708	b	{$ $}	0.0pt
0.76718	0.7895105326421096	g	{$ $}	0.0pt
0.81864	0.8028184188347078	c	{$ $}	0.0pt
0.8203	0.7889669382343212	c	{$\cdot$}	0.0pt
0.82938	0.7961077791466672	c	{$ $}	0.0pt
0.83892	0.7988930933227132	c	{$\cdot$}	0.0pt
0.84876	0.8087566947663564	c	{$ $}	0.0pt
0.86744	0.8087566946958217	c	{$\cdot$}	0.0pt
0.8576	0.8134728057674454	c	{$ $}	0.0pt
0.87264	0.8134728057837924	c	{$\cdot$}	0.0pt
0.86068	0.9249990410714748	h	{$ $}	0.0pt
0.88264	0.7872615951755806	h	{$\cdot$}	0.0pt
0.84088	0.8133592038779812	f	{$ $}	0.0pt
0.835	0.7992291119625519	f	{$ $}	0.0pt
0.85602	0.8127843612122255	h	{$ $}	0.0pt
0.81226	0.8039852392507255	a	{$ $}	0.0pt
0.79078	0.7936476278006992	a	{$ $}	0.0pt
0.80314	0.7829827193612791	e	{\fontsize{4pt}{4pt}$\star$}	0.0pt
0.81138	0.765670699790019	e	{\fontsize{4pt}{4pt}$\star$}	0.0pt
0.84278	0.8050931182847553	e	{\fontsize{4pt}{4pt}$\star$}	0.0pt
0.73286	0.7677016762551706	e	{\fontsize{4pt}{4pt}$\star$}	0.0pt
0.83742	0.7971647605207252	h	{$ $}	0.0pt
0.66754	0.7914152576670295	h	{$ $}	0.0pt
0.65178	0.7483741893560183	h	{$ $}	0.0pt
0.78064	0.7673648192242305	h	{$ $}	0.0pt
0.82858	0.8018797997876895	h	{$ $}	0.0pt
0.75292	0.7982981579020811	c	{$ $}	0.0pt
0.78456	0.836347034171477	c	{$ $}	0.0pt
0.86398	0.8087376495556823	h	{\fontsize{4pt}{4pt}$\star$}	0.0pt
0.75466	0.760414	f	{$\cdot$}	0.3pt
0.7607	0.8995921776627556	l	{\fontsize{4pt}{4pt}$\star$}	0.0pt
0.68364	0.9380046884928999	l	{\fontsize{4pt}{4pt}$\star$}	0.0pt
0.82038	0.9061943400472994	l	{\fontsize{4pt}{4pt}$\star$}	0.0pt
0.63376	0.8799441791497208	l	{\fontsize{4pt}{4pt}$\star$}	0.0pt
0.76764	0.8876052243248322	l	{\fontsize{4pt}{4pt}$\star$}	0.0pt
0.81632	0.7375654601944512	d	{$ $}	0.0pt
0.82414	0.7467967640325963	d	{$ $}	0.0pt
0.81958	0.736096	d	{$ $}	0.0pt
0.80532	0.733626	d	{$ $}	0.0pt
0.81516	0.7535395897494822	d	{$ $}	0.0pt
0.82388	0.736132	d	{$ $}	0.0pt
0.8212	0.747663110523794	d	{$ $}	0.0pt
0.81772	0.7457386689778723	d	{$ $}	0.0pt
0.80308	0.740643	d	{$ $}	0.0pt
0.81678	0.7558980796063501	d	{$ $}	0.0pt
0.8229	0.749271784068492	d	{$ $}	0.0pt
0.8229	0.7505696484877644	d	{$ $}	0.0pt
0.80546	0.7516262492189092	d	{$ $}	0.0pt
0.78838	0.7545444854790587	d	{$ $}	0.0pt
0.8159	0.7561835639756832	d	{$ $}	0.0pt
0.79482	0.7733551226920953	k	{$ $}	0.0pt
0.77104	0.7603165380514947	k	{$ $}	0.0pt
0.76626	0.7418649340062764	k	{$ $}	0.0pt
0.76492	0.7711458055496444	k	{$ $}	0.0pt
0.74408	0.7415040433939094	k	{$ $}	0.0pt
0.79228	0.758098595047476	a	{$ $}	0.0pt
0.7725	0.897411071968837	l	{\fontsize{4pt}{4pt}$\star$}	0.0pt
0.76902	0.7612339115103315	h	{$ $}	0.0pt
0.80462	0.9046768168611052	l	{\fontsize{4pt}{4pt}$\star$}	0.0pt
0.76526	0.734841663613316	k	{$ $}	0.0pt
0.7213	0.8912717861540803	l	{\fontsize{4pt}{4pt}$\star$}	0.0pt
0.75898	0.896321716016582	l	{\fontsize{4pt}{4pt}$\star$}	0.0pt
0.79168	0.8969375798408549	l	{\fontsize{4pt}{4pt}$\star$}	0.0pt
0.72786	0.7831856755042765	g	{$ $}	0.0pt
    };
\end{axis}
\end{tikzpicture}
        \end{subfigure}
        \hspace{-2.5em}
        \begin{subfigure}{0.275\linewidth}
           \centering
           \begin{tikzpicture}[every node/.style={font=\sffamily}]
\sffamily
\scriptsize

\begin{axis}[
            title={Obj. Focus\strut},
            title style={yshift=-1em},
            width=1\linewidth,
            minor y tick num=0,
            xtick={0.6,0.7, 0.8, 0.9},
            xticklabels={0.6,0.7, 0.8, 0.9},
            ytick={0.7,0.8, 0.9},
            yticklabels={0.7,0.8, 0.9},
            label style={font=\scriptsize},
            tick label style={font=\scriptsize},
            x label style={at={(axis description cs:0.5,+0.05)},anchor=north},
            y label style={at={(axis description cs:+0.08,.5)},anchor=center},
            ]
    \addplot [
        scatter,
        only marks,
        point meta=explicit symbolic,
        scatter/classes={
            a={mark=square*,fill=cnncolor,draw=cnncolor,mark size=2pt, opacity=0.5, draw opacity=0.7},
            b={mark=*,fill=cnncolor,draw=cnncolor, mark size=2.2pt, opacity=0.5, draw opacity=0.7},
            c={mark=triangle*,fill=cnncolor,draw=cnncolor, mark size=3pt, opacity=0.5, draw opacity=0.7},
            d={mark=pentagon*,fill=cnncolor,draw=cnncolor, mark size=2.5pt, opacity=0.5, draw opacity=0.7},
            e={mark=diamond*,fill=cnncolor,draw=cnncolor, mark size=3pt, opacity=0.5, draw opacity=0.7},
            f={mark=square*,fill=transformercolor,draw=transformercolor,mark size=2pt, opacity=0.5, draw opacity=0.7},
            g={mark=*,fill=transformercolor,draw=transformercolor, mark size=2.2pt, opacity=0.5, draw opacity=0.7},
            h={mark=triangle*,fill=transformercolor,draw=transformercolor, mark size=3pt, opacity=0.5, draw opacity=0.7},
            i={mark=pentagon*,fill=transformercolor,draw=transformercolor, mark size=2.5pt, opacity=0.5, draw opacity=0.7},
            j={mark=diamond*,fill=transformercolor,draw=transformercolor, mark size=3pt, opacity=0.5, draw opacity=0.7},
            k={mark=square*,fill=bcoscolor,draw=bcoscolor,mark size=2pt, opacity=0.5, draw opacity=0.7},
            l={mark=triangle*,fill=vilcolor,draw=vilcolor, mark size=3pt, opacity=0.5, draw opacity=0.7}
        },
        visualization depends on={value \thisrow{symbol} \as \symbol},
        visualization depends on={value \thisrow{centeroffset} \as \centeroffset},
        nodes near coords*={\symbol},
        every node near coord/.append style={anchor=center, yshift=-\centeroffset, opacity=0.5},
    ] table [meta=label] {
x	y	label	symbol	centeroffset
0.56516	0.8000310868703381	a	{$ $}	0.0pt
0.69784	0.91402	a	{$ $}	0.0pt
0.74216	0.86160995893356	a	{$ $}	0.0pt
0.7832	0.8793074584097396	a	{$ $}	0.0pt
0.78844	0.9364516688200742	a	{$ $}	0.0pt
0.58182	0.771258430866786	a	{$ $}	0.0pt
0.80132	0.9386661223285024	a	{$ $}	0.0pt
0.80446	0.9484356195690508	a	{$ $}	0.0pt
0.83246	0.934779945946891	a	{$ $}	0.0pt
0.77114	0.919393841689505	a	{$ $}	0.0pt
0.7904	0.9322112260708751	a	{$ $}	0.0pt
0.7187	0.8790683452804875	a	{$ $}	0.0pt
0.76226	0.8475187757796214	a	{$ $}	0.0pt
0.82628	0.9381431099492774	a	{$ $}	0.0pt
0.74046	0.8372543717818057	a	{$ $}	0.0pt
0.779	0.884308784608314	a	{$\cdot$}	0.3pt
0.64526	0.805329338540369	a	{$ $}	0.0pt
0.76514	0.8389472249732368	a	{$ $}	0.0pt
0.84124	0.8948046194437929	a	{$ $}	0.0pt
0.86828	0.8929683333788454	e	{\fontsize{4pt}{4pt}$\star$}	0.0pt
0.68404	0.9068740991964102	b	{$ $}	0.0pt
0.63862	0.8774593654096007	b	{$ $}	0.0pt
0.84508	0.91333993424898	a	{$\cdot$}	0.3pt
0.80874	0.8645787782937795	a	{$ $}	0.0pt
0.81066	0.9348207601361331	f	{$ $}	0.0pt
0.8358	0.9242354676808204	f	{$ $}	0.0pt
0.83162	0.9165692070303952	a	{$ $}	0.0pt
0.83332	0.9348153362921245	a	{$\cdot$}	0.3pt
0.84556	0.9465618227149046	a	{$ $}	0.0pt
0.84772	0.9507745722994828	a	{$\cdot$}	0.3pt
0.852	0.9356633628934099	a	{$ $}	0.0pt
0.85822	0.9605845115543483	a	{$\cdot$}	0.3pt
0.8198	0.9011168661779306	f	{$ $}	0.0pt
0.8229	0.909153417925347	f	{$ $}	0.0pt
0.85044	0.937167460308745	f	{$ $}	0.0pt
0.82502	0.928157532570456	f	{$ $}	0.0pt
0.82896	0.927643934005288	f	{$ $}	0.0pt
0.82588	0.918501022985228	f	{$ $}	0.0pt
0.82446	0.896058483523746	f	{$ $}	0.0pt
0.836	0.9267337657952445	f	{$ $}	0.0pt
0.8493	0.9303787367221779	f	{$ $}	0.0pt
0.81358	0.9471927002102276	f	{$ $}	0.0pt
0.83068	0.9236086217033554	f	{$\cdot$}	0.3pt
0.83094	0.9500788486756186	f	{$ $}	0.0pt
0.84558	0.9267067472717243	f	{$\cdot$}	0.3pt
0.83786	0.9543221959059132	f	{$ $}	0.0pt
0.85708	0.9492966932801409	f	{$\cdot$}	0.3pt
0.84778	0.9586428927374299	f	{$ $}	0.0pt
0.86976	0.9554297368253708	f	{$\cdot$}	0.3pt
0.85252	0.949359972983995	f	{$ $}	0.0pt
0.84598	0.922982409904462	f	{$ $}	0.0pt
0.86942	0.9602186201470652	f	{$\cdot$}	0.3pt
0.79172	0.9270224602412751	f	{$ $}	0.0pt
0.80882	0.9206281609453308	f	{$\cdot$}	0.3pt
0.81502	0.9360767031090049	f	{$ $}	0.0pt
0.8324	0.9477933804379424	f	{$\cdot$}	0.3pt
0.83248	0.9463153973372437	f	{$ $}	0.0pt
0.85084	0.9515366368515413	f	{$\cdot$}	0.3pt
0.78836	0.9119910750806957	f	{$ $}	0.0pt
0.85838	0.9371945058846017	f	{$\cdot$}	0.3pt
0.82516	0.9245919211016044	a	{$ $}	0.0pt
0.829	0.9481749542063174	a	{$\cdot$}	0.3pt
0.83634	0.9380014684795037	a	{$ $}	0.0pt
0.8459	0.963221289619209	a	{$\cdot$}	0.3pt
0.84062	0.9497725223974897	a	{$ $}	0.0pt
0.85812	0.9634572965420245	a	{$\cdot$}	0.3pt
0.84414	0.9473494391399006	a	{$ $}	0.0pt
0.86606	0.9694943155365451	a	{$\cdot$}	0.3pt
0.8522	0.9608855511225847	h	{$\cdot$}	0.0pt
0.83374	0.9392699698282169	f	{$ $}	0.0pt
0.8465	0.9408840521295176	f	{$ $}	0.0pt
0.78744	0.9408343767705474	g	{$ $}	0.0pt
0.78118	0.946549959944476	b	{$ $}	0.0pt
0.76074	0.9457810867548067	b	{$ $}	0.0pt
0.7743	0.9505751931473623	b	{$ $}	0.0pt
0.76718	0.9504357629786689	g	{$ $}	0.0pt
0.81864	0.9442013432697169	c	{$ $}	0.0pt
0.8203	0.9404012082300514	c	{$\cdot$}	0.0pt
0.82938	0.9395651489183204	c	{$ $}	0.0pt
0.83892	0.9487877712261444	c	{$\cdot$}	0.0pt
0.84876	0.9566542680418609	c	{$ $}	0.0pt
0.86744	0.9607188115903416	c	{$\cdot$}	0.0pt
0.8576	0.9691933849272157	c	{$ $}	0.0pt
0.87264	0.9654800871033606	c	{$\cdot$}	0.0pt
0.86068	0.928200185064023	h	{$ $}	0.0pt
0.88264	0.9298697568073038	h	{$\cdot$}	0.0pt
0.84088	0.9416315333396192	f	{$ $}	0.0pt
0.835	0.938923595555714	f	{$ $}	0.0pt
0.85602	0.9561565568590327	h	{$ $}	0.0pt
0.81226	0.9273620640992849	a	{$ $}	0.0pt
0.79078	0.946080476308045	a	{$ $}	0.0pt
0.80314	0.932535574271493	e	{\fontsize{4pt}{4pt}$\star$}	0.0pt
0.81138	0.8975967842097535	e	{\fontsize{4pt}{4pt}$\star$}	0.0pt
0.84278	0.9457744493148849	e	{\fontsize{4pt}{4pt}$\star$}	0.0pt
0.73286	0.9048726748127137	e	{\fontsize{4pt}{4pt}$\star$}	0.0pt
0.83742	0.9472217873179557	h	{$ $}	0.0pt
0.66754	0.8505661287417187	h	{$ $}	0.0pt
0.65178	0.8450586788164908	h	{$ $}	0.0pt
0.78064	0.9104881338262495	h	{$ $}	0.0pt
0.82858	0.9427847553556599	h	{$ $}	0.0pt
0.75292	0.6898070599913144	c	{$ $}	0.0pt
0.78456	0.923022582994258	c	{$ $}	0.0pt
0.86398	0.9425569693399967	h	{\fontsize{4pt}{4pt}$\star$}	0.0pt
0.75466	0.8652	f	{$\cdot$}	0.3pt
0.7607	0.9487152022097906	l	{\fontsize{4pt}{4pt}$\star$}	0.0pt
0.68364	0.8637797888834401	l	{\fontsize{4pt}{4pt}$\star$}	0.0pt
0.82038	0.9680966875031414	l	{\fontsize{4pt}{4pt}$\star$}	0.0pt
0.63376	0.9065466528945945	l	{\fontsize{4pt}{4pt}$\star$}	0.0pt
0.76764	0.8842935144687915	l	{\fontsize{4pt}{4pt}$\star$}	0.0pt
0.81632	0.9066898319301311	d	{$ $}	0.0pt
0.82414	0.9396490451259412	d	{$ $}	0.0pt
0.81958	0.918838	d	{$ $}	0.0pt
0.80532	0.917033	d	{$ $}	0.0pt
0.81516	0.93170607536545	d	{$ $}	0.0pt
0.82388	0.900603	d	{$ $}	0.0pt
0.8212	0.9369446185525871	d	{$ $}	0.0pt
0.81772	0.9274128807850776	d	{$ $}	0.0pt
0.80308	0.941233	d	{$ $}	0.0pt
0.81678	0.9401484869997909	d	{$ $}	0.0pt
0.8229	0.864675188982672	d	{$ $}	0.0pt
0.8229	0.9416318069889623	d	{$ $}	0.0pt
0.80546	0.9263958600502207	d	{$ $}	0.0pt
0.78838	0.9032530751701106	d	{$ $}	0.0pt
0.8159	0.9362082942647288	d	{$ $}	0.0pt
0.79482	0.9369373483428658	k	{$ $}	0.0pt
0.77104	0.875108559482911	k	{$ $}	0.0pt
0.76626	0.9023745590156068	k	{$ $}	0.0pt
0.76492	0.9080289818720283	k	{$ $}	0.0pt
0.74408	0.8894981423741772	k	{$ $}	0.0pt
0.79228	0.8968715379996747	a	{$ $}	0.0pt
0.7725	0.9510967277957569	l	{\fontsize{4pt}{4pt}$\star$}	0.0pt
0.76902	0.9139517314163834	h	{$ $}	0.0pt
0.80462	0.9656305821534957	l	{\fontsize{4pt}{4pt}$\star$}	0.0pt
0.76526	0.9314577988577378	k	{$ $}	0.0pt
0.7213	0.9195553980895264	l	{\fontsize{4pt}{4pt}$\star$}	0.0pt
0.75898	0.94371913702243	l	{\fontsize{4pt}{4pt}$\star$}	0.0pt
0.79168	0.9666282423625457	l	{\fontsize{4pt}{4pt}$\star$}	0.0pt
0.72786	0.9392774285381839	g	{$ $}	0.0pt
    };
\end{axis}
\end{tikzpicture}
        \end{subfigure}
        \hspace{-2.5em}
        \begin{subfigure}{0.275\linewidth}
           \centering
           \begin{tikzpicture}[every node/.style={font=\sffamily}]
\sffamily
\scriptsize

\begin{axis}[
            title={Shape Bias\strut},
            title style={yshift=-1em},
            width=1\linewidth,
            minor y tick num=0,
            xtick={0.6,0.7, 0.8, 0.9},
            xticklabels={0.6,0.7, 0.8, 0.9},
            ytick={0.0, 0.2, 0.4,0.6,0.8},
            yticklabels={0, 0.2,0.4,0.6,0.8},
            label style={font=\scriptsize},
            tick label style={font=\scriptsize},
            x label style={at={(axis description cs:0.5,+0.05)},anchor=north},
            y label style={at={(axis description cs:+0.08,.5)},anchor=center},
            ]
    \addplot [
        scatter,
        only marks,
        point meta=explicit symbolic,
        scatter/classes={
            a={mark=square*,fill=cnncolor,draw=cnncolor,mark size=2pt, opacity=0.5, draw opacity=0.7},
            b={mark=*,fill=cnncolor,draw=cnncolor, mark size=2.2pt, opacity=0.5, draw opacity=0.7},
            c={mark=triangle*,fill=cnncolor,draw=cnncolor, mark size=3pt, opacity=0.5, draw opacity=0.7},
            d={mark=pentagon*,fill=cnncolor,draw=cnncolor, mark size=2.5pt, opacity=0.5, draw opacity=0.7},
            e={mark=diamond*,fill=cnncolor,draw=cnncolor, mark size=3pt, opacity=0.5, draw opacity=0.7},
            f={mark=square*,fill=transformercolor,draw=transformercolor,mark size=2pt, opacity=0.5, draw opacity=0.7},
            g={mark=*,fill=transformercolor,draw=transformercolor, mark size=2.2pt, opacity=0.5, draw opacity=0.7},
            h={mark=triangle*,fill=transformercolor,draw=transformercolor, mark size=3pt, opacity=0.5, draw opacity=0.7},
            i={mark=pentagon*,fill=transformercolor,draw=transformercolor, mark size=2.5pt, opacity=0.5, draw opacity=0.7},
            j={mark=diamond*,fill=transformercolor,draw=transformercolor, mark size=3pt, opacity=0.5, draw opacity=0.7},
            k={mark=square*,fill=bcoscolor,draw=bcoscolor,mark size=2pt, opacity=0.5, draw opacity=0.7},
            l={mark=triangle*,fill=vilcolor,draw=vilcolor, mark size=3pt, opacity=0.5, draw opacity=0.7}
        },
        visualization depends on={value \thisrow{symbol} \as \symbol},
        visualization depends on={value \thisrow{centeroffset} \as \centeroffset},
        nodes near coords*={\symbol},
        every node near coord/.append style={anchor=center, yshift=-\centeroffset, opacity=0.5},
    ] table [meta=label] {
x	y	label	symbol	centeroffset
0.56516	0.2635869565217391	a	{$ $}	0.0pt
0.69784	0.2505694760820046	a	{$ $}	0.0pt
0.74216	0.1052631578947368	a	{$ $}	0.0pt
0.7832	0.2366863905325444	a	{$ $}	0.0pt
0.78844	0.230379746835443	a	{$ $}	0.0pt
0.58182	0.213468869123252	a	{$ $}	0.0pt
0.80132	0.295369211514393	a	{$ $}	0.0pt
0.80446	0.2640845070422535	a	{$ $}	0.0pt
0.83246	0.218911917098445	a	{$ $}	0.0pt
0.77114	0.22209026128266	a	{$ $}	0.0pt
0.7904	0.2159090909090909	a	{$ $}	0.0pt
0.7187	0.1804511278195489	a	{$ $}	0.0pt
0.76226	0.267022696929239	a	{$ $}	0.0pt
0.82628	0.239154616240267	a	{$ $}	0.0pt
0.74046	0.3189189189189189	a	{$ $}	0.0pt
0.779	0.2699468085106383	a	{$\cdot$}	0.3pt
0.64526	0.0389908256880733	a	{$ $}	0.0pt
0.76514	0.2346437346437346	a	{$ $}	0.0pt
0.84124	0.2488038277511962	a	{$ $}	0.0pt
0.86828	0.2354368932038835	e	{\fontsize{4pt}{4pt}$\star$}	0.0pt
0.68404	0.6670918367346937	b	{$ $}	0.0pt
0.63862	0.6429495472186286	b	{$ $}	0.0pt
0.84508	0.1851441241685144	a	{$\cdot$}	0.3pt
0.80874	0.2218045112781955	a	{$ $}	0.0pt
0.81066	0.398046398046398	f	{$ $}	0.0pt
0.8358	0.2190692395005676	f	{$ $}	0.0pt
0.83162	0.2398568019093079	a	{$ $}	0.0pt
0.83332	0.3403019744483159	a	{$\cdot$}	0.3pt
0.84556	0.2515262515262515	a	{$ $}	0.0pt
0.84772	0.3646080760095012	a	{$\cdot$}	0.3pt
0.852	0.265792610250298	a	{$ $}	0.0pt
0.85822	0.3826291079812207	a	{$\cdot$}	0.3pt
0.8198	0.2434584755403868	f	{$ $}	0.0pt
0.8229	0.2668918918918919	f	{$ $}	0.0pt
0.85044	0.2285714285714286	f	{$ $}	0.0pt
0.82502	0.262202043132803	f	{$ $}	0.0pt
0.82896	0.321644498186215	f	{$ $}	0.0pt
0.82588	0.263157894736842	f	{$ $}	0.0pt
0.82446	0.3017142857142857	f	{$ $}	0.0pt
0.836	0.2538552787663108	f	{$ $}	0.0pt
0.8493	0.275	f	{$ $}	0.0pt
0.81358	0.3404255319148937	f	{$ $}	0.0pt
0.83068	0.3502475247524753	f	{$\cdot$}	0.3pt
0.83094	0.423030303030303	f	{$ $}	0.0pt
0.84558	0.3554006968641115	f	{$\cdot$}	0.3pt
0.83786	0.4318936877076412	f	{$ $}	0.0pt
0.85708	0.3985932004689332	f	{$\cdot$}	0.3pt
0.84778	0.5017142857142857	f	{$ $}	0.0pt
0.86976	0.5130718954248366	f	{$\cdot$}	0.3pt
0.85252	0.2922732362821948	f	{$ $}	0.0pt
0.84598	0.1843003412969283	f	{$ $}	0.0pt
0.86942	0.4076840981856991	f	{$\cdot$}	0.3pt
0.79172	0.2435754189944134	f	{$ $}	0.0pt
0.80882	0.3252858958068615	f	{$\cdot$}	0.3pt
0.81502	0.2603911980440098	f	{$ $}	0.0pt
0.8324	0.3328964613368283	f	{$\cdot$}	0.3pt
0.83248	0.3163636363636364	f	{$ $}	0.0pt
0.85084	0.3698296836982969	f	{$\cdot$}	0.3pt
0.78836	0.2987551867219917	f	{$ $}	0.0pt
0.85838	0.3819163292847504	f	{$\cdot$}	0.3pt
0.82516	0.2513721185510428	a	{$ $}	0.0pt
0.829	0.2865429234338747	a	{$\cdot$}	0.3pt
0.83634	0.2535991140642303	a	{$ $}	0.0pt
0.8459	0.3115124153498872	a	{$\cdot$}	0.3pt
0.84062	0.2982261640798227	a	{$ $}	0.0pt
0.85812	0.3347826086956522	a	{$\cdot$}	0.3pt
0.84414	0.3112582781456954	a	{$ $}	0.0pt
0.86606	0.4008620689655172	a	{$\cdot$}	0.3pt
0.8522	0.5292682926829269	h	{$\cdot$}	0.0pt
0.83374	0.2057142857142857	f	{$ $}	0.0pt
0.8465	0.3393501805054152	f	{$ $}	0.0pt
0.78744	0.7301401869158879	g	{$ $}	0.0pt
0.78118	0.7264472190692396	b	{$ $}	0.0pt
0.76074	0.7361894024802706	b	{$ $}	0.0pt
0.7743	0.7441601779755284	b	{$ $}	0.0pt
0.76718	0.7729852440408627	g	{$ $}	0.0pt
0.81864	0.287317620650954	c	{$ $}	0.0pt
0.8203	0.3158522050059595	c	{$\cdot$}	0.0pt
0.82938	0.3190529875986471	c	{$ $}	0.0pt
0.83892	0.3480861244019139	c	{$\cdot$}	0.0pt
0.84876	0.4229432213209733	c	{$ $}	0.0pt
0.86744	0.4023479188900747	c	{$\cdot$}	0.0pt
0.8576	0.4187298170075349	c	{$ $}	0.0pt
0.87264	0.4472807991120977	c	{$\cdot$}	0.0pt
0.86068	0.2702380952380952	h	{$ $}	0.0pt
0.88264	0.2050113895216401	h	{$\cdot$}	0.0pt
0.84088	0.3078848560700876	f	{$ $}	0.0pt
0.835	0.207525655644241	f	{$ $}	0.0pt
0.85602	0.451685393258427	h	{$ $}	0.0pt
0.81226	0.3099173553719008	a	{$ $}	0.0pt
0.79078	0.2065934065934066	a	{$ $}	0.0pt
0.80314	0.2444444444444444	e	{\fontsize{4pt}{4pt}$\star$}	0.0pt
0.81138	0.2174462705436157	e	{\fontsize{4pt}{4pt}$\star$}	0.0pt
0.84278	0.4234620886981402	e	{\fontsize{4pt}{4pt}$\star$}	0.0pt
0.73286	0.2718327183271833	e	{\fontsize{4pt}{4pt}$\star$}	0.0pt
0.83742	0.3646759847522236	h	{$ $}	0.0pt
0.66754	0.2872727272727273	h	{$ $}	0.0pt
0.65178	0.351575456053068	h	{$ $}	0.0pt
0.78064	0.3601532567049808	h	{$ $}	0.0pt
0.82858	0.4229346485819975	h	{$ $}	0.0pt
0.75292	0.2004357298474945	c	{$ $}	0.0pt
0.78456	0.226606538895152	c	{$ $}	0.0pt
0.86398	0.5989583333333334	h	{\fontsize{4pt}{4pt}$\star$}	0.0pt
0.75466	0.26513	f	{$\cdot$}	0.3pt
0.7607	0.4868131868131868	l	{\fontsize{4pt}{4pt}$\star$}	0.0pt
0.68364	0.1952054794520548	l	{\fontsize{4pt}{4pt}$\star$}	0.0pt
0.82038	0.6381514257620453	l	{\fontsize{4pt}{4pt}$\star$}	0.0pt
0.63376	0.2769679300291545	l	{\fontsize{4pt}{4pt}$\star$}	0.0pt
0.76764	0.6461001164144353	l	{\fontsize{4pt}{4pt}$\star$}	0.0pt
0.81632	0.2597560975609756	d	{$ $}	0.0pt
0.82414	0.2792792792792793	d	{$ $}	0.0pt
0.81958	0.268722	d	{$ $}	0.0pt
0.80532	0.302817	d	{$ $}	0.0pt
0.81516	0.3014705882352942	d	{$ $}	0.0pt
0.82388	0.182955	d	{$ $}	0.0pt
0.8212	0.2412231030577577	d	{$ $}	0.0pt
0.81772	0.2473246135552913	d	{$ $}	0.0pt
0.80308	0.19515	d	{$ $}	0.0pt
0.81678	0.3048368953880765	d	{$ $}	0.0pt
0.8229	0.1581986143187067	d	{$ $}	0.0pt
0.8229	0.1647901740020471	d	{$ $}	0.0pt
0.80546	0.165929203539823	d	{$ $}	0.0pt
0.78838	0.1334723670490094	d	{$ $}	0.0pt
0.8159	0.1812080536912752	d	{$ $}	0.0pt
0.79482	0.232508073196986	k	{$ $}	0.0pt
0.77104	0.1970884658454647	k	{$ $}	0.0pt
0.76626	0.2276119402985075	k	{$ $}	0.0pt
0.76492	0.3354037267080745	k	{$ $}	0.0pt
0.74408	0.4359313077939234	k	{$ $}	0.0pt
0.79228	0.2463768115942029	a	{$ $}	0.0pt
0.7725	0.6229665071770334	l	{\fontsize{4pt}{4pt}$\star$}	0.0pt
0.76902	0.2889733840304182	h	{$ $}	0.0pt
0.80462	0.6273885350318471	l	{\fontsize{4pt}{4pt}$\star$}	0.0pt
0.76526	0.2840158520475561	k	{$ $}	0.0pt
0.7213	0.6368653421633554	l	{\fontsize{4pt}{4pt}$\star$}	0.0pt
0.75898	0.5720122574055158	l	{\fontsize{4pt}{4pt}$\star$}	0.0pt
0.79168	0.6835187057633973	l	{\fontsize{4pt}{4pt}$\star$}	0.0pt
0.72786	0.7202797202797203	g	{$ $}	0.0pt
    };
\end{axis}
\end{tikzpicture}
        \end{subfigure}
        \hspace{-2.5em}
        \begin{subfigure}{0.275\linewidth}
           \centering
           \begin{tikzpicture}[every node/.style={font=\sffamily}]
\sffamily
\scriptsize

\begin{axis}[
            title={Parameters in Mil.\strut},
            title style={yshift=-1em},
            width=1\linewidth,
            minor y tick num=0,
            xtick={0.6,0.7, 0.8, 0.9},
            xticklabels={0.6,0.7, 0.8, 0.9},
            ytick={0, 250, 500,750},
            yticklabels={0, 250, 500,750},
            label style={font=\scriptsize},
            tick label style={font=\scriptsize},
            x label style={at={(axis description cs:0.5,+0.05)},anchor=north},
            y label style={at={(axis description cs:+0.08,.5)},anchor=center},
            ]
    \addplot [
        scatter,
        only marks,
        point meta=explicit symbolic,
        scatter/classes={
            a={mark=square*,fill=cnncolor,draw=cnncolor,mark size=2pt, opacity=0.5, draw opacity=0.7},
            b={mark=*,fill=cnncolor,draw=cnncolor, mark size=2.2pt, opacity=0.5, draw opacity=0.7},
            c={mark=triangle*,fill=cnncolor,draw=cnncolor, mark size=3pt, opacity=0.5, draw opacity=0.7},
            d={mark=pentagon*,fill=cnncolor,draw=cnncolor, mark size=2.5pt, opacity=0.5, draw opacity=0.7},
            e={mark=diamond*,fill=cnncolor,draw=cnncolor, mark size=3pt, opacity=0.5, draw opacity=0.7},
            f={mark=square*,fill=transformercolor,draw=transformercolor,mark size=2pt, opacity=0.5, draw opacity=0.7},
            g={mark=*,fill=transformercolor,draw=transformercolor, mark size=2.2pt, opacity=0.5, draw opacity=0.7},
            h={mark=triangle*,fill=transformercolor,draw=transformercolor, mark size=3pt, opacity=0.5, draw opacity=0.7},
            i={mark=pentagon*,fill=transformercolor,draw=transformercolor, mark size=2.5pt, opacity=0.5, draw opacity=0.7},
            j={mark=diamond*,fill=transformercolor,draw=transformercolor, mark size=3pt, opacity=0.5, draw opacity=0.7},
            k={mark=square*,fill=bcoscolor,draw=bcoscolor,mark size=2pt, opacity=0.5, draw opacity=0.7},
            l={mark=triangle*,fill=vilcolor,draw=vilcolor, mark size=3pt, opacity=0.5, draw opacity=0.7}
        },
        visualization depends on={value \thisrow{symbol} \as \symbol},
        visualization depends on={value \thisrow{centeroffset} \as \centeroffset},
        nodes near coords*={\symbol},
        every node near coord/.append style={anchor=center, yshift=-\centeroffset, opacity=0.5, color=black},
    ] table [meta=label] {
x	y	label	symbol	centeroffset
0.56516	61.1	a	{$ $}	0.0pt
0.69784	6.6	a	{$ $}	0.0pt
0.74216	132.9	a	{$ $}	0.0pt
0.7832	11.7	a	{$ $}	0.0pt
0.78844	68.9	a	{$ $}	0.0pt
0.58182	1.2	a	{$ $}	0.0pt
0.80132	27.2	a	{$ $}	0.0pt
0.80446	55.8	a	{$ $}	0.0pt
0.83246	25.0	a	{$ $}	0.0pt
0.77114	8.0	a	{$ $}	0.0pt
0.7904	22.9	a	{$ $}	0.0pt
0.7187	3.5	a	{$ $}	0.0pt
0.76226	1.4	a	{$ $}	0.0pt
0.82628	88.8	a	{$ $}	0.0pt
0.74046	2.5	a	{$ $}	0.0pt
0.779	5.5	a	{$\cdot$}	0.3pt
0.64526	15.7	a	{$ $}	0.0pt
0.76514	2.2	a	{$ $}	0.0pt
0.84124	5.3	a	{$ $}	0.0pt
0.86828	5.3	e	{\fontsize{4pt}{4pt}$\star$}	0.0pt
0.68404	68.9	b	{$ $}	0.0pt
0.63862	25.6	b	{$ $}	0.0pt
0.84508	25.5	a	{$\cdot$}	0.3pt
0.80874	4.3	a	{$ $}	0.0pt
0.81066	86.6	f	{$ $}	0.0pt
0.8358	28.3	f	{$ $}	0.0pt
0.83162	21.5	a	{$ $}	0.0pt
0.83332	21.5	a	{$\cdot$}	0.3pt
0.84556	54.1	a	{$ $}	0.0pt
0.84772	54.1	a	{$\cdot$}	0.3pt
0.852	118.5	a	{$ $}	0.0pt
0.85822	118.5	a	{$\cdot$}	0.3pt
0.8198	5.7	f	{$ $}	0.0pt
0.8229	5.7	f	{$ $}	0.0pt
0.85044	12.0	f	{$ $}	0.0pt
0.82502	8.8	f	{$ $}	0.0pt
0.82896	47.7	f	{$ $}	0.0pt
0.82588	9.2	f	{$ $}	0.0pt
0.82446	4.8	f	{$ $}	0.0pt
0.836	5.7	f	{$ $}	0.0pt
0.8493	30.9	f	{$ $}	0.0pt
0.81358	22.1	f	{$ $}	0.0pt
0.83068	22.1	f	{$\cdot$}	0.3pt
0.83094	38.8	f	{$ $}	0.0pt
0.84558	38.8	f	{$\cdot$}	0.3pt
0.83786	86.6	f	{$ $}	0.0pt
0.85708	86.6	f	{$\cdot$}	0.3pt
0.84778	304.4	f	{$ $}	0.0pt
0.86976	304.4	f	{$\cdot$}	0.3pt
0.85252	24.2	f	{$ $}	0.0pt
0.84598	28.4	f	{$ $}	0.0pt
0.86942	87.9	f	{$\cdot$}	0.3pt
0.79172	5.4	f	{$ $}	0.0pt
0.80882	5.4	f	{$\cdot$}	0.3pt
0.81502	11.0	f	{$ $}	0.0pt
0.8324	11.0	f	{$\cdot$}	0.3pt
0.83248	21.2	f	{$ $}	0.0pt
0.85084	21.2	f	{$\cdot$}	0.3pt
0.78836	22.1	f	{$ $}	0.0pt
0.85838	22.1	f	{$\cdot$}	0.3pt
0.82516	28.6	a	{$ $}	0.0pt
0.829	28.6	a	{$\cdot$}	0.3pt
0.83634	50.2	a	{$ $}	0.0pt
0.8459	50.2	a	{$\cdot$}	0.3pt
0.84062	88.6	a	{$ $}	0.0pt
0.85812	88.6	a	{$\cdot$}	0.3pt
0.84414	197.8	a	{$ $}	0.0pt
0.86606	197.8	a	{$\cdot$}	0.3pt
0.8522	86.5	h	{$\cdot$}	0.0pt
0.83374	12.3	f	{$ $}	0.0pt
0.8465	28.4	f	{$ $}	0.0pt
0.78744	87.8	g	{$ $}	0.0pt
0.78118	88.6	b	{$ $}	0.0pt
0.76074	88.8	b	{$ $}	0.0pt
0.7743	198.1	b	{$ $}	0.0pt
0.76718	87.1	g	{$ $}	0.0pt
0.81864	15.6	c	{$ $}	0.0pt
0.8203	15.6	c	{$\cdot$}	0.0pt
0.82938	28.6	c	{$ $}	0.0pt
0.83892	28.6	c	{$\cdot$}	0.0pt
0.84876	88.7	c	{$ $}	0.0pt
0.86744	88.7	c	{$\cdot$}	0.0pt
0.8576	198.0	c	{$ $}	0.0pt
0.87264	198.0	c	{$\cdot$}	0.0pt
0.86068	27.9	h	{$ $}	0.0pt
0.88264	5.8	h	{$\cdot$}	0.0pt
0.84088	28.1	f	{$ $}	0.0pt
0.835	11.6	f	{$ $}	0.0pt
0.85602	86.5	h	{$ $}	0.0pt
0.81226	115.1	a	{$ $}	0.0pt
0.79078	25.6	a	{$ $}	0.0pt
0.80314	25.0	e	{\fontsize{4pt}{4pt}$\star$}	0.0pt
0.81138	25.6	e	{\fontsize{4pt}{4pt}$\star$}	0.0pt
0.84278	88.8	e	{\fontsize{4pt}{4pt}$\star$}	0.0pt
0.73286	11.7	e	{\fontsize{4pt}{4pt}$\star$}	0.0pt
0.83742	86.6	h	{$ $}	0.0pt
0.66754	86.6	h	{$ $}	0.0pt
0.65178	51.5	h	{$ $}	0.0pt
0.78064	87.3	h	{$ $}	0.0pt
0.82858	86.6	h	{$ $}	0.0pt
0.75292	25.6	c	{$ $}	0.0pt
0.78456	25.6	c	{$ $}	0.0pt
0.86398	309.5	h	{\fontsize{4pt}{4pt}$\star$}	0.0pt
0.75466	5.7	f	{$\cdot$}	0.3pt
0.7607	203.2	l	{\fontsize{4pt}{4pt}$\star$}	0.0pt
0.68364	102.0	l	{\fontsize{4pt}{4pt}$\star$}	0.0pt
0.82038	877.4	l	{\fontsize{4pt}{4pt}$\star$}	0.0pt
0.63376	119.7	l	{\fontsize{4pt}{4pt}$\star$}	0.0pt
0.76764	53.8	l	{\fontsize{4pt}{4pt}$\star$}	0.0pt
0.81632	5.3	d	{$ $}	0.0pt
0.82414	20.6	d	{$ $}	0.0pt
0.81958	44.5	d	{$ $}	0.0pt
0.80532	25.0	d	{$ $}	0.0pt
0.81516	115.1	d	{$ $}	0.0pt
0.82388	5.3	d	{$ $}	0.0pt
0.8212	20.6	d	{$ $}	0.0pt
0.81772	44.5	d	{$ $}	0.0pt
0.80308	25.0	d	{$ $}	0.0pt
0.81678	115.1	d	{$ $}	0.0pt
0.8229	5.3	d	{$ $}	0.0pt
0.8229	20.6	d	{$ $}	0.0pt
0.80546	44.5	d	{$ $}	0.0pt
0.78838	25.0	d	{$ $}	0.0pt
0.8159	115.1	d	{$ $}	0.0pt
0.79482	88.5	k	{$ $}	0.0pt
0.77104	28.5	k	{$ $}	0.0pt
0.76626	7.9	k	{$ $}	0.0pt
0.76492	60.1	k	{$ $}	0.0pt
0.74408	86.9	k	{$ $}	0.0pt
0.79228	20.6	a	{$ $}	0.0pt
0.7725	149.8	l	{\fontsize{4pt}{4pt}$\star$}	0.0pt
0.76902	23.2	h	{$ $}	0.0pt
0.80462	652.2	l	{\fontsize{4pt}{4pt}$\star$}	0.0pt
0.76526	44.5	k	{$ $}	0.0pt
0.7213	149.6	l	{\fontsize{4pt}{4pt}$\star$}	0.0pt
0.75898	351.8	l	{\fontsize{4pt}{4pt}$\star$}	0.0pt
0.79168	427.6	l	{\fontsize{4pt}{4pt}$\star$}	0.0pt
0.72786	22.8	g	{$ $}	0.0pt
    };
\end{axis}
\end{tikzpicture}
        \end{subfigure}
        
    \caption{\textit{Different quality dimensions (y axis) \vsiccv accuracy (x axis).} To reduce clutter in the plots, we only plot representative models instead of our full model zoo; please refer to the project page for interactive plots with all models. To emphasize the effect of different training strategies and model architectures, we group models visually: the training dataset size is marked by symbols within each marker (no symbol for ImageNet-1k, dot ($\cdot$) for ImageNet-21k, star ($\star$) for large-scale datasets); different training strategies by shapes (standard supervised training as squares~{\protect\squared{gray!60}}, adversarial training as circles~{\protect\circled{gray!60}}, self-supervised \mbox{(pre-)training} as triangles~{\protect\triangled{gray!60}}, semi-supervised training as diamonds~{\protect\diamonded{gray!60}}, A[1,2,3] training as pentagons~{\protect\pentagoned{gray!60}}); and different architectures by color (blue~\colorindicator{cnncolor} for CNNs, orange~\colorindicator{transformercolor} for Transformers, green~\colorindicator{bcoscolor} for B-cos models, and yellow~\colorindicator{vilcolor} for vision-language (ViL) models). 
    }
    \label{fig:scatter_dimensions_acc}
\end{figure*}
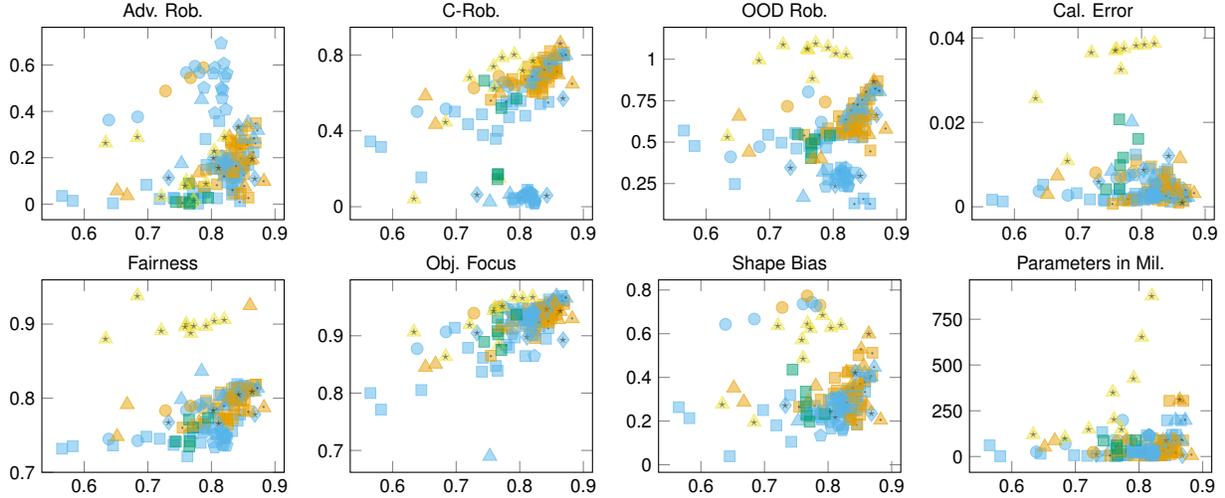

\subsection{Different training strategies}

\paragraph{Training dataset size.} We compare models trained on ImageNet-1k~\citep{Russakovsky:2015:ILS} to those trained on the larger ImageNet-21k~\citep{Deng:2009:ILS} dataset in \cref{fig:scatter_dimensions_acc} (no symbol \vsiccv \mbox{dot ($\cdot$)}) and \cref{tab:comparisons} \tablesetupcnndataset\ and \tablesetuptransdataset. %
Training on a larger dataset improves nearly all quality dimensions for CNNs and Transformers, particularly accuracy, C-robustness, and calibration -- likely by promoting more general and less overfitted features.
Interestingly, with a larger training dataset, the adversarial robustness of Transformers decreases.%
\nopagebreak
\answerbox{Training on a larger dataset improves most of the considered quality dimensions.}

\myparagraph{Adversarial training.} We compare adversarially trained models against the corresponding standard supervised models in \cref{fig:scatter_dimensions_acc} (circles~\circled{gray!60} \vsiccv squares~\squared{gray!60}) and \cref{tab:comparisons} \tablesetupat. Adversarial training (AT) improves shape bias~\citep{Engstrom:2019:ARA, Wang:2020:HFC,Geirhos:2021:PSI}, OOD robustness~\citep{Engstrom:2019:ARA}, and, expectedly, adversarial robustness. Accuracy and class balance significantly worsen with AT. %
The latter is in line with results of~\citet{Benz:2020:RMB,Xu:2021:TBR}. Interestingly, we observe a trend of increasing calibration error with AT, which extends findings in \citep{Grabinski:2022:RBA} that found evidence that adversarially trained ResNets~\citep{He:2016:DRL} exhibit improved calibration errors. 
\nopagebreak
\answerbox{AT improves adversarial/OOD robustness and shape bias. It impairs accuracy and class balance.}

\paragraph{Self-supervised training.} Self-supervised learning eliminates the need for dataset annotations and thus allows for training on significantly larger datasets. We compare models initialized with weights obtained through self-supervised learning to standard supervised models in \cref{fig:scatter_dimensions_acc} (triangles~\triangled{gray!60} \vsiccv squares~\squared{gray!60}) and \cref{tab:comparisons} \tablesetupselfsllp\ and \tablesetupselfsletoe. 
We consider self-supervised models in two standard transfer settings: \textit{(i)} models where only the final classification layer is trained (linear probing, LP) and \textit{(ii)} models that are fully fine-tuned on ImageNet-1k (E2E). We analyze how each approach affects the different quality dimensions.
LP models (\cref{tab:comparisons} \tablesetupselfsllp) generally underperform compared to supervised models, except in OOD robustness, calibration, and shape bias -- likely due to the larger gap between training and testing distributions.
The reduced object focus is notable, as \citet{Caron:2021:EPS} found that self-supervised Transformers produce attention maps that closely align with objects. This suggests that attention maps may not serve as reliable explanations, a finding consistent with \citep{Hesse:2023:SVD, Hesse:2024:IDSDS}.
The slight parameter difference stems from DINOv2~\citep{Oquab:2024:DIN} using a $14 \times 14$ patch size instead of the original $16 \times 16$.

On the other hand, fine-tuning (E2E) self-supervised models (\cref{tab:comparisons} \tablesetupselfsletoe) improves most quality dimensions (except calibration) -- 
probably due to larger pre-training datasets typically used for self-supervised learning and a smaller domain gap than LP.
The improvement in class balance is particularly surprising, as we expected the larger training datasets used for self-supervised training to have much stronger class imbalances than the evenly distributed ImageNet dataset. We hypothesize that the class balance is nonetheless improved because the self-supervised models are pre-trained without any class information, resulting in features less tailored to specific classes. As a result, these features yield a more balanced distribution of class accuracies and class confidences. 
\nopagebreak
\answerbox{Self-supervised models with linear probing perform worse than supervised models in most quality dimensions. Fine-tuning self-supervised models in an end-to-end fashion improves most quality dimensions.}

\begin{table*}[t!]
  \centering
  \tiny
    \caption{\textit{Average quality dimensions for models with different configurations.} We evaluate various setups, focusing on different training strategies (a--g) and different architectural choices (h--j). In each configuration, we report the average score for each quality dimension across the models associated with that configuration.
  As different models are available for different setups, each setup considers a distinct selection of models being compared. As a result, both the models and their total number (indicated by the number beside each setup) vary across setups to maintain a fair basis for comparison.
  The number of asterisks represents the statistical significance of differences in the average scores of a quality dimension across configurations within each setup:~$^{***}$~for~$p < 0.05$,~$^{**}$~for $p < 0.1$, and $^{*}$ for $p < 0.2$, based on $t$-test results.
  }
  \begin{tabularx}{\linewidth}{@{}p{0.1cm}p{0.1cm}XS[table-format=1.2, table-space-text-post ={*}, table-space-text-pre ={+}]S[table-format=1.2, table-space-text-post ={*}, table-space-text-pre ={+}]S[table-format=1.2, table-space-text-post ={*}, table-space-text-pre ={+}]S[table-format=1.2, table-space-text-post ={*}, table-space-text-pre ={+}]S[table-format=1.4, table-space-text-post ={*}, table-space-text-pre ={+}]S[table-format=1.2, table-space-text-post ={*}, table-space-text-pre ={+}]S[table-format=1.2, table-space-text-post ={*}, table-space-text-pre ={+}]S[table-format=1.2, table-space-text-post ={*}, table-space-text-pre ={+}]S[table-format=2, table-space-text-pre ={+}, table-space-text-post ={***}]@{}}%
    \toprule
    \textbf{Setup} & &\textbf{Configuration}  & {\tablearrowcenter\textbf{Acc.}$\uparrow$} & {\tablearrowcenter\makecell{\textbf{Adv.} \\ \textbf{Rob.}}$\uparrow$} & {\tablearrowcenter\textbf{C-Rob.}$\uparrow$} & {\tablearrowcenter\makecell{\textbf{OOD} \\ \textbf{Rob.}}$\uparrow$} & {\tablearrowcenter\makecell{\textbf{Cal.} \\ \textbf{Error}}$\downarrow$} & {\tablearrowcenter\makecell{\textbf{Class} \\ \textbf{Balance}}$\uparrow$} & {\tablearrowcenter\makecell{\textbf{Obj.} \\ \textbf{Focus}}$\uparrow$} & {\tablearrowcenter\makecell{\textbf{Shape} \\ \textbf{Bias}}$\uparrow$} & {\hspace{-0.8em}\tablearrowcenter\makecell{\textbf{Params.} \\ \textbf{in Mil.}}$\downarrow$} \\
\midrule
\tablesetupcnndataset\ & 14 & CNNs (IN-1k)& 0.82& 0.14& 0.67& 0.60& 0.0048& 0.80& 0.93& 0.29& 69\\
&& CNNs (IN-21k)& \bfseries 0.84$^{*}$ & \bfseries 0.15& \bfseries 0.71& 0.60& \bfseries 0.0026$^{***}$ & 0.80& \bfseries 0.95& \bfseries 0.33$^{*}$ & 69\\
\midrule
\tablesetuptransdataset\ & 16 & Transformers (IN-1k)& 0.81& \bfseries 0.21& 0.69& 0.61& 0.0046& 0.79& 0.94& 0.39& 85\\
&& Transformers (IN-21k)& \bfseries 0.84$^{***}$ & 0.17$^{*}$ & \bfseries 0.75$^{**}$ & \bfseries 0.69$^{**}$ & \bfseries 0.0027$^{***}$ & \bfseries 0.80$^{**}$ & \bfseries 0.95& \bfseries 0.42& 85\\
\midrule
\tablesetupat\ & 11 & Supervised models& \bfseries 0.82& 0.12& \bfseries 0.66& 0.60& \bfseries 0.0056& \bfseries 0.80& \bfseries 0.94& 0.29& 86\\
&& Adversarially trained models& 0.74$^{***}$ & \bfseries 0.52$^{***}$ & 0.62& \bfseries 0.63& 0.0068$^{*}$ & 0.78$^{**}$ & 0.93& \bfseries 0.72$^{***}$ & 86\\
\midrule
\tablesetupselfsllp\ & 13 & Supervised models& \bfseries 0.81& \bfseries 0.20& \bfseries 0.66& 0.58& 0.0047& \bfseries 0.81& \bfseries 0.93& 0.34& \bfseries 89\\
&& Self-supervised models (LP)& 0.75$^{*}$ & 0.10$^{***}$ & 0.61& \bfseries 0.59& \bfseries 0.0029$^{**}$ & 0.78$^{**}$ & 0.88$^{**}$ & \bfseries 0.40$^{*}$ & 91\\
\midrule
\tablesetupselfsletoe\ & 25 & Supervised models& 0.81& 0.16& 0.67& 0.58& \bfseries 0.0034& 0.79& 0.93& 0.38& \bfseries 94\\
&& Self-supervised models (E2E)& \bfseries 0.84$^{***}$ & \bfseries 0.24$^{***}$ & \bfseries 0.73& \bfseries 0.73$^{***}$ & 0.0045& \bfseries 0.81$^{***}$ & \bfseries 0.95$^{***}$ & \bfseries 0.39$^{*}$ & 95\\
\midrule
\tablesetupsemisl\ & 13 & Supervised models& 0.80& 0.13& \bfseries 0.58& 0.59& \bfseries 0.0048& 0.78& 0.92& 0.24& 28\\
&& Semi-supervised models& \bfseries 0.82$^{*}$ & \bfseries 0.20$^{***}$ & 0.45$^{*}$ & 0.59& 0.0059& \bfseries 0.80$^{*}$ & \bfseries 0.93& \bfseries 0.29$^{***}$ & 28\\
\midrule
\tablesetuplong\ & 19 & Supervised models& 0.79& 0.12& \bfseries 0.50& \bfseries 0.52& 0.0044& \bfseries 0.78& 0.92& 0.25& 39\\
& & A1 supervised models (600 epochs)& \bfseries 0.80& \bfseries 0.47$^{***}$ & 0.06$^{***}$ & 0.31$^{***}$ & 0.0030$^{*}$ & 0.75$^{***}$ & 0.92& \bfseries 0.27$^{**}$ & 39\\
& & A2 supervised models (300 epochs)& \bfseries 0.80& 0.41$^{***}$ & 0.06$^{***}$ & 0.30$^{***}$ & \bfseries 0.0026$^{***}$ & 0.75$^{***}$ & \bfseries 0.93$^{***}$ & 0.25& 39\\
& & A3 supervised models (100 epochs)& 0.78& 0.32$^{***}$ & 0.04$^{***}$ & 0.27$^{***}$ & 0.0048& 0.76$^{***}$ & 0.91& 0.17$^{***}$ & 39\\
\midrule
\tablesetupcnnvstrans\ & 46 & CNNs& 0.81& 0.11& 0.62& 0.54& 0.0048& 0.79& 0.92& 0.29& 40\\
& & Transformers& 0.81& \bfseries 0.20& \bfseries 0.69$^{***}$ & \bfseries 0.62$^{***}$ & \bfseries 0.0046& \bfseries 0.80$^{*}$ & \bfseries 0.93$^{***}$ & \bfseries 0.32& 40\\
\midrule
\tablesetupbcos\ & 12 & Standard models& \bfseries 0.77& \bfseries 0.07& \bfseries 0.58& \bfseries 0.56& \bfseries 0.0033& \bfseries 0.77& \bfseries 0.93& 0.27& 37\\
&& B-cos models& 0.75$^{*}$ & 0.02$^{***}$ & 0.26$^{***}$ & 0.46$^{***}$ & 0.0115$^{***}$ & 0.75$^{*}$ & 0.90$^{***}$ & 0.27& \bfseries 36\\
\midrule
\tablesetupzeroshot\ & 24 & Standard models& \bfseries 0.81& \bfseries 0.18& \bfseries 0.62& 0.56& \bfseries 0.0044& 0.79& 0.93& 0.35& \bfseries 152\\
&& Vision-language models& 0.74$^{***}$ & 0.10$^{*}$ & 0.60& \bfseries 1.00$^{***}$ & 0.0337$^{***}$ & \bfseries 0.90$^{***}$ & 0.93& \bfseries 0.56$^{***}$ & 275$^{***}$ \\

    \bottomrule
  \end{tabularx}

  \label{tab:comparisons}
\end{table*}

\myparagraph{Semi-supervised training.} We also measure how semi-supervised training~\citep{Xie:2020:STW, Yalniz:2019:BSS}, \ie, training on a combination of labeled and unlabeled data, compares to supervised training in \cref{fig:scatter_dimensions_acc} (diamonds~\diamonded{gray!60} \vsiccv squares~\squared{gray!60}) and \cref{tab:comparisons} \tablesetupsemisl. Among the dimensions with statistically significant changes, semi-supervised training has similar effects as self-supervised training with E2E fine-tuning, probably also due to the combination of a large-scale training dataset and relatively close training and testing domains. Only C-robustness is negatively affected statistically significantly.
\nopagebreak
\answerbox{Semi-supervised training improves accuracy, adversarial robustness, class balance, and shape bias. Only C-robustness is clearly impaired.}

\myparagraph{A[1,2,3] training.} \citet{Wightman:2021:RSB} introduce several training strategies, termed A1, A2, and A3, which incorporate best practices for training DNNs -- \eg, multi-label classification objectives, data augmentation techniques, and the use of advanced optimizers. Most importantly, the three strategies vary in their training duration: A1 is trained for 600, A2 for 300, and A3 for 100 epochs. We compare the training strategies to standard supervised models in \cref{fig:scatter_dimensions_acc} (pentagons~\pentagoned{gray!60} \vsiccv squares~\squared{gray!60}) and \cref{tab:comparisons} \tablesetuplong. While some training strategies of the standard supervised models might overlap with the A[1,2,3] training, the long training of A1 is not utilized in any of the standard models. The accuracy is slightly increasing for the setups with increased training times (A[1,2]; statistically insignificant). Interestingly, adversarial robustness significantly improves with the A[1,2,3] training, while C-robustness and OOD robustness decrease. We believe the improved training enhances adversarial robustness by expanding the distance between decision boundaries and data points, but reduces generalizability by encouraging ``overfitting'' to the training distribution.
Calibration error decreases for the setups with increased training times (A[1,2]), which extends findings of \citet{Minderer:2021:RCM} that showed that calibration error increases with longer training when measured only on BiT models~\citep{Kolesnikov:2020:BIT}. Class balance is reduced for all the setups; object focus remains fairly stable, and the shape bias increases with longer training, confirming results of \citet{Hermann:2020:TOP}. 
\nopagebreak
\answerbox{Adversarial robustness, calibration, and shape bias improve with longer training times. C/OOD-robustness and class balance are impeded.} %

\subsection{Different model designs}

Now that we have covered various training strategies and their effect on different quality dimensions, we analyze the effect of specific architectural choices.

\myparagraph{Is the time of CNNs over?} 
We compare models based on convolutions (CNNs) and attention (Transformers) in \cref{fig:scatter_dimensions_acc} (blue~\colorindicator{cnncolor} \vsiccv orange~\colorindicator{transformercolor}) and \cref{tab:comparisons} \tablesetupcnnvstrans. Since Vision Transformers~\citep{Dosovitskiy:2021:IWW} were introduced only in 2020, we exclude CNNs proposed before 2020 and compare only newer CNN architectures with Transformers for a fairer evaluation. To further improve the fairness of our comparison, we make sure that we have an equal number of CNNs and Transformers from different setups (\eg, adversarial training) and only compare them when they have a similar number of parameters (within a 1-million difference). 
Despite our efforts to ensure a balanced comparison, this setup gives us less control over certain variables than our other experiments.
Therefore, these results should be interpreted with caution. Remarkably, CNNs and Transformers perform equally in accuracy. However, Transformers outperform CNNs in all the other quality dimensions. %
Our results on robustness nicely complement those of \citet{Bai:2021:TMR}, who compared the robustness of CNNs and Transformers but considered only ResNet50~\citep{He:2016:DRL} and DeiT-S/16~\citep{Touvron:2021:TDE}.
\nopagebreak
\answerbox{Transformers consistently outperform CNNs across almost all quality dimensions.}

\myparagraph{B-cos transform.} Initially introduced to improve interpretability, the B-cos transform~\citep{Boehle:2022:BCN} can substitute the linear transformations in a DNN. It encourages the weights to align with the input and potentially affects the model beyond the improved interpretability. We thus analyze B-cos models in \cref{fig:scatter_dimensions_acc} (green~\colorindicator{bcoscolor}) and compare them to the corresponding standard models in \cref{tab:comparisons} \tablesetupbcos. Besides shape bias and the number of parameters, all considered quality dimensions drop significantly when using the B-cos transform. A potential reason for this is the inductive bias of weight-input alignment, limiting the model's expressiveness.
\nopagebreak
\answerbox{The B-cos transform negatively affects most of the considered quality dimensions.}

\myparagraph{Vision-language (ViL) models.} %
With ViL models becoming increasingly relevant, we study their performance across the considered quality dimensions in \cref{fig:scatter_dimensions_acc} (yellow~\colorindicator{vilcolor}) and compare them to their corresponding backbones trained in a supervised fashion in \cref{tab:comparisons} \tablesetupzeroshot. 
Please note that \citet{Tu:2023:CLR} conducted a similar study, however, focusing exclusively on CLIP models~\citep{Radford:2021:LTV} and covering a slightly different set of quality dimensions.
Since ViL models perform zero-shot classification by mapping the 1000 ImageNet-1k class labels into their feature space and then predicting the class label closest to the feature embedding of the given image, %
their accuracy is notably lower than that of the supervised models. 
Also, they contain significantly more parameters due to the additional language encoder. %
They exhibit decreased adversarial robustness and C-robustness while strongly improving OOD robustness~\citep{Radford:2021:LTV}. At first glance, one might attribute the improved OOD robustness to the models being trained on significantly larger datasets that include domains similar to those in the OOD datasets~\citep{Liu:2023:TTA}. While this is certainly a factor~\citep{Mayilvahanan:2024:ISF}, a closer look reveals that ViL models still outperform other models trained on similarly large datasets (see \cref{appendix:sec:ood}), suggesting that they offer advantages beyond just dataset size. While \citet{Minderer:2021:RCM} found that CLIP is fairly well calibrated when trained on WebImageText (WIT)~\citep{Radford:2021:LTV}, \citet{Tu:2023:CLR} found that CLIP calibration can decrease when trained on other datasets. We extend their finding by observing that other ViL models also exhibit significantly worse calibration than standard models. Class balance and shape bias improve by a large margin -- the former probably for similar reasons as for self-supervised models.
\nopagebreak
\answerbox{%
ViL models excel in OOD robustness, class balance, and shape bias. However, they fall behind in accuracy (zero-shot), calibration, and parameters.}

\section{Relationships between quality dimensions}
\label{sec:experiments_correlations}

\begin{figure}%
\centering
  \begin{tikzpicture}[every node/.style={font=\sffamily}]
\sffamily

\scriptsize

\def\yticklabeloffset{-0.55};
\def\xticklabeloffset{-0.6};
\def\xticklabeloffsetx{-0.1};
\def\xticklabelrotation{60};

\begin{axis}[%
    axis equal image, %
    width=7cm,
    axis line style = {line width=.5pt,draw=gray!50},
    scatter, %
    colormap name=correlation_cm, %
    colorbar, %
    point meta min=-1,
    point meta max=1,
    clip=false,
    grid=minor, %
    minor grid style={line width=.5pt,draw=gray!50},
    minor tick num=1, %
    minor tick length=0pt, 
    tickwidth=0pt, %
    ticks=none, %
    try min ticks=10, %
    y dir=reverse, %
    colorbar style={
        at={(1.05,0.5)},anchor=west,width=0.25cm,
        ytick={-1,-0.5,0,0.5,1},
        yticklabels={-1, -0.5, 0, 0.5, 1},%
    },
    xticklabel pos=right, %
    enlargelimits={abs=0.5}, %
    scatter/@pre marker code/.append code={%
      \pgfplotstransformcoordinatex{sqrt(abs(\pgfplotspointmeta))}%
      \scope[mark size=\pgfplotsunitxlength*\pgfmathresult/3.7 + \pgfplotsunitxlength*20.7/2, fill=mapped color]
    },
    scatter/@post marker code/.append code={%
      \endscope%
    }
    ]

\addplot +[
    point meta=explicit, %
    only marks, %
    every node near coord/.append style={font=\small, color=white,anchor=center},
    visualization depends on={value \thisrow{label} \as \Label}, %
    visualization depends on={value \thisrow{size} \as \Size}, %
    ] table [
    x expr={int(mod(\coordindex+0.01,9))}, %
    y expr={int((\coordindex+0.01)/9))},
    meta=value,
] {
X   Y   value label size
0 0 1.0 $+$ 0.1em
0 0 0 $$ 0.1em
0 0 0 $$ 0.1em
0 0 0 $$ 0.1em
0 0 0 $$ 0.1em
0 0 0 $$ 0.1em
0 0 0 $$ 0.1em
0 0 0 $$ 0.1em
0 0 0 $$ 0.1em
 
0 0 0.4399 $+$ 0.1em
0 0 1.0 $+$ 0.1em
0 0 0 $$ 0.1em
0 0 0 $$ 0.1em
0 0 0 $$ 0.1em
0 0 0 $$ 0.1em
0 0 0 $$ 0.1em
0 0 0 $$ 0.1em
0 0 0 $$ 0.1em
 
0 0 0.6239 $+$ 0.1em
0 0 0.0116 $+$ 0.1em
0 0 1.0 $+$ 0.1em
0 0 0 $$ 0.1em
0 0 0 $$ 0.1em
0 0 0 $$ 0.1em
0 0 0 $$ 0.1em
0 0 0 $$ 0.1em
0 0 0 $$ 0.1em
 
0 0 0.3523 $+$ 0.1em
0 0 -0.0807 $-$ 0.1em
0 0 0.7957 $+$ 0.1em
0 0 1.0 $+$ 0.1em
0 0 0 $$ 0.1em
0 0 0 $$ 0.1em
0 0 0 $$ 0.1em
0 0 0 $$ 0.1em
0 0 0 $$ 0.1em
 
0 0 -0.1175 $-$ 0.1em
0 0 0.066 $+$ 0.1em
0 0 0.0779 $+$ 0.1em
0 0 0.2519 $+$ 0.1em
0 0 1.0 $+$ 0.1em
0 0 0 $$ 0.1em
0 0 0 $$ 0.1em
0 0 0 $$ 0.1em
0 0 0 $$ 0.1em
 
0 0 0.5338 $+$ 0.1em
0 0 0.0821 $+$ 0.1em
0 0 0.7349 $+$ 0.1em
0 0 0.7625 $+$ 0.1em
0 0 0.4903 $+$ 0.1em
0 0 1.0 $+$ 0.1em
0 0 0 $$ 0.1em
0 0 0 $$ 0.1em
0 0 0 $$ 0.1em
 
0 0 0.7169 $+$ 0.1em
0 0 0.4489 $+$ 0.1em
0 0 0.6397 $+$ 0.1em
0 0 0.4709 $+$ 0.1em
0 0 0.0892 $+$ 0.1em
0 0 0.5691 $+$ 0.1em
0 0 1.0 $+$ 0.1em
0 0 0 $$ 0.1em
0 0 0 $$ 0.1em
 
0 0 0.2612 $+$ 0.1em
0 0 0.1673 $+$ 0.1em
0 0 0.6071 $+$ 0.1em
0 0 0.623 $+$ 0.1em
0 0 0.2409 $+$ 0.1em
0 0 0.5399 $+$ 0.1em
0 0 0.5213 $+$ 0.1em
0 0 1.0 $+$ 0.1em
0 0 0 $$ 0.1em
 
0 0 0.3119 $+$ 0.1em
0 0 0.1781 $+$ 0.1em
0 0 0.4356 $+$ 0.1em
0 0 0.4339 $+$ 0.1em
0 0 0.108 $+$ 0.1em
0 0 0.4748 $+$ 0.1em
0 0 0.4821 $+$ 0.1em
0 0 0.5299 $+$ 0.1em
0 0 1.0 $+$ 0.1em
 
};

\addplot[
    mark=x,only marks, mark size=6pt, thick, gray!50,
    point meta =explicit symbolic,
    nodes near coords,
]
table[x=x,y=y, meta=label]{
    x   y   label
    1 2 {} 
    1 3 {} 
    1 4 {} 
    2 4 {} 
    1 5 {} 
    4 6 {} 
    4 8 {}

};

\node[anchor=east] at (axis cs:0+\yticklabeloffset,0) {Accuracy\strut};
\node[anchor=east] at (axis cs:0+\yticklabeloffset,1) {Adv. Rob. \strut};
\node[anchor=east] at (axis cs:0+\yticklabeloffset,2) {C-Rob.\strut};
\node[anchor=east] at (axis cs:0+\yticklabeloffset,3) {OOD Rob.\strut};
\node[anchor=east] at (axis cs:0+\yticklabeloffset,4) {Cal. Error\strut};
\node[anchor=east] at (axis cs:0+\yticklabeloffset,5) {Class Balance\strut};
\node[anchor=east] at (axis cs:0+\yticklabeloffset,6) {Obj. Focus\strut};
\node[anchor=east] at (axis cs:0+\yticklabeloffset,7) {Shape Bias\strut};
\node[anchor=east] at (axis cs:0+\yticklabeloffset,8) {Parameters\strut};

\node[anchor=east] at (axis cs:0+\yticklabeloffset,11.3) {\strut};

\node[anchor=west, rotate=\xticklabelrotation] at (axis cs:0+\xticklabeloffsetx,0+\xticklabeloffset) {Accuracy\strut};
\node[anchor=west, rotate=\xticklabelrotation] at (axis cs:1+\xticklabeloffsetx,0+\xticklabeloffset) {Adv. Rob. \strut};
\node[anchor=west, rotate=\xticklabelrotation] at (axis cs:2+\xticklabeloffsetx,0+\xticklabeloffset) {C-Rob.\strut};
\node[anchor=west, rotate=\xticklabelrotation] at (axis cs:3+\xticklabeloffsetx,0+\xticklabeloffset) {OOD Rob.\strut};
\node[anchor=west, rotate=\xticklabelrotation] at (axis cs:4+\xticklabeloffsetx,0+\xticklabeloffset) {Cal. Error\strut};
\node[anchor=west, rotate=\xticklabelrotation] at (axis cs:5+\xticklabeloffsetx,0+\xticklabeloffset) {Class Balance\strut};
\node[anchor=west, rotate=\xticklabelrotation] at (axis cs:6+\xticklabeloffsetx,0+\xticklabeloffset) {Obj. Focus\strut};
\node[anchor=west, rotate=\xticklabelrotation] at (axis cs:7+\xticklabeloffsetx,0+\xticklabeloffset) {Shape Bias\strut};
\node[anchor=west, rotate=\xticklabelrotation] at (axis cs:8+\xticklabeloffsetx,0+\xticklabeloffset) {Parameters\strut};

\end{axis}

\end{tikzpicture}
\vspace{-4em}
\caption{\textit{Rank correlation matrix for the considered quality dimensions among our full model zoo.} 
All non-crossed-out entries have a $p$-value below 0.05, indicating statistical significance. 
Crossed-out entries correspond to $p$-values above 0.05 and are therefore not statistically significant.
}
    \label{fig:correlation_matrix}
\end{figure}

\paragraph{Comparison to related work.}\label{sec:relationship_confirmation} %
While most previous work is concerned with \textit{improving} quality dimensions, there are also studies examining their \textit{relationships}.
However, not all relationships have been explored, and prior studies used fewer and older models, which can lead to contradictory findings (see \cref{appendix:sec:conflicting_results}).
To address this gap, we investigate the relationship between \textit{numerous} quality dimensions for our \textit{extensive} model zoo, plotting the Spearman's rank correlation matrix for all nine considered quality dimensions across \nrmodels models in \cref{fig:correlation_matrix} --  
please refer to \cref{appendix:sec:subgroups} for correlation matrices of specific model subgroups.
Our analysis confirms that accuracy is positively correlated with OOD robustness~\citep{Miller:2021:AOT}, object focus~\citep{Xiao:2021:NSR}, shape bias \citep{Hermann:2020:TOP}, and with the number of parameters~\citep{Liu:2023:CSR}. 
The number of parameters positively correlates with OOD robustness~\citep{Liu:2023:CSR} and adversarial robustness~\citep{Nakkiran:2019:ARM,Madry:2018:TDL}. Increasing the shape bias improves adversarial robustness~\citep{Jo:2017:MTT,Geirhos:2021:PSI}, accuracy~\citep{Geirhos:2019:ITC}, and OOD robustness~\citep{Geirhos:2019:ITC}.
Accuracy and calibration error exhibit a negative correlation, aligning with \citet{Minderer:2021:RCM}, who found a negative correlation in more recent Transformer models, and contradicting \citet{Guo:2017:OCM}, who observed a positive correlation between these metrics in older backbone models.
Unlike hypothesized by \citet{Tsipras:2019:RMB} and confirming \citet{Yang:2020:ACL}, accuracy and adversarial robustness are positively correlated.
Contrary to \citet{Liu:2023:CSR} and \citet{Jo:2017:MTT}, adversarial robustness is not statistically significantly correlated with C-robustness and OOD robustness. 
While \citet{Grabinski:2022:RBA} found that adversarial training improves calibration, we find no statistically significant correlation between adversarial robustness and calibration. 
\nopagebreak
\answerbox{We provide a bigger picture of related work using our extensive model zoo to validate known quality relationships, resolve conflicting findings, and extend recent findings regarding a link between adversarial and OOD robustness / C-robustness / calibration.%
}

\paragraph{Discovering new relationships.}
While we cannot discuss all findings in \cref{fig:correlation_matrix}, some insights --~to our knowledge~-- have not been reported for backbone models in image classification. 
For example, accuracy and \updated{class balance} are strongly correlated, meaning higher-accuracy models have less discrepancy between the best and worst-performing classes. %
Further, object focus is strongly correlated with all quality dimensions but the calibration error, rendering models with improved object focus an interesting research direction. This may be because models with greater object focus capture fewer surface-level statistical regularities and instead develop higher-level conceptual understanding~\citep{Jo:2017:MTT}.
Interestingly, calibration error is only statistically significantly correlated with OOD robustness, class balance, and shape bias, highlighting the need for dedicated calibration research. %
Lastly, most considered quality dimensions (excluding the number of parameters and calibration) improve together, indicating that developing models that excel in a wide range of quality dimensions is feasible. This observation aligns with \citet{Liu:2023:TTA}, who argue that many desirable properties of trustworthy machine learning are underpinned by shared foundations, suggesting that improvements in one aspect may benefit others.%
\nopagebreak
\answerbox{Accuracy and class balance are strongly correlated, object focus is strongly correlated with most quality dimensions, calibration error is not correlated with most quality dimensions, and there are only a few trade-offs between the considered dimensions.}

\section{Which backbone to use?}
\label{sec:experiments_rankingdetails}

We conclude our analysis by ranking models to provide recommendations on the best model choices.
Ranking models across multiple quality dimensions is a non-trivial task with no one-size-fits-all solution, as user priorities vary depending on specific needs.
Nonetheless, we aim to identify models that perform well across a wide range of dimensions and, thus, require an effective way to summarize the different quality scores with flexible weightings to reflect different user needs. %

Probably the most straightforward way to summarize our results would be to take the average of all quality dimensions. However, given that these dimensions have vastly different ranges and scales, this approach would not treat all dimensions fairly.
Another alternative would be to compute the mean rank: for each quality dimension, the models are ranked, and then the geometric mean of these individual ranks is calculated. However, using ranks has two key limitations. First, ranks are uniformly distributed, whereas the raw scores are not, meaning that the difference in mean rank between two models would not accurately capture the actual difference in their model quality. 
Second, if future studies introduce new models, the set of models will change, altering most of the rankings. As a result, mean ranks would no longer be consistent across different papers.

\begin{table*}[t]
  \centering
  \tiny
      \caption{\textit{QUBA score and quality dimensions for the five top-performing models.} The configuration lists the architecture, training dataset, and training paradigm. \textsuperscript{\textdagger} indicates models trained with knowledge distillation.}
  \setlength{\tabcolsep}{4.5pt}
  \begin{tabularx}{\textwidth}{@{}lp{2.55cm}S[table-format=1.2]S[table-format=1.2]S[table-format=1.2]S[table-format=1.2]S[table-format=1.2]S[table-format=1.4]S[table-format=1.2]S[table-format=1.2]S[table-format=1.2]S[table-format=2]@{}}%
    \toprule
    \makecell[l]{\textbf{Model}} & {\makecell[l]{\textbf{Configuration}}} & {\makecell{\textbf{QUBA} \\ \textbf{Score}}$\uparrow$} & {\textbf{Acc.}$\uparrow$} & {\makecell{\textbf{Adv.} \\ \textbf{Rob.}}$\uparrow$} & {\textbf{C-Rob.}$\uparrow$} & {\makecell{\textbf{OOD} \\ \textbf{Rob.}}$\uparrow$} & {\makecell{\textbf{Cal.} \\ \textbf{Error}}$\downarrow$} & {\makecell{\textbf{Class} \\ \textbf{Balance}}$\uparrow$} & {\makecell{\textbf{Obj.} \\ \textbf{Focus}}$\uparrow$} & {\makecell{\textbf{Shape} \\ \textbf{Bias}}$\uparrow$} & {\makecell{\textbf{Params.} \\ \textbf{in Mil.}}$\downarrow$} \\
\midrule
\makecell[l]{EfficientNet-B6 \\ \citep{Xie:2020:STW}}&\makecell[l]{CNN, JFT-300M \\ \citep{Hinton:2015:DTK}; \\ \citep{Sun:2017:RUE} \\ + IN1k, semi-SL\textsuperscript{\textdagger}} & 0.94& 0.86& 0.25& 0.77& 0.83& 0.0048& 0.82& 0.95& 0.35& \bfseries 43\\[4mm]
\makecell[l]{Hiera-B \\ \citep{Ryali:2023:HHV}}&\makecell[l]{Transformer, IN1k, \\ self-SL (E2E)}& 0.95& 0.85& 0.23& 0.76& 0.76& 0.0130& 0.93& 0.94& 0.34& 51\\[2mm]
\makecell[l]{ConvNeXtV2-B \\ \citep{Woo:2023:CCS}}&\makecell[l]{CNN, IN21k, \\self-SL (E2E)}& 0.96& 0.87& \bfseries 0.28& 0.79& 0.82& \bfseries 0.0023& 0.81& 0.96& 0.40& 88\\[2mm]
\makecell[l]{Hiera-B-Plus \\ \citep{Ryali:2023:HHV}}&\makecell[l]{Transformer, IN1k, \\self-SL (E2E)} & 1.03& 0.85& 0.24& 0.78& 0.74& 0.0130& \bfseries 0.93& 0.95& \bfseries 0.43& 69\\[2mm]
\makecell[l]{EVA02-B/14 \\ \citep{Fang:2023:EVA}}&\makecell[l]{Transformer, IN21k, \\self-SL (E2E)}& \bfseries 1.08& \bfseries 0.88& 0.21& \bfseries 0.81& \bfseries 0.86& 0.0039& 0.83& \bfseries 0.97& 0.34& 87\\

    \bottomrule
  \end{tabularx}

  \label{tab:top5}
\end{table*}
\myparagraph{QUBA score.}
To address these issues, we leverage an intriguing property of our large model zoo: its size makes it representative of a broad range of models, enabling us to estimate a meaningful mean $\mu_i$ and standard deviation $\sigma_i$ for each quality dimension $i$ (we exclude the bottom and top 10\% models to reduce outlier sensitivity).
We then express each model's quality scores $s_i^{\text{model}}$ in terms of how many standard deviations they deviate from the mean.
The final \textit{QUBA score} (\textbf{Q}uality \textbf{U}nderstanding \textbf{B}eyond \textbf{A}ccuracy) for a model is the weighted arithmetic mean of these scores:
\begin{equation}\label{eq:quba}
    \operatorname{QUBA}_{\text{model}} = \left(\frac{1}{\sum^{9}_{i=1} w_i}\right)\sum^{9}_{i=1} w_i \frac{s_i^{\text{model}} - \mu_i}{\sigma_i} .
\end{equation}
\updated{By default, we use a balanced weighting where the three robustness dimensions are weighted} at $w_i=\sfrac{1}{3}$ to prevent them from overshadowing the results and\updated{, similarly,} assign $w_i=\sfrac{1}{2}$ to object focus and shape bias, as both are related to shortcut learning. All other weights are set to 1 \updated{(different weightings are analyzed below)}. Since calibration errors and the number of parameters should be as small as possible, they are multiplied by $-1$ before computing the mean, so that higher values indicate better performance.
Intuitively, the QUBA score reflects how many standard deviations a model deviates from the ``average model'' across the considered dimensions.

Our approach solves both limitations of the mean rank: the distances now have a consistent and meaningful interpretation, and the score can be calculated independently of the considered model set (since the mean and standard deviation for each quality dimension are assumed to be fixed).

\updated{To validate that our model zoo is large enough to produce reliable estimates of the mean and standard deviation for each quality dimension -- and to assess the robustness of the QUBA score -- we randomly sample \num{100} models from our full model zoo and compute the QUBA mean and standard deviation. This process is repeated five times. For each of the five resulting QUBA variants, we rank all \nrmodels models and compute the rank correlation between the resulting rankings. The average rank correlation is very high (\num{0.97}), indicating that the QUBA rankings are stable and not overly dependent on the specific subset of models used.}

\myparagraph{The best models.} We report results for the top five QUBA score models in \cref{tab:top5}. %
Of our \nrmodels models, EVA02-B/14 (IN21k)~\citep{Fang:2023:EET} achieves the best QUBA score. Compared to the other top-performing models, it achieves the highest accuracy, C-robustness, OOD robustness, and object focus. It lags behind in adversarial robustness, calibration, class balance, shape bias, and the number of parameters.
The second-best model, Hiera-B-Plus~\citep{Ryali:2023:HHV}, ranks lower in accuracy and calibration but performs well in the other dimensions, excelling in class balance and shape bias.
In third place, the convolutional model ConvNeXtV2-B (IN21k)~\citep{Woo:2023:CCS} leads in adversarial robustness and calibration while achieving good results in all other dimensions but the parameter count. %
The last two models, Hiera-B~\citep{Ryali:2023:HHV} and EfficientNet-B6~\citep{Xie:2020:STW} have a particularly low parameter count. 
Remarkably, all five models have been trained semi- or self-supervised, making these promising training paradigms for developing well-behaved models. %
Our analysis highlights that even the five top-performing models vary strongly among the quality dimensions, highlighting the need to consider a wide range of quality dimensions simultaneously in the design process of new models. %
\nopagebreak
\answerbox{The models with the highest QUBA scores excel across various quality dimensions, with each model showcasing distinct strengths.} %

\myparagraph{A closer look at popular models.}
\updated{There are many popular models that did not make it into the top five above. We here go over some of the most popular models and briefly discuss their performance according to the considered quality dimensions (see \cref{tab:popular_models}). %
SwinV2-b/12to16~\citep{Liu:2022:STS} is the best \textit{supervised} model, particularly excelling in accuracy and object focus, while having quite a large number of parameters. DINOv2-B-reg (LP)~\citep{Darcet:2024:VTN} exhibits a very good calibration and achieves good results in most other metrics. ViT-b/16-MAE (E2E)~\citep{He:2022:MAA} is in no dimension particularly good or bad. Although ViT-b/16~\citep{Dosovitskiy:2021:IWW} and ResNet50~\citep{He:2016:DRL} are still two of the most popular backbones, they perform quite poorly, with QUBA ranks of 124 and 214, respectively. The ResNet50 has a comparably low number of parameters. CLIP-L/14~\citep{Radford:2021:LTV} suffers particularly in the calibration error and the number of parameters. On the other hand, it exhibits a high shape bias. Based on these findings, we suggest that the vision community should reconsider its selection of canonical backbone models.}

\answerbox{Widely used models such as ResNet50 and ViT underperform in several of the evaluated quality dimensions. This suggests that the vision community should critically reconsider its choice of canonical backbone models.}

\begin{table*}[t]
  \centering
  \tiny
      \caption{\textit{QUBA score and quality dimensions for particularly popular models that did not make it in the top five.} The configuration lists the architecture, the training dataset, and the training paradigm.} 
  \setlength{\tabcolsep}{4.2pt}
  \begin{tabularx}{\textwidth}{@{}p{2.5cm}p{2.3cm}S[table-format=1.2, table-space-text-pre ={-}]S[table-format=1.0]S[table-format=1.2]S[table-format=1.2]S[table-format=1.2]S[table-format=1.2]S[table-format=1.4]S[table-format=1.2]S[table-format=1.2]S[table-format=1.2]S[table-format=2]@{}}%
    \toprule
    \textbf{Model} & \textbf{Configuration}& {\makecell{\textbf{QUBA} \\ \textbf{Score}$\uparrow$
    \\ /\textbf{Rank}$\downarrow$}} & {\textbf{Acc.}$\uparrow$} & {\makecell{\textbf{Adv.} \\ \textbf{Rob.}}$\uparrow$} & {\textbf{C-Rob.}$\uparrow$} & {\makecell{\textbf{OOD} \\ \textbf{Rob.}}$\uparrow$} & {\makecell{\textbf{Cal.} \\ \textbf{Error}}$\downarrow$} & {\makecell{\textbf{Class} \\ \textbf{Balance}}$\uparrow$} & {\makecell{\textbf{Obj.} \\ \textbf{Focus}}$\uparrow$} & {\makecell{\textbf{Shape} \\ \textbf{Bias}}$\uparrow$} & {\makecell{\textbf{Params.} \\ \textbf{in Mil.}}$\downarrow$} \\
\midrule
\makecell[l]{CLIP-L/14 \\ \citep{Radford:2021:LTV}}& {\makecell[l]{ViL, WIT400m \\ \citep{Radford:2021:LTV}, \\ self-SL}} & {\makecell{-0.65/243}}  & {\makecell{0.76}}& {\makecell{\bfseries 0.32}}& {\makecell{0.76}}& {\makecell{\bfseries 1.04}}& {\makecell{0.0110}}& {\makecell{\bfseries 0.89}}& {\makecell{0.94}}& {\makecell{\bfseries 0.60}}& {\makecell{427}}\\\addlinespace[-0.7mm]
\makecell[l]{ResNet50 \\ \citep{He:2016:DRL}}&{\makecell[l]{CNN, IN1k, SL}} & {\makecell{-0.31/214}}  & {\makecell{0.76}}& {\makecell{0.03}}& {\makecell{0.51}}& {\makecell{0.50}}& {\makecell{0.0021}}& {\makecell{0.75}}& {\makecell{0.93}}& {\makecell{0.22}}& {\makecell{\bfseries 25}}\\\addlinespace[1mm]
\makecell[l]{ViT-b/16 \\ \citep{Dosovitskiy:2021:IWW}}&{\makecell[l]{Transformer, \\ IN1k, SL}} & {\makecell{0.20/124}}  & {\makecell{0.81}}& {\makecell{0.18}}& {\makecell{0.66}}& {\makecell{0.56}}& {\makecell{0.0034}}& {\makecell{0.79}}& {\makecell{0.93}}& {\makecell{0.40}}& {\makecell{86}}\\\addlinespace[0.25mm]
\makecell[l]{ViT-b/16-MAE \\ \citep{He:2022:MAA}}&{\makecell[l]{Transformer, \\ IN1k, self-SL (E2E)}} &  {\makecell{0.36/84}}  & {\makecell{0.84}}& {\makecell{0.25}}& {\makecell{0.71}}& {\makecell{0.58}}& {\makecell{0.0049}}& {\makecell{0.80}}& {\makecell{0.95}}& {\makecell{0.36}}& {\makecell{86}}\\\addlinespace[0.4mm]
\makecell[l]{DINOv2-B-reg \\ \citep{Darcet:2024:VTN}}&{\makecell[l]{Transformer, LVD142m \\ \citep{Oquab:2024:DIN}, \\ self-SL (LP)}} & {\makecell{0.74/25}}  & {\makecell{0.85}}& {\makecell{0.12}}& {\makecell{0.79}}& {\makecell{0.79}}& {\makecell{\bfseries 0.0011}}& {\makecell{0.80}}& {\makecell{0.94}}& {\makecell{0.49}}& {\makecell{90}}\\\addlinespace[-0.5mm]
\makecell[l]{SwinV2-b/12to16 \\ \citep{Liu:2022:STS}}&{\makecell[l]{Transformer, \\ IN21k, SL}}&{\makecell{\bfseries{0.90}/\bfseries{8}}} & {\makecell{\bfseries 0.86}}& {\makecell{0.26}}& {\makecell{\bfseries 0.81}}& {\makecell{0.81}}& {\makecell{0.0040}}& {\makecell{0.82}}& {\makecell{\bfseries 0.96}}& {\makecell{0.41}}& {\makecell{87}}\\

    \bottomrule
  \end{tabularx}
  \vspace{-0.5em}
  \label{tab:popular_models}
  \vspace{-0.5em}
\end{table*}

\myparagraph{Different weightings.} As outlined above, different practitioners could have different requirements on their models, depending on the task at hand. To reflect this in our analysis, in \cref{fig:topfive_weighting_barplot}, we plot the five top-performing models when weighting one (group) of the considered quality dimensions twice as much as the other dimensions when computing the weighted mean for the QUBA score. EVA02-B/14 (IN21k)~\citep{Fang:2023:EET} leads in five setups, highlighting its versatility and quality beyond accuracy. 
The top five models remain fairly consistent when emphasizing accuracy, robustness, calibration, and shortcut learning, though their ranking within the top five varies slightly. Interestingly, the Hiera~\citep{Ryali:2023:HHV} model family, self-supervised models only trained on ImageNet-1k, dominates strongly when focusing on class balance. Besides for class balance, the training dataset and architecture are quite heterogeneous for most of the setups. For the training paradigm, semi- and self-supervised learning dominate.

\answerbox{When focusing on specific quality dimensions, the top five models remain fairly stable. %
For class balance, the Hiera model family is dominating.}

\myparagraph{Limitations.}%
Naturally, our work comes with limitations.
First, while we focus on image classification, which certainly is a relevant field, some downstream tasks rely on the evaluated backbone models for other purposes, and there is no guarantee that our findings will directly translate. %
Second, similar to the previous point, our analysis is limited to models trained on ImageNet-1k, and we cannot guarantee that the results generalize to other datasets. However, extending our analysis to another dataset is challenging: assuming an average training time of only \num{10} hours per model, retraining all \nrmodels models on another dataset would require \num{3260} hours ($\sim136$ days) of compute, which is infeasible with our compute resources.
Further, ImageNet-1k remains highly relevant, with numerous impactful papers focusing primarily on it and its variations. %
Third, we acknowledge that there are numerous different protocols to assess different dimensions of DNN quality. While we cannot include all protocols, we aimed for a \textit{(i)} representative selection of \textit{(ii)} established and \textit{(iii)} easy-to-use (requiring no fine-tuning) protocols. 
However, our benchmark can easily be adapted or even extended with other protocols and quality dimensions.
Fourth, to provide a comprehensive bird’s-eye view, we prioritize breadth over depth here. While we briefly discuss various findings \updated{and why they might occur}, this means we cannot provide detailed analyses for specific observations -- \updated{indeed, studying theoretical connections for just two quality dimensions is often the scope of a full paper \citep[\eg,][]{Minderer:2021:RCM,Tsipras:2019:RMB,Xiao:2021:NSR}.} We rather consider this paper as groundwork, paving the way for future research to conduct more fine-grained, in-depth investigations.
Fifth, our chosen evaluation protocols capture only certain aspects of the evaluated dimensions, and different protocols might yield different conclusions.
That said, our benchmark design and its simple applicability allow us to conduct one of the largest studies to date. We thus argue that our design choices are justified and that our analysis %
makes numerous valuable contributions.

\begin{figure}%
\centering
\input{figures/topfive_weighting_barplot_tmlr}
\caption{\textit{Top five QUBA score models under different weightings.} We report the top five models when weighing specific (groups of) quality dimensions twice as strongly. See \cref{fig:scatter_dimensions_acc} for color and marker details.}
    \label{fig:topfive_weighting_barplot}
\end{figure}

\section{Conclusion}
\label{sec_conclusion}

In this work, we provide a bird’s-eye view of nine quality dimensions for ImageNet-1k image classification across \nrmodels vision backbones by conducting one of the largest studies to date.
This broad perspective allows us to examine how various training strategies and model architectures impact these dimensions, finding that larger training datasets and self-supervised pre-training followed by end-to-end fine-tuning enhance almost all measured quality dimensions.
Additionally, we explore the relationships between these quality dimensions, providing novel insights such that object focus is strongly correlated with most of the considered dimensions.
Our analysis is rounded off by ranking models based on our proposed QUBA score, which is possible due to our large model zoo. We highlight that no single model is universally superior and instead provide recommendations on which models excel for specific requirements.
To conclude, we encourage researchers to consider a broad range of quality dimensions together, rather than focusing on individual ones, to foster the development of more well-behaved image classification models. 
Our work facilitates this by offering an easy-to-use benchmark, along with a comprehensive analysis of how design choices influence these quality dimensions, their interrelationships, and how models can be ranked across multiple quality dimensions.

\section*{Acknowledgments}

RH and SR have received funding from the European Research Council (ERC) under the European Union's Horizon 2020 research and innovation programme (grant agreement No.~866008). SSM has been funded by the Deutsche Forschungsgemeinschaft (DFG, German Research Foundation) --~project No.~529680848. Further, SR was supported by the DFG under Germany's Excellence Strategy (EXC 3066/1 ``The Adaptive Mind,'' project No.~533717223). Additionally, SR and SSM have received funding from the DFG under Germany's Excellence Strategy (EXC-3057/1 ``Reasonable Artificial Intelligence'', Project No.~533677015). Moreover, DB is supported by the Konrad Zuse School of Excellence in Learning and Intelligent Systems (\href{https://eliza.school}{ELIZA}) through the DAAD programme Konrad Zuse Schools of Excellence in Artificial Intelligence, sponsored by the Federal Ministry of Education and Research.

{
    \small
    \bibliographystyle{tmlr}
    \bibliography{bibtex/short,bibtex/external,bibtex/local}

\begin{thebibliography}{134}
\providecommand{\natexlab}[1]{#1}
\providecommand{\url}[1]{\texttt{#1}}
\expandafter\ifx\csname urlstyle\endcsname\relax
  \providecommand{\doi}[1]{doi: #1}\else
  \providecommand{\doi}{doi: \begingroup \urlstyle{rm}\Url}\fi

\bibitem[Ali et~al.(2021)Ali, Touvron, Caron, Bojanowski, Douze, Joulin, Laptev, Neverova, Synnaeve, Verbeek, and J{\'{e}}gou]{Ali:2021:XCC}
Alaaeldin Ali, Hugo Touvron, Mathilde Caron, Piotr Bojanowski, Matthijs Douze, Armand Joulin, Ivan Laptev, Natalia Neverova, Gabriel Synnaeve, Jakob Verbeek, and Herv{\'{e}} J{\'{e}}gou.
\newblock {XCiT}: Cross-covariance image transformers.
\newblock In \emph{NeurIPS}, pp.\  20014--20027, 2021.

\bibitem[Andriushchenko et~al.(2020)Andriushchenko, Croce, Flammarion, and Hein]{Andriushchenko:2020:SAA}
Maksym Andriushchenko, Francesco Croce, Nicolas Flammarion, and Matthias Hein.
\newblock Square attack: {A} query-efficient black-box adversarial attack via random search.
\newblock In \emph{ECCV}, volume 12368, pp.\  484--501, 2020.

\bibitem[Bai et~al.(2021)Bai, Mei, Yuille, and Xie]{Bai:2021:TMR}
Yutong Bai, Jieru Mei, Alan~L. Yuille, and Cihang Xie.
\newblock Are transformers more robust than {CNN}s?
\newblock In \emph{NeurIPS}, pp.\  26831--26843, 2021.

\bibitem[Bao et~al.(2022)Bao, Dong, Piao, and Wei]{Bao:2022:BBP}
Hangbo Bao, Li~Dong, Songhao Piao, and Furu Wei.
\newblock {BEiT}: {BERT} pre-training of image transformers.
\newblock In \emph{ICLR}, 2022.

\bibitem[Benz et~al.(2020)Benz, Zhang, Karjauv, and Kweon]{Benz:2020:RMB}
Philipp Benz, Chaoning Zhang, Adil Karjauv, and In~So Kweon.
\newblock Robustness may be at odds with fairness: An empirical study on class-wise accuracy.
\newblock In \emph{NeurIPS}, pp.\  325--342, 2020.

\bibitem[Brendel \& Bethge(2019)Brendel and Bethge]{Brendel:2019:ACB}
Wieland Brendel and Matthias Bethge.
\newblock Approximating {CNN}s with {Bag-of-local-Features} models works surprisingly well on {ImageNet}.
\newblock In \emph{ICLR}, 2019.

\bibitem[Böhle et~al.(2022)Böhle, Fritz, and Schiele]{Boehle:2022:BCN}
Moritz Böhle, Mario Fritz, and Bernt Schiele.
\newblock B-cos networks: {A}lignment is all we need for interpretability.
\newblock In \emph{CVPR}, pp.\  10319--10328, 2022.

\bibitem[Caron et~al.(2021)Caron, Touvron, Misra, J{\'{e}}gou, Mairal, Bojanowski, and Joulin]{Caron:2021:EPS}
Mathilde Caron, Hugo Touvron, Ishan Misra, Herv{\'{e}} J{\'{e}}gou, Julien Mairal, Piotr Bojanowski, and Armand Joulin.
\newblock Emerging properties in self-supervised vision transformers.
\newblock In \emph{ICCV}, pp.\  9630--9640, 2021.

\bibitem[Chen et~al.(2021{\natexlab{a}})Chen, Fan, and Panda]{Chen:2021:CCA}
Chun{-}Fu~(Richard) Chen, Quanfu Fan, and Rameswar Panda.
\newblock {CrossViT}: Cross-attention multi-scale vision transformer for image classification.
\newblock In \emph{ICCV}, pp.\  347--356, 2021{\natexlab{a}}.

\bibitem[Chen et~al.(2023)Chen, Wang, Changpinyo, Piergiovanni, Padlewski, Salz, Goodman, Grycner, Mustafa, Beyer, Kolesnikov, Puigcerver, Ding, Rong, Akbari, Mishra, Xue, Thapliyal, Bradbury, and Kuo]{Chen:2023:PAL}
Xi~Chen, Xiao Wang, Soravit Changpinyo, A.~J. Piergiovanni, Piotr Padlewski, Daniel Salz, Sebastian Goodman, Adam Grycner, Basil Mustafa, Lucas Beyer, Alexander Kolesnikov, Joan Puigcerver, Nan Ding, Keran Rong, Hassan Akbari, Gaurav Mishra, Linting Xue, Ashish~V. Thapliyal, James Bradbury, and Weicheng Kuo.
\newblock {PaLI}: {A} jointly-scaled multilingual language-image model.
\newblock In \emph{ICLR}, 2023.

\bibitem[Chen et~al.(2021{\natexlab{b}})Chen, Xie, and He]{Chen:2021:AES}
Xinlei Chen, Saining Xie, and Kaiming He.
\newblock An empirical study of training self-supervised vision transformers.
\newblock In \emph{ICCV}, pp.\  9620--9629, 2021{\natexlab{b}}.

\bibitem[Cherti et~al.(2023)Cherti, Beaumont, Wightman, Wortsman, Ilharco, Gordon, Schuhmann, Schmidt, and Jitsev]{Cherti:2023:RSL}
Mehdi Cherti, Romain Beaumont, Ross Wightman, Mitchell Wortsman, Gabriel Ilharco, Cade Gordon, Christoph Schuhmann, Ludwig Schmidt, and Jenia Jitsev.
\newblock Reproducible scaling laws for contrastive language-image learning.
\newblock In \emph{CVPR}, pp.\  2818--2829, 2023.

\bibitem[Chollet(2017)]{Chollet:2017:XDL}
Fran{\c{c}}ois Chollet.
\newblock Xception: Deep learning with depthwise separable convolutions.
\newblock In \emph{CVPR}, pp.\  1800--1807, 2017.

\bibitem[Croce \& Hein(2020)Croce and Hein]{Croce:2020:MDA}
Francesco Croce and Matthias Hein.
\newblock Minimally distorted adversarial examples with a fast adaptive boundary attack.
\newblock In \emph{ICML}, volume 119, pp.\  2196--2205, 2020.

\bibitem[Croce et~al.(2021)Croce, Andriushchenko, Sehwag, Debenedetti, Flammarion, Chiang, Mittal, and Hein]{Croce:2021:RBS}
Francesco Croce, Maksym Andriushchenko, Vikash Sehwag, Edoardo Debenedetti, Nicolas Flammarion, Mung Chiang, Prateek Mittal, and Matthias Hein.
\newblock {RobustBench}: A standardized adversarial robustness benchmark.
\newblock In \emph{NeurIPS Datasets and Benchmarks}, 2021.

\bibitem[Darcet et~al.(2024)Darcet, Oquab, Mairal, and Bojanowski]{Darcet:2024:VTN}
Timoth{\'{e}}e Darcet, Maxime Oquab, Julien Mairal, and Piotr Bojanowski.
\newblock Vision transformers need registers.
\newblock In \emph{ICLR}, 2024.

\bibitem[d'Ascoli et~al.(2021)d'Ascoli, Touvron, Leavitt, Morcos, Biroli, and Sagun]{Ascoli:2021:CIV}
St{\'{e}}phane d'Ascoli, Hugo Touvron, Matthew~L. Leavitt, Ari~S. Morcos, Giulio Biroli, and Levent Sagun.
\newblock {ConViT}: Improving vision transformers with soft convolutional inductive biases.
\newblock In \emph{ICML}, pp.\  2286--2296, 2021.

\bibitem[Dehghani et~al.(2022)Dehghani, Tay, Arnab, Beyer, and Vaswani]{Dehghani:2022:TEM}
Mostafa Dehghani, Yi~Tay, Anurag Arnab, Lucas Beyer, and Ashish Vaswani.
\newblock The efficiency misnomer.
\newblock In \emph{ICLR}, 2022.

\bibitem[Deng et~al.(2009)Deng, Dong, Socher, Li, Li, and Fei{-}Fei]{Deng:2009:ILS}
Jia Deng, Wei Dong, Richard Socher, Li{-}Jia Li, Kai Li, and Li~Fei{-}Fei.
\newblock {ImageNet}: {A} large-scale hierarchical image database.
\newblock In \emph{CVPR}, pp.\  248--255, 2009.

\bibitem[Ding et~al.(2022)Ding, Xiao, Codella, Luo, Wang, and Yuan]{Ding:2022:DAV}
Mingyu Ding, Bin Xiao, Noel Codella, Ping Luo, Jingdong Wang, and Lu~Yuan.
\newblock {DaViT}: Dual attention vision transformers.
\newblock In \emph{ECCV}, pp.\  74--92, 2022.

\bibitem[Dosovitskiy et~al.(2021)Dosovitskiy, Beyer, Kolesnikov, Weissenborn, Zhai, Unterthiner, Dehghani, Minderer, Heigold, Gelly, Uszkoreit, and Houlsby]{Dosovitskiy:2021:IWW}
Alexey Dosovitskiy, Lucas Beyer, Alexander Kolesnikov, Dirk Weissenborn, Xiaohua Zhai, Thomas Unterthiner, Mostafa Dehghani, Matthias Minderer, Georg Heigold, Sylvain Gelly, Jakob Uszkoreit, and Neil Houlsby.
\newblock An image is worth 16x16 words: Transformers for image recognition at scale.
\newblock In \emph{ICLR}, 2021.

\bibitem[Du et~al.(2021)Du, Yang, Zou, and Hu]{Du:FDL:2021}
Mengnan Du, Fan Yang, Na~Zou, and Xia Hu.
\newblock Fairness in deep learning: {A} computational perspective.
\newblock \emph{{IEEE} Intell. Syst.}, 36\penalty0 (4):\penalty0 25--34, 2021.

\bibitem[Engstrom et~al.(2019)Engstrom, Ilyas, Santurkar, DimitrisTsipras, Tran, and Madry]{Engstrom:2019:ARA}
Logan Engstrom, Andrew Ilyas, Shibani Santurkar, DimitrisTsipras, Brandon Tran, and Aleksander Madry.
\newblock Adversarial robustness as a prior for learned representations.
\newblock \emph{arXiv:1906.00945 [stat.ML]}, 2019.

\bibitem[Fang et~al.(2024{\natexlab{a}})Fang, Jose, Jain, Schmidt, Toshev, and Shankar]{Fang:2024:DFN}
Alex Fang, Albin~Madappally Jose, Amit Jain, Ludwig Schmidt, Alexander~T. Toshev, and Vaishaal Shankar.
\newblock Data filtering networks.
\newblock In \emph{ICLR}, 2024{\natexlab{a}}.

\bibitem[Fang et~al.(2023)Fang, Wang, Xie, Sun, Wu, Wang, Huang, Wang, and Cao]{Fang:2023:EET}
Yuxin Fang, Wen Wang, Binhui Xie, Quan Sun, Ledell Wu, Xinggang Wang, Tiejun Huang, Xinlong Wang, and Yue Cao.
\newblock {EVA}:robustness met limits of masked visual representation learning at scale.
\newblock In \emph{CVPR}, pp.\  19358--19369, 2023.

\bibitem[Fang et~al.(2024{\natexlab{b}})Fang, Sun, Wang, Huang, Wang, and Cao]{Fang:2023:EVA}
Yuxin Fang, Quan Sun, Xinggang Wang, Tiejun Huang, Xinlong Wang, and Yue Cao.
\newblock {EVA-02:} {A} visual representation for neon genesis.
\newblock \emph{Image Vis. Comput.}, 149:\penalty0 105171, 2024{\natexlab{b}}.

\bibitem[Gadre et~al.(2023)Gadre, Ilharco, Fang, Hayase, Smyrnis, Nguyen, Marten, Wortsman, Ghosh, Zhang, Orgad, Entezari, Daras, Pratt, Ramanujan, Bitton, Marathe, Mussmann, Vencu, Cherti, Krishna, Koh, Saukh, Ratner, Song, Hajishirzi, Farhadi, Beaumont, Oh, Dimakis, Jitsev, Carmon, Shankar, and Schmidt]{Gadre:2023:DIS}
Samir~Yitzhak Gadre, Gabriel Ilharco, Alex Fang, Jonathan Hayase, Georgios Smyrnis, Thao Nguyen, Ryan Marten, Mitchell Wortsman, Dhruba Ghosh, Jieyu Zhang, Eyal Orgad, Rahim Entezari, Giannis Daras, Sarah~M. Pratt, Vivek Ramanujan, Yonatan Bitton, Kalyani Marathe, Stephen Mussmann, Richard Vencu, Mehdi Cherti, Ranjay Krishna, Pang~Wei Koh, Olga Saukh, Alexander~J. Ratner, Shuran Song, Hannaneh Hajishirzi, Ali Farhadi, Romain Beaumont, Sewoong Oh, Alex Dimakis, Jenia Jitsev, Yair Carmon, Vaishaal Shankar, and Ludwig Schmidt.
\newblock Datacomp: In search of the next generation of multimodal datasets.
\newblock In \emph{NeurIPS Datasets and Benchmarks}, 2023.

\bibitem[Geirhos et~al.(2019)Geirhos, Rubisch, Michaelis, Bethge, Wichmann, and Brendel]{Geirhos:2019:ITC}
Robert Geirhos, Patricia Rubisch, Claudio Michaelis, Matthias Bethge, Felix~A. Wichmann, and Wieland Brendel.
\newblock {ImageNet}-trained {CNN}s are biased towards texture; increasing shape bias improves accuracy and robustness.
\newblock In \emph{ICLR}, 2019.

\bibitem[Geirhos et~al.(2021)Geirhos, Narayanappa, Mitzkus, Thieringer, Bethge, Wichmann, and Brendel]{Geirhos:2021:PSI}
Robert Geirhos, Kantharaju Narayanappa, Benjamin Mitzkus, Tizian Thieringer, Matthias Bethge, Felix~A. Wichmann, and Wieland Brendel.
\newblock Partial success in closing the gap between human and machine vision.
\newblock In \emph{NeurIPS}, pp.\  23885--23899, 2021.

\bibitem[Goldblum et~al.(2023)Goldblum, Souri, Ni, Shu, Prabhu, Somepalli, Chattopadhyay, Ibrahim, Bardes, Hoffman, Chellappa, Wilson, and Goldstein]{Goldblum:2023:BBL}
Micah Goldblum, Hossein Souri, Renkun Ni, Manli Shu, Viraj Prabhu, Gowthami Somepalli, Prithvijit Chattopadhyay, Mark Ibrahim, Adrien Bardes, Judy Hoffman, Rama Chellappa, Andrew~Gordon Wilson, and Tom Goldstein.
\newblock Battle of the backbones: {A} large-scale comparison of pretrained models across computer vision tasks.
\newblock In \emph{NeurIPS}, 2023.

\bibitem[Goodfellow et~al.(2015)Goodfellow, Shlens, and Szegedy]{Goodfellow:2015:EHA}
Ian Goodfellow, Jonathon Shlens, and Christian Szegedy.
\newblock Explaining and harnessing adversarial examples.
\newblock In \emph{ICLR}, 2015.

\bibitem[Grabinski et~al.(2022)Grabinski, Gavrikov, Keuper, and Keuper]{Grabinski:2022:RBA}
Julia Grabinski, Paul Gavrikov, Janis Keuper, and Margret Keuper.
\newblock Robust models are less over-confident.
\newblock In \emph{{NeurIPS}}, 2022.

\bibitem[Graham et~al.(2021)Graham, El{-}Nouby, Touvron, Stock, Joulin, J{\'{e}}gou, and Douze]{Graham:2021:LAV}
Benjamin Graham, Alaaeldin El{-}Nouby, Hugo Touvron, Pierre Stock, Armand Joulin, Herv{\'{e}} J{\'{e}}gou, and Matthijs Douze.
\newblock {LeViT}: A vision transformer in {ConvNet}'s clothing for faster inference.
\newblock In \emph{ICCV}, pp.\  12239--12249, 2021.

\bibitem[Guo et~al.(2017)Guo, Pleiss, Sun, and Weinberger]{Guo:2017:OCM}
Chuan Guo, Geoff Pleiss, Yu~Sun, and Kilian~Q. Weinberger.
\newblock On calibration of modern neural networks.
\newblock In \emph{ICML}, pp.\  1321--1330, 2017.

\bibitem[He et~al.(2016)He, Zhang, Ren, and Sun]{He:2016:DRL}
Kaiming He, Xiangyu Zhang, Shaoqing Ren, and Jian Sun.
\newblock Deep residual learning for image recognition.
\newblock In \emph{CVPR}, pp.\  770--778, 2016.

\bibitem[He et~al.(2022)He, Chen, Xie, Li, Doll{\'{a}}r, and Girshick]{He:2022:MAA}
Kaiming He, Xinlei Chen, Saining Xie, Yanghao Li, Piotr Doll{\'{a}}r, and Ross~B. Girshick.
\newblock Masked autoencoders are scalable vision learners.
\newblock In \emph{CVPR}, pp.\  15979--15988, 2022.

\bibitem[He et~al.(2019)He, Zhang, Zhang, Zhang, Xie, and Li]{He:2019:BOT}
Tong He, Zhi Zhang, Hang Zhang, Zhongyue Zhang, Junyuan Xie, and Mu~Li.
\newblock Bag of tricks for image classification with convolutional neural networks.
\newblock In \emph{CVPR}, pp.\  558--567, 2019.

\bibitem[Hendrycks \& Dietterich(2019)Hendrycks and Dietterich]{Hendrycks:2019:BNN}
Dan Hendrycks and Thomas~G. Dietterich.
\newblock Benchmarking neural network robustness to common corruptions and perturbations.
\newblock In \emph{ICLR}, 2019.

\bibitem[Hendrycks et~al.(2019)Hendrycks, Mazeika, Kadavath, and Song]{Hendrycks:2019:USS}
Dan Hendrycks, Mantas Mazeika, Saurav Kadavath, and Dawn Song.
\newblock Using self-supervised learning can improve model robustness and uncertainty.
\newblock In \emph{{NeurIPS}}, pp.\  15637--15648, 2019.

\bibitem[Hendrycks et~al.(2020)Hendrycks, Mu, Cubuk, Zoph, Gilmer, and Lakshminarayanan]{Hendrycks:2020:SDP}
Dan Hendrycks, Norman Mu, Ekin~Dogus Cubuk, Barret Zoph, Justin Gilmer, and Balaji Lakshminarayanan.
\newblock {AugMix}: {A} simple data processing method to improve robustness and uncertainty.
\newblock In \emph{ICLR}, 2020.

\bibitem[Hendrycks et~al.(2021)Hendrycks, Basart, Mu, Kadavath, Wang, Dorundo, Desai, Zhu, Parajuli, Guo, Song, Steinhardt, and Gilmer]{Hendrycks:2021:MFR}
Dan Hendrycks, Steven Basart, Norman Mu, Saurav Kadavath, Frank Wang, Evan Dorundo, Rahul Desai, Tyler Zhu, Samyak Parajuli, Mike Guo, Dawn Song, Jacob Steinhardt, and Justin Gilmer.
\newblock The many faces of robustness: {A} critical analysis of out-of-distribution generalization.
\newblock In \emph{ICCV}, pp.\  8320--8329, 2021.

\bibitem[Heo et~al.(2021)Heo, Yun, Han, Chun, Choe, and Oh]{Heo:2021:RSD}
Byeongho Heo, Sangdoo Yun, Dongyoon Han, Sanghyuk Chun, Junsuk Choe, and Seong~Joon Oh.
\newblock Rethinking spatial dimensions of vision transformers.
\newblock In \emph{ICCV}, pp.\  11916--11925, 2021.

\bibitem[Hermann et~al.(2020)Hermann, Chen, and Kornblith]{Hermann:2020:TOP}
Katherine~L. Hermann, Ting Chen, and Simon Kornblith.
\newblock The origins and prevalence of texture bias in convolutional neural networks.
\newblock In \emph{NeurIPS}, 2020.

\bibitem[Hesse et~al.(2023{\natexlab{a}})Hesse, Schaub-Meyer, and Roth]{Hesse:2023:CAD}
Robin Hesse, Simone Schaub-Meyer, and Stefan Roth.
\newblock Content-adaptive downsampling in convolutional neural networks.
\newblock In \emph{{CVPR} Workshops}, pp.\  4544--4553, 2023{\natexlab{a}}.

\bibitem[Hesse et~al.(2023{\natexlab{b}})Hesse, Schaub-Meyer, and Roth]{Hesse:2023:SVD}
Robin Hesse, Simone Schaub-Meyer, and Stefan Roth.
\newblock Funny{B}irds: {A} synthetic vision dataset for a part-based analysis of explainable {AI} methods.
\newblock In \emph{ICCV}, pp.\  3981--3991, 2023{\natexlab{b}}.

\bibitem[Hesse et~al.(2024)Hesse, Schaub-Meyer, and Roth]{Hesse:2024:IDSDS}
Robin Hesse, Simone Schaub-Meyer, and Stefan Roth.
\newblock Benchmarking the attribution quality of vision models.
\newblock In \emph{NeurIPS}, 2024.

\bibitem[Hinton et~al.(2015)Hinton, Vinyals, and Dean]{Hinton:2015:DTK}
Geoffrey~E. Hinton, Oriol Vinyals, and Jeffrey Dean.
\newblock Distilling the knowledge in a neural network.
\newblock \emph{arXiv:1503.02531 [stat.ML]}, 2015.

\bibitem[Howard et~al.(2019)Howard, Pang, Adam, Le, Sandler, Chen, Wang, Chen, Tan, Chu, Vasudevan, and Zhu]{Howard:2019:SFM}
Andrew Howard, Ruoming Pang, Hartwig Adam, Quoc~V. Le, Mark Sandler, Bo~Chen, Weijun Wang, Liang{-}Chieh Chen, Mingxing Tan, Grace Chu, Vijay Vasudevan, and Yukun Zhu.
\newblock Searching for {MobileNetV3}.
\newblock In \emph{ICCV}, pp.\  1314--1324, 2019.

\bibitem[Hu et~al.(2018)Hu, Shen, and Sun]{Hu:2018:SEN}
Jie Hu, Li~Shen, and Gang Sun.
\newblock Squeeze-and-excitation networks.
\newblock In \emph{CVPR}, pp.\  7132--7141, 2018.

\bibitem[Huang et~al.(2017)Huang, Liu, van~der Maaten, and Weinberger]{Huang:2017:DCC}
Gao Huang, Zhuang Liu, Laurens van~der Maaten, and Kilian~Q. Weinberger.
\newblock Densely connected convolutional networks.
\newblock In \emph{CVPR}, pp.\  770--778, 2017.

\bibitem[Iandola et~al.(2016)Iandola, Moskewicz, Ashraf, Han, Dally, and Keutzer]{Iandola:2016:SAA}
Forrest~N. Iandola, Matthew~W. Moskewicz, Khalid Ashraf, Song Han, William~J. Dally, and Kurt Keutzer.
\newblock {SqueezeNet}: {AlexNet}-level accuracy with 50x fewer parameters and less than 1{MB} model size.
\newblock \emph{arXiv:1602.07360 [cs.CV]}, 2016.

\bibitem[Jo \& Bengio(2017)Jo and Bengio]{Jo:2017:MTT}
Jason Jo and Yoshua Bengio.
\newblock Measuring the tendency of {CNN}s to learn surface statistical regularities.
\newblock \emph{arXiv:1711.11561 [cs.LG]}, 2017.

\bibitem[Kaplan et~al.(2020)Kaplan, McCandlish, Henighan, Brown, Chess, Child, Gray, Radford, Wu, and Amodei]{Kaplan:2020:SLF}
Jared Kaplan, Sam McCandlish, Tom Henighan, Tom~B. Brown, Benjamin Chess, Rewon Child, Scott Gray, Alec Radford, Jeffrey Wu, and Dario Amodei.
\newblock Scaling laws for neural language models.
\newblock \emph{CoRR}, 2020.

\bibitem[Kim(2020)]{Kim:2020:TPR}
Hoki Kim.
\newblock Torchattacks: A {PyTorch} repository for adversarial attacks.
\newblock \emph{arXiv:2010.01950 [cs.LG]}, 2020.

\bibitem[Knuth(1992)]{Knuth:1992:TNN}
Donald~E. Knuth.
\newblock Two notes on notation.
\newblock \emph{The American Mathematical Monthly}, 99\penalty0 (5):\penalty0 403--422, 1992.

\bibitem[Kolesnikov et~al.(2020)Kolesnikov, Beyer, Zhai, Puigcerver, Yung, Gelly, and Houlsby]{Kolesnikov:2020:BIT}
Alexander Kolesnikov, Lucas Beyer, Xiaohua Zhai, Joan Puigcerver, Jessica Yung, Sylvain Gelly, and Neil Houlsby.
\newblock {Big Transfer (BiT)}: General visual representation learning.
\newblock In \emph{ECCV}, pp.\  491--507, 2020.

\bibitem[Krizhevsky et~al.(2012)Krizhevsky, Sutskever, and Hinton]{Krizhevsky:2012:INC}
Alex Krizhevsky, Ilya Sutskever, and Geoffrey~E. Hinton.
\newblock {ImageNet} classification with deep convolutional neural networks.
\newblock In \emph{{NIPS}}, pp.\  1106--1114, 2012.

\bibitem[Kuzucu et~al.(2024)Kuzucu, Cheong, Gunes, and Kalkan]{Kuzucu:2024:UFM}
Selim Kuzucu, Jiaee Cheong, Hatice Gunes, and Sinan Kalkan.
\newblock Uncertainty as a fairness measure.
\newblock \emph{J. Artif. Intell. Res.}, 81:\penalty0 307--335, 2024.

\bibitem[Lakshminarayanan et~al.(2017)Lakshminarayanan, Pritzel, and Blundell]{Lakshminarayanan:2017:SSP}
Balaji Lakshminarayanan, Alexander Pritzel, and Charles Blundell.
\newblock Simple and scalable predictive uncertainty estimation using deep ensembles.
\newblock In \emph{NIPS}, pp.\  6402--6413, 2017.

\bibitem[Li et~al.(2022{\natexlab{a}})Li, Wu, Fan, Mangalam, Xiong, Malik, and Feichtenhofer]{Li:2022:MIM}
Yanghao Li, Chao{-}Yuan Wu, Haoqi Fan, Karttikeya Mangalam, Bo~Xiong, Jitendra Malik, and Christoph Feichtenhofer.
\newblock {MViTv2}: Improved multiscale vision transformers for classification and detection.
\newblock In \emph{CVPR}, pp.\  4794--4804, 2022{\natexlab{a}}.

\bibitem[Li et~al.(2022{\natexlab{b}})Li, Yuan, Wen, Hu, Evangelidis, Tulyakov, Wang, and Ren]{Li:2022:EVT}
Yanyu Li, Geng Yuan, Yang Wen, Ju~Hu, Georgios Evangelidis, Sergey Tulyakov, Yanzhi Wang, and Jian Ren.
\newblock {EfficientFormer}: Vision transformers at {MobileNet} speed.
\newblock In \emph{NeurIPS}, 2022{\natexlab{b}}.

\bibitem[Liu et~al.(2025)Liu, Dong, Xiang, Yang, Su, Zhu, Chen, He, Xue, and Zheng]{Liu:2023:CSR}
Chang Liu, Yinpeng Dong, Wenzhao Xiang, Xiao Yang, Hang Su, Jun Zhu, Yuefeng Chen, Yuan He, Hui Xue, and Shibao Zheng.
\newblock A comprehensive study on robustness of image classification models: Benchmarking and rethinking.
\newblock \emph{Int. J. Comput. Vis.}, 133\penalty0 (2):\penalty0 567--589, 2025.

\bibitem[Liu et~al.(2023)Liu, Chaudhary, and Wang]{Liu:2023:TTA}
Haoyang Liu, Maheep Chaudhary, and Haohan Wang.
\newblock Towards trustworthy and aligned machine learning: {A} data-centric survey with causality perspectives.
\newblock \emph{arXiv:2307.16851 [cs.CL]}, 2023.

\bibitem[Liu et~al.(2021)Liu, Lin, Cao, Hu, Wei, Zhang, Lin, and Guo]{Liu:2021:STH}
Ze~Liu, Yutong Lin, Yue Cao, Han Hu, Yixuan Wei, Zheng Zhang, Stephen Lin, and Baining Guo.
\newblock {Swin Transformer}: Hierarchical vision transformer using shifted windows.
\newblock In \emph{ICCV}, pp.\  9992--10002, 2021.

\bibitem[Liu et~al.(2022{\natexlab{a}})Liu, Hu, Lin, Yao, Xie, Wei, Ning, Cao, Zhang, Dong, Wei, and Guo]{Liu:2022:STS}
Ze~Liu, Han Hu, Yutong Lin, Zhuliang Yao, Zhenda Xie, Yixuan Wei, Jia Ning, Yue Cao, Zheng Zhang, Li~Dong, Furu Wei, and Baining Guo.
\newblock {Swin Transformer V2}: Scaling up capacity and resolution.
\newblock In \emph{CVPR}, pp.\  11999--12009, 2022{\natexlab{a}}.

\bibitem[Liu et~al.(2022{\natexlab{b}})Liu, Mao, Wu, Feichtenhofer, Darrell, and Xie]{Liu:2022:ACF}
Zhuang Liu, Hanzi Mao, Chao{-}Yuan Wu, Christoph Feichtenhofer, Trevor Darrell, and Saining Xie.
\newblock A {ConvNet} for the 2020s.
\newblock In \emph{CVPR}, pp.\  11966--11976, 2022{\natexlab{b}}.

\bibitem[Lorenz et~al.(2021)Lorenz, Strassel, Keuper, and Keuper]{Lorenz:2021:IRA}
Peter Lorenz, Dominik Strassel, Margret Keuper, and Janis Keuper.
\newblock Is robustbench/autoattack a suitable benchmark for adversarial robustness?
\newblock \emph{CoRR}, 2021.

\bibitem[Loshchilov \& Hutter(2017)Loshchilov and Hutter]{Loshchilov:2017:FWD}
Ilya Loshchilov and Frank Hutter.
\newblock Fixing weight decay regularization in {A}dam.
\newblock In \emph{ICLR}, 2017.

\bibitem[Ma et~al.(2018)Ma, Zhang, Zheng, and Sun]{Ma:2018:SPG}
Ningning Ma, Xiangyu Zhang, Hai{-}Tao Zheng, and Jian Sun.
\newblock {ShuffleNet} {V2}: Practical guidelines for efficient {CNN} architecture design.
\newblock In \emph{ECCV}, pp.\  122--138, 2018.

\bibitem[Madry et~al.(2018)Madry, Makelov, Schmidt, Tsipras, and Vladu]{Madry:2018:TDL}
Aleksander Madry, Aleksandar Makelov, Ludwig Schmidt, Dimitris Tsipras, and Adrian Vladu.
\newblock Towards deep learning models resistant to adversarial attacks.
\newblock In \emph{ICLR}, 2018.

\bibitem[Mayilvahanan et~al.(2025)Mayilvahanan, Zimmermann, Wiedemer, Rusak, Juhos, Bethge, and Brendel]{Mayilvahanan:2024:ISF}
Prasanna Mayilvahanan, Roland~S. Zimmermann, Thadd{\"{a}}us Wiedemer, Evgenia Rusak, Attila Juhos, Matthias Bethge, and Wieland Brendel.
\newblock In search of forgotten domain generalization.
\newblock In \emph{ICLR}, 2025.

\bibitem[Miller et~al.(2021)Miller, Taori, Raghunathan, Sagawa, Koh, Shankar, Liang, Carmon, and Schmidt]{Miller:2021:AOT}
John Miller, Rohan Taori, Aditi Raghunathan, Shiori Sagawa, Pang~Wei Koh, Vaishaal Shankar, Percy Liang, Yair Carmon, and Ludwig Schmidt.
\newblock Accuracy on the line: On the strong correlation between out-of-distribution and in-distribution generalization.
\newblock In \emph{ICML}, pp.\  7721--7735, 2021.

\bibitem[Minderer et~al.(2021)Minderer, Djolonga, Romijnders, Hubis, Zhai, Houlsby, Tran, and Lucic]{Minderer:2021:RCM}
Matthias Minderer, Josip Djolonga, Rob Romijnders, Frances Hubis, Xiaohua Zhai, Neil Houlsby, Dustin Tran, and Mario Lucic.
\newblock Revisiting the calibration of modern neural networks.
\newblock In \emph{NeurIPS}, pp.\  15682--15694, 2021.

\bibitem[M{\"{u}}ller et~al.(2019)M{\"{u}}ller, Kornblith, and Hinton]{Mueller:2019:WDL}
Rafael M{\"{u}}ller, Simon Kornblith, and Geoffrey~E. Hinton.
\newblock When does label smoothing help?
\newblock In \emph{NeurIPS}, pp.\  4696--4705, 2019.

\bibitem[Nakkiran(2019)]{Nakkiran:2019:ARM}
Preetum Nakkiran.
\newblock Adversarial robustness may be at odds with simplicity.
\newblock \emph{arXiv:1901.00532 [cs.LG]}, 2019.

\bibitem[Nam et~al.(2021)Nam, Lee, Park, Yoon, and Yoo]{Nam:2021:RDG}
Hyeonseob Nam, HyunJae Lee, Jongchan Park, Wonjun Yoon, and Donggeun Yoo.
\newblock Reducing domain gap by reducing style bias.
\newblock In \emph{CVPR}, pp.\  8690--8699, 2021.

\bibitem[Nixon et~al.(2019)Nixon, Dusenberry, Zhang, Jerfel, and Tran]{Nixon:2019:MCD}
Jeremy Nixon, Michael~W. Dusenberry, Linchuan Zhang, Ghassen Jerfel, and Dustin Tran.
\newblock Measuring calibration in deep learning.
\newblock In \emph{{CVPR} Workshops}, pp.\  38--41, 2019.

\bibitem[Oquab et~al.(2024)Oquab, Darcet, Moutakanni, Vo, Szafraniec, Khalidov, Fernandez, Haziza, Massa, El{-}Nouby, Assran, Ballas, Galuba, Howes, Huang, Li, Misra, Rabbat, Sharma, Synnaeve, Xu, J{\'{e}}gou, Mairal, Labatut, Joulin, and Bojanowski]{Oquab:2024:DIN}
Maxime Oquab, Timoth{\'{e}}e Darcet, Th{\'{e}}o Moutakanni, Huy~V. Vo, Marc Szafraniec, Vasil Khalidov, Pierre Fernandez, Daniel Haziza, Francisco Massa, Alaaeldin El{-}Nouby, Mido Assran, Nicolas Ballas, Wojciech Galuba, Russell Howes, Po{-}Yao Huang, Shang{-}Wen Li, Ishan Misra, Michael Rabbat, Vasu Sharma, Gabriel Synnaeve, Hu~Xu, Herv{\'{e}} J{\'{e}}gou, Julien Mairal, Patrick Labatut, Armand Joulin, and Piotr Bojanowski.
\newblock {DINOv2}: {L}earning robust visual features without supervision.
\newblock \emph{Trans. Mach. Learn. Res.}, 2024, 2024.

\bibitem[Papernot et~al.(2016)Papernot, McDaniel, Wu, Jha, and Swami]{Papernot:2016:DDA}
Nicolas Papernot, Patrick McDaniel, Xi~Wu, Somesh Jha, and Ananthram Swami.
\newblock Distillation as a defense to adversarial perturbations against deep neural networks.
\newblock In \emph{IEEE Symposium on Security and Privacy}, pp.\  582--597, 2016.

\bibitem[Paszke et~al.(2019)Paszke, Gross, Massa, Lerer, Bradbury, Chanan, Killeen, Lin, Gimelshein, Antiga, Desmaison, K{\"{o}}pf, Yang, DeVito, Raison, Tejani, Chilamkurthy, Steiner, Fang, Bai, and Chintala]{Paszke:2019:PAI}
Adam Paszke, Sam Gross, Francisco Massa, Adam Lerer, James Bradbury, Gregory Chanan, Trevor Killeen, Zeming Lin, Natalia Gimelshein, Luca Antiga, Alban Desmaison, Andreas K{\"{o}}pf, Edward~Z. Yang, Zachary DeVito, Martin Raison, Alykhan Tejani, Sasank Chilamkurthy, Benoit Steiner, Lu~Fang, Junjie Bai, and Soumith Chintala.
\newblock {PyTorch}: An imperative style, high-performance deep learning library.
\newblock In \emph{NeurIPS}, pp.\  8024--8035, 2019.

\bibitem[Peng et~al.(2022)Peng, Dong, Bao, Ye, and Wei]{Peng:2022:BMI}
Zhiliang Peng, Li~Dong, Hangbo Bao, Qixiang Ye, and Furu Wei.
\newblock {BEiT v2}: Masked image modeling with vector-quantized visual tokenizers.
\newblock \emph{arXiv:2208.06366 [cs.CV]}, 2022.

\bibitem[Radford et~al.(2021)Radford, Kim, Hallacy, Ramesh, Goh, Agarwal, Sastry, Askell, Mishkin, Clark, Krueger, and Sutskever]{Radford:2021:LTV}
Alec Radford, Jong~Wook Kim, Chris Hallacy, Aditya Ramesh, Gabriel Goh, Sandhini Agarwal, Girish Sastry, Amanda Askell, Pamela Mishkin, Jack Clark, Gretchen Krueger, and Ilya Sutskever.
\newblock Learning transferable visual models from natural language supervision.
\newblock In \emph{{ICML}}, pp.\  8748--8763, 2021.

\bibitem[Radosavovic et~al.(2020)Radosavovic, Kosaraju, Girshick, He, and Doll{\'{a}}r]{Radosavovic:2020:DND}
Ilija Radosavovic, Raj~Prateek Kosaraju, Ross~B. Girshick, Kaiming He, and Piotr Doll{\'{a}}r.
\newblock Designing network design spaces.
\newblock In \emph{CVPR}, pp.\  10425--10433, 2020.

\bibitem[Russakovsky et~al.(2015)Russakovsky, Deng, Su, Krause, Satheesh, Ma, Huang, Karpathy, Khosla, Bernstein, Berg, and Fei-Fei]{Russakovsky:2015:ILS}
Olga Russakovsky, Jia Deng, Hao Su, Jonathan Krause, Sanjeev Satheesh, Sean Ma, Zhiheng Huang, Andrej Karpathy, Aditya Khosla, Michael Bernstein, Alexander~C. Berg, and Li~Fei-Fei.
\newblock {ImageNet} large scale visual recognition challenge.
\newblock \emph{Int. J. Comput. Vision}, 115\penalty0 (13):\penalty0 211--252, 2015.

\bibitem[Ryali et~al.(2023)Ryali, Hu, Bolya, Wei, Fan, Huang, Aggarwal, Chowdhury, Poursaeed, Hoffman, Malik, Li, and Feichtenhofer]{Ryali:2023:HHV}
Chaitanya Ryali, Yuan{-}Ting Hu, Daniel Bolya, Chen Wei, Haoqi Fan, Po{-}Yao Huang, Vaibhav Aggarwal, Arkabandhu Chowdhury, Omid Poursaeed, Judy Hoffman, Jitendra Malik, Yanghao Li, and Christoph Feichtenhofer.
\newblock Hiera: {A} hierarchical vision transformer without the bells-and-whistles.
\newblock In \emph{ICML}, pp.\  29441--29454, 2023.

\bibitem[Salman et~al.(2020)Salman, Ilyas, Engstrom, Kapoor, and Madry]{Salman:2020:ARI}
Hadi Salman, Andrew Ilyas, Logan Engstrom, Ashish Kapoor, and Aleksander Madry.
\newblock Do adversarially robust imagenet models transfer better?
\newblock In \emph{NeurIPS}, 2020.

\bibitem[Sandler et~al.(2018)Sandler, Howard, Zhu, Zhmoginov, and Chen]{Sandler:MIR:2018}
Mark Sandler, Andrew~G. Howard, Menglong Zhu, Andrey Zhmoginov, and Liang{-}Chieh Chen.
\newblock {MobileNetV2}: Inverted residuals and linear bottlenecks.
\newblock In \emph{CVPR}, pp.\  4510--4520, 2018.

\bibitem[Schuhmann et~al.(2022)Schuhmann, Beaumont, Vencu, Gordon, Wightman, Cherti, Coombes, Katta, Mullis, Wortsman, Schramowski, Kundurthy, Crowson, Schmidt, Kaczmarczyk, and Jitsev]{Schuhmann:2022:AOL}
Christoph Schuhmann, Romain Beaumont, Richard Vencu, Cade Gordon, Ross Wightman, Mehdi Cherti, Theo Coombes, Aarush Katta, Clayton Mullis, Mitchell Wortsman, Patrick Schramowski, Srivatsa Kundurthy, Katherine Crowson, Ludwig Schmidt, Robert Kaczmarczyk, and Jenia Jitsev.
\newblock {LAION-5B}: An open large-scale dataset for training next generation image-text models.
\newblock In \emph{NeurIPS}, 2022.

\bibitem[Shi et~al.(2020)Shi, Zhang, Dai, Zhu, Mu, and Wang]{Shi:2020:IDR}
Baifeng Shi, Dinghuai Zhang, Qi~Dai, Zhanxing Zhu, Yadong Mu, and Jingdong Wang.
\newblock Informative dropout for robust representation learning: {A} shape-bias perspective.
\newblock In \emph{ICML}, pp.\  8828--8839, 2020.

\bibitem[Simonyan \& Zisserman(2015)Simonyan and Zisserman]{Simonyan:2015:VDC}
Karen Simonyan and Andrew Zisserman.
\newblock Very deep convolutional networks for large-scale image recognition.
\newblock In \emph{ICLR}, 2015.

\bibitem[Singh et~al.(2023)Singh, Croce, and Hein]{Singh:2023:RAT}
Naman~Deep Singh, Francesco Croce, and Matthias Hein.
\newblock Revisiting adversarial training for {ImageNet}: Architectures, training and generalization across threat models.
\newblock In \emph{{NeurIPS}}, 2023.

\bibitem[Steiner et~al.(2022)Steiner, Kolesnikov, Zhai, Wightman, Uszkoreit, and Beyer]{Steiner:2022:HTT}
Andreas Steiner, Alexander Kolesnikov, Xiaohua Zhai, Ross Wightman, Jakob Uszkoreit, and Lucas Beyer.
\newblock How to train your {ViT}? {D}ata, augmentation, and regularization in vision transformers.
\newblock \emph{Trans. Mach. Learn. Res.}, 2022.

\bibitem[Sun et~al.(2017)Sun, Shrivastava, Singh, and Gupta]{Sun:2017:RUE}
Chen Sun, Abhinav Shrivastava, Saurabh Singh, and Abhinav Gupta.
\newblock Revisiting unreasonable effectiveness of data in deep learning era.
\newblock In \emph{ICCV}, pp.\  843--852, 2017.

\bibitem[Szegedy et~al.(2014)Szegedy, Zaremba, Sutskever, Bruna, Erhan, Goodfellow, and Fergus]{Szegedy:2014:IPN}
Christian Szegedy, Wojciech Zaremba, Ilya Sutskever, Joan Bruna, Dumitru Erhan, Ian~J. Goodfellow, and Rob Fergus.
\newblock Intriguing properties of neural networks.
\newblock In \emph{ICLR}, 2014.

\bibitem[Szegedy et~al.(2015)Szegedy, Liu, Jia, Sermanet, Reed, Anguelov, Erhan, Vanhoucke, and Rabinovich]{Szegedy:2015:GDC}
Christian Szegedy, Wei Liu, Yangqing Jia, Pierre Sermanet, Scott~E. Reed, Dragomir Anguelov, Dumitru Erhan, Vincent Vanhoucke, and Andrew Rabinovich.
\newblock Going deeper with convolutions.
\newblock In \emph{CVPR}, pp.\  1--9, 2015.

\bibitem[Szegedy et~al.(2016)Szegedy, Vanhoucke, Ioffe, Shlens, and Wojna]{Szegedy:2016:RTI}
Christian Szegedy, Vincent Vanhoucke, Sergey Ioffe, Jonathon Shlens, and Zbigniew Wojna.
\newblock Rethinking the {Inception} architecture for computer vision.
\newblock In \emph{CVPR}, pp.\  2818--2826, 2016.

\bibitem[Szegedy et~al.(2017)Szegedy, Ioffe, Vanhoucke, and Alemi]{Szegedy:2017:IIA}
Christian Szegedy, Sergey Ioffe, Vincent Vanhoucke, and Alexander~A. Alemi.
\newblock Inception-v4, {Inception-ResNet} and the impact of residual connections on learning.
\newblock In \emph{AAAI}, pp.\  4278--4284, 2017.

\bibitem[Tan \& Le(2019)Tan and Le]{Tan:2019:ERM}
Mingxing Tan and Quoc~V. Le.
\newblock {EfficientNet}: Rethinking model scaling for convolutional neural networks.
\newblock In \emph{ICML}, volume~97, pp.\  6105--6114, 2019.

\bibitem[Tan \& Le(2021)Tan and Le]{Tan:2021:ESM}
Mingxing Tan and Quoc~V. Le.
\newblock {EfficientNetV2}: Smaller models and faster training.
\newblock In \emph{{ICML}}, pp.\  10096--10106, 2021.

\bibitem[Tan et~al.(2019)Tan, Chen, Pang, Vasudevan, Sandler, Howard, and Le]{Tan:2019:MPA}
Mingxing Tan, Bo~Chen, Ruoming Pang, Vijay Vasudevan, Mark Sandler, Andrew Howard, and Quoc~V. Le.
\newblock {MnasNet}: Platform-aware neural architecture search for mobile.
\newblock In \emph{CVPR}, pp.\  2820--2828, 2019.

\bibitem[Tay et~al.(2022)Tay, Dehghani, Rao, Fedus, Abnar, Chung, Narang, Yogatama, Vaswani, and Metzler]{Tay:SEI:2021}
Yi~Tay, Mostafa Dehghani, Jinfeng Rao, William Fedus, Samira Abnar, Hyung~Won Chung, Sharan Narang, Dani Yogatama, Ashish Vaswani, and Donald Metzler.
\newblock Scale efficiently: Insights from pretraining and finetuning transformers.
\newblock In \emph{ICLR}, 2022.

\bibitem[Thomee et~al.(2016)Thomee, Shamma, Friedland, Elizalde, Ni, Poland, Borth, and Li]{Thomee:2016:YTD}
Bart Thomee, David~A. Shamma, Gerald Friedland, Benjamin Elizalde, Karl Ni, Douglas Poland, Damian Borth, and Li{-}Jia Li.
\newblock {YFCC100M}: The new data in multimedia research.
\newblock \emph{Commun. {ACM}}, 59\penalty0 (2):\penalty0 64--73, 2016.

\bibitem[Touvron et~al.(2021{\natexlab{a}})Touvron, Cord, Douze, Massa, Sablayrolles, and J{\'{e}}gou]{Touvron:2021:TDE}
Hugo Touvron, Matthieu Cord, Matthijs Douze, Francisco Massa, Alexandre Sablayrolles, and Herv{\'{e}} J{\'{e}}gou.
\newblock Training data-efficient image transformers {\&} distillation through attention.
\newblock In \emph{ICML}, pp.\  10347--10357, 2021{\natexlab{a}}.

\bibitem[Touvron et~al.(2021{\natexlab{b}})Touvron, Cord, Sablayrolles, Synnaeve, and J{\'{e}}gou]{Touvron:2021:GDW}
Hugo Touvron, Matthieu Cord, Alexandre Sablayrolles, Gabriel Synnaeve, and Herv{\'{e}} J{\'{e}}gou.
\newblock Going deeper with image transformers.
\newblock In \emph{ICCV}, pp.\  32--42, 2021{\natexlab{b}}.

\bibitem[Touvron et~al.(2022)Touvron, Cord, and J{\'{e}}gou]{Touvron:2022:DRO}
Hugo Touvron, Matthieu Cord, and Herv{\'{e}} J{\'{e}}gou.
\newblock {DeiT III}: Revenge of the {ViT}.
\newblock In \emph{{ECCV}}, pp.\  516--533, 2022.

\bibitem[Tschannen et~al.(2025)Tschannen, Gritsenko, Wang, Naeem, Alabdulmohsin, Parthasarathy, Evans, Beyer, Xia, Mustafa, Hénaff, Harmsen, Steiner, and Zhai]{Tschannen:2025:SMV}
Michael Tschannen, Alexey Gritsenko, Xiao Wang, Muhammad~Ferjad Naeem, Ibrahim Alabdulmohsin, Nikhil Parthasarathy, Talfan Evans, Lucas Beyer, Ye~Xia, Basil Mustafa, Olivier Hénaff, Jeremiah Harmsen, Andreas Steiner, and Xiaohua Zhai.
\newblock Siglip 2: Multilingual vision-language encoders with improved semantic understanding, localization, and dense features.
\newblock \emph{arXiv:2502.14786 [cs.CV]}, 2025.

\bibitem[Tsipras et~al.(2019)Tsipras, Santurkar, Engstrom, Turner, and Madry]{Tsipras:2019:RMB}
Dimitris Tsipras, Shibani Santurkar, Logan Engstrom, Alexander Turner, and Aleksander Madry.
\newblock Robustness may be at odds with accuracy.
\newblock In \emph{ICLR}, 2019.

\bibitem[Tu et~al.(2023)Tu, Deng, and Gedeon]{Tu:2023:CLR}
Weijie Tu, Weijian Deng, and Tom Gedeon.
\newblock A closer look at the robustness of contrastive language-image pre-training {(CLIP)}.
\newblock In \emph{NeurIPS}, 2023.

\bibitem[Tu et~al.(2022)Tu, Talebi, Zhang, Yang, Milanfar, Bovik, and Li]{Tu:2022:MAV}
Zhengzhong Tu, Hossein Talebi, Han Zhang, Feng Yang, Peyman Milanfar, Alan~C. Bovik, and Yinxiao Li.
\newblock {MaxViT}: Multi-axis vision transformer.
\newblock In \emph{ECCV}, pp.\  459--479, 2022.

\bibitem[Vasu et~al.(2023)Vasu, Gabriel, Zhu, Tuzel, and Ranjan]{Vasu:2023:FAF}
Pavan Kumar~Anasosalu Vasu, James Gabriel, Jeff Zhu, Oncel Tuzel, and Anurag Ranjan.
\newblock {FastViT}: {A} fast hybrid vision transformer using structural reparameterization.
\newblock In \emph{ICCV}, pp.\  5762--5772, 2023.

\bibitem[Vasu et~al.(2024)Vasu, Pouransari, Faghri, Vemulapalli, and Tuzel]{Vase:2024:MFI}
Pavan Kumar~Anasosalu Vasu, Hadi Pouransari, Fartash Faghri, Raviteja Vemulapalli, and Oncel Tuzel.
\newblock {MobileCLIP}: Fast image-text models through multi-modal reinforced training.
\newblock In \emph{CVPR}, pp.\  15963--15974, 2024.

\bibitem[Verma \& Rubin(2018)Verma and Rubin]{Verma:2018:FDE}
Sahil Verma and Julia Rubin.
\newblock Fairness definitions explained.
\newblock In \emph{FairWare@ICSE}, pp.\  1--7, 2018.

\bibitem[Wang et~al.(2019)Wang, Ge, Lipton, and Xing]{Wang:2019:LRG}
Haohan Wang, Songwei Ge, Zachary Lipton, and Eric~P Xing.
\newblock Learning robust global representations by penalizing local predictive power.
\newblock In \emph{NeurIPS}, pp.\  10506--10518, 2019.

\bibitem[Wang et~al.(2020)Wang, Wu, Huang, and Xing]{Wang:2020:HFC}
Haohan Wang, Xindi Wu, Zeyi Huang, and Eric~P. Xing.
\newblock High-frequency component helps explain the generalization of convolutional neural networks.
\newblock In \emph{CVPR}, pp.\  8681--8691, 2020.

\bibitem[Wightman et~al.(2021)Wightman, Touvron, and J{\'{e}}gou]{Wightman:2021:RSB}
Ross Wightman, Hugo Touvron, and Herv{\'{e}} J{\'{e}}gou.
\newblock {ResNet} strikes back: An improved training procedure in timm.
\newblock \emph{arXiv:2110.00476 [cs.CV]}, 2021.

\bibitem[Woo et~al.(2023)Woo, Debnath, Hu, Chen, Liu, Kweon, and Xie]{Woo:2023:CCS}
Sanghyun Woo, Shoubhik Debnath, Ronghang Hu, Xinlei Chen, Zhuang Liu, In~So Kweon, and Saining Xie.
\newblock {ConvNeXt V2}: Co-designing and scaling convnets with masked autoencoders.
\newblock In \emph{CVPR}, pp.\  16133--16142, 2023.

\bibitem[Wortsman et~al.(2022)Wortsman, Ilharco, Gadre, Roelofs, Lopes, Morcos, Namkoong, Farhadi, Carmon, Kornblith, and Schmidt]{Wortsman:2022:MSA}
Mitchell Wortsman, Gabriel Ilharco, Samir~Yitzhak Gadre, Rebecca Roelofs, Raphael~Gontijo Lopes, Ari~S. Morcos, Hongseok Namkoong, Ali Farhadi, Yair Carmon, Simon Kornblith, and Ludwig Schmidt.
\newblock Model soups: averaging weights of multiple fine-tuned models improves accuracy without increasing inference time.
\newblock In \emph{ICML}, pp.\  23965--23998, 2022.

\bibitem[Wu et~al.(2022)Wu, Zhang, Peng, Liu, Xiao, Fu, and Yuan]{Wu:2022:TFP}
Kan Wu, Jinnian Zhang, Houwen Peng, Mengchen Liu, Bin Xiao, Jianlong Fu, and Lu~Yuan.
\newblock {TinyViT}: Fast pretraining distillation for small vision transformers.
\newblock In \emph{ECCV}, pp.\  68--85, 2022.

\bibitem[Xiao et~al.(2021)Xiao, Engstrom, Ilyas, and Madry]{Xiao:2021:NSR}
Kai~Yuanqing Xiao, Logan Engstrom, Andrew Ilyas, and Aleksander Madry.
\newblock Noise or signal: The role of image backgrounds in object recognition.
\newblock In \emph{ICLR}, 2021.

\bibitem[Xie et~al.(2020)Xie, Luong, Hovy, and Le]{Xie:2020:STW}
Qizhe Xie, Minh{-}Thang Luong, Eduard~H. Hovy, and Quoc~V. Le.
\newblock Self-training with noisy student improves imagenet classification.
\newblock In \emph{CVPR}, pp.\  10684--10695, 2020.

\bibitem[Xie et~al.(2017)Xie, Girshick, Doll{\'{a}}r, Tu, and He]{Xie:2017:ART}
Saining Xie, Ross~B. Girshick, Piotr Doll{\'{a}}r, Zhuowen Tu, and Kaiming He.
\newblock Aggregated residual transformations for deep neural networks.
\newblock In \emph{CVPR}, pp.\  5987--5995, 2017.

\bibitem[Xu et~al.(2021{\natexlab{a}})Xu, Liu, Li, Jain, and Tang]{Xu:2021:TBR}
Han Xu, Xiaorui Liu, Yaxin Li, Anil~K. Jain, and Jiliang Tang.
\newblock To be robust or to be fair: Towards fairness in adversarial training.
\newblock In \emph{ICML}, pp.\  11492--11501, 2021{\natexlab{a}}.

\bibitem[Xu et~al.(2024)Xu, Xie, Tan, Huang, Howes, Sharma, Li, Ghosh, Zettlemoyer, and Feichtenhofer]{Xu:2024:DCD}
Hu~Xu, Saining Xie, Xiaoqing~Ellen Tan, Po{-}Yao Huang, Russell Howes, Vasu Sharma, Shang{-}Wen Li, Gargi Ghosh, Luke Zettlemoyer, and Christoph Feichtenhofer.
\newblock Demystifying {CLIP} data.
\newblock In \emph{ICLR}, 2024.

\bibitem[Xu et~al.(2021{\natexlab{b}})Xu, Xu, Chang, and Tu]{Xu:2021:CSC}
Weijian Xu, Yifan Xu, Tyler~A. Chang, and Zhuowen Tu.
\newblock Co-scale {Conv}-attentional image transformers.
\newblock In \emph{ICCV}, pp.\  9961--9970, 2021{\natexlab{b}}.

\bibitem[Yalniz et~al.(2019)Yalniz, J{\'{e}}gou, Chen, Paluri, and Mahajan]{Yalniz:2019:BSS}
I.~Zeki Yalniz, Herv{\'{e}} J{\'{e}}gou, Kan Chen, Manohar Paluri, and Dhruv Mahajan.
\newblock Billion-scale semi-supervised learning for image classification.
\newblock \emph{arXiv:1905.00546 [cs.CV]}, 2019.

\bibitem[Yang et~al.(2020)Yang, Rashtchian, Zhang, Salakhutdinov, and Chaudhuri]{Yang:2020:ACL}
Yao{-}Yuan Yang, Cyrus Rashtchian, Hongyang Zhang, Ruslan Salakhutdinov, and Kamalika Chaudhuri.
\newblock A closer look at accuracy vs. robustness.
\newblock In \emph{NeurIPS}, 2020.

\bibitem[You et~al.(2017)You, Gitman, and Ginsburg]{You:2017:LBT}
Yang You, Igor Gitman, and Boris Ginsburg.
\newblock Large batch training of convolutional networks.
\newblock \emph{arXiv:1708.03888 [cs.CV]}, 2017.

\bibitem[Yu et~al.(2024)Yu, Zhou, Yan, and Wang]{Yu:2023:IWI}
Weihao Yu, Pan Zhou, Shuicheng Yan, and Xinchao Wang.
\newblock {InceptionNeXt}: When inception meets {ConvNeXt}.
\newblock In \emph{CVPR}, pp.\  5672--5683, 2024.

\bibitem[Yuan et~al.(2021)Yuan, Chen, Wang, Yu, Shi, Jiang, Tay, Feng, and Yan]{Yuan:2021:TTT}
Li~Yuan, Yunpeng Chen, Tao Wang, Weihao Yu, Yujun Shi, Zihang Jiang, Francis E.~H. Tay, Jiashi Feng, and Shuicheng Yan.
\newblock Tokens-to-token vit: Training vision transformers from scratch on imagenet.
\newblock In \emph{ICCV}, pp.\  538--547, 2021.

\bibitem[Yun et~al.(2019)Yun, Han, Chun, Oh, Yoo, and Choe]{Yun:2019:CRS}
Sangdoo Yun, Dongyoon Han, Sanghyuk Chun, Seong~Joon Oh, Youngjoon Yoo, and Junsuk Choe.
\newblock Cutmix: Regularization strategy to train strong classifiers with localizable features.
\newblock In \emph{ICCV}, pp.\  6022--6031, 2019.

\bibitem[Zagoruyko \& Komodakis(2016)Zagoruyko and Komodakis]{Zagoruyko:2016:WRN}
Sergey Zagoruyko and Nikos Komodakis.
\newblock Wide residual networks.
\newblock In \emph{BMVC}, 2016.

\bibitem[Zhai et~al.(2023)Zhai, Mustafa, Kolesnikov, and Beyer]{Zhai:2023:SLF}
Xiaohua Zhai, Basil Mustafa, Alexander Kolesnikov, and Lucas Beyer.
\newblock Sigmoid loss for language image pre-training.
\newblock In \emph{ICCV}, pp.\  11941--11952, 2023.

\bibitem[Zhu et~al.(2017)Zhu, Xie, and Yuille]{Zhu:2017:ORO}
Zhuotun Zhu, Lingxi Xie, and Alan~L. Yuille.
\newblock Object recognition with and without objects.
\newblock In \emph{{IJCAI}}, pp.\  3609--3615, 2017.

\bibitem[Zoph et~al.(2018)Zoph, Vasudevan, Shlens, and Le]{Zoph:2018:LTA}
Barret Zoph, Vijay Vasudevan, Jonathon Shlens, and Quoc~V. Le.
\newblock Learning transferable architectures for scalable image recognition.
\newblock In \emph{CVPR}, pp.\  8697--8710, 2018.

\end{thebibliography}
}
\newpage
\appendix

\section{Details on the considered quality dimensions}\label{appendix:sec:quality_dimensions}

In the following, we provide more details on the considered quality dimensions and the corresponding evaluation protocols. We let $f$ be the model of interest and use the ImageNet-1k evaluation split with $N$ images $\{x_n \mid n \in 1\ldots N\}$ belonging to one of $C$ classes $\{c_n\in 1\ldots C \mid n \in 1\ldots N\}$ for most protocols. %

\myparagraph{Accuracy.} To measure the predictive performance of the considered models, we let $[\cdot]$ denote the Iverson bracket~\citep{Knuth:1992:TNN} and report the ImageNet-1k top-1 accuracy
\begin{equation}
\operatorname{A} = \frac{1}{N} \sum_{n=1}^{N} \big[f(x_n) = c_n \big].
\label{eq:app:acc}
\end{equation}

\myparagraph{Adversarial robustness.} To assess the adversarial robustness of a model, we use the two popular attacks, FGSM~\citep{Goodfellow:2015:EHA} and PGD~\citep{Madry:2018:TDL}. In the Fast Gradient Sign Method (FGSM), adversarial examples are generated by computing
the sign of the gradient of the cross-entropy loss $\mathcal{L}$ with respect to the original input, scaling it with a small factor $\epsilon$, and adding the result to the original image (see \cref{fig:adversarial_attack}). Formally, we obtain the FGSM accuracy ($\operatorname{FGSM-A}$) via
\begin{equation}
    \operatorname{FGSM-A} = \frac{1}{N} \sum_{n=1}^{N} \big[f(\hat{x}_n) = c_n \big],
\end{equation}
with $\hat{x}_n = x + \epsilon \cdot \text{sign}(\nabla_x \mathcal{L})$.
Projected Gradient Descent (PGD) extends FGSM by applying it repeatedly, yielding
\begin{equation}
    \operatorname{PGD-A} = \frac{1}{N} \sum_{n=1}^{N} \Big[f(\hat{x}_n^{(I)}) = c_n \Big],
\end{equation}
with $\hat{x}_n^{(i+1)} = \hat{x}_n^{(i)} + \epsilon \cdot \text{sign}\big(\nabla_{\hat{x}_n^{(i)}} \mathcal{L}\big) \; \text{and}\; \hat{x}^{(0)} = x$.
We use $\epsilon=\sfrac{8}{255}$ and $I=10$~\citep{Kim:2020:TPR}. To reduce the dependence on the clean accuracy of the model, we report adversarial robustness relative to the accuracy $\operatorname{A}$ from \cref{eq:app:acc}. We combine the results from the two attacks using their geometric mean ($\operatorname{GM}$), resulting in the final adversarial robustness 
\begin{equation}
    \operatorname{AR} = \operatorname{GM}\left(\frac{\operatorname{FGSM-A}}{\operatorname{A}}, \frac{\operatorname{PGD-A}}{\operatorname{A}}\right).
\end{equation}

\myparagraph{Corruption robustness.} To assess a model's robustness to common corruptions ($\operatorname{CR}$) like JPEG compression or contrast changes (see \cref{fig:imagenet_c}), we measure the accuracy on ImageNet-C~\citep{Hendrycks:2019:BNN}, \textit{i.e.}, the ImageNet evaluation split with different corruption types of increasing strength. We here follow \citet{Hendrycks:2019:BNN} and use the standard mean instead of the geometric mean to summarize the results for different corruption types and strengths. To normalize the C-robustness and to be consistent with our other robustness metrics, we deviate from \citet{Hendrycks:2019:BNN} and again report the top-1 accuracy on the corrupted data ($\operatorname{A}_\text{Corr}$) relative to the clean ImageNet-1k accuracy ($\operatorname{A}$), yielding
\begin{equation}
\operatorname{CR} = \frac{\operatorname{A}_\text{Corr}}{\operatorname{A}}.    
\end{equation}

\begin{figure}[t]
    \centering
    \input{figures/quality_dimensions/adv_rob}    
    \caption{\textit{Illustration of an adversarial attack.}}
    \label{fig:adversarial_attack}
\end{figure}

\begin{figure}[t]
    \centering
    \input{figures/quality_dimensions/c_rob}    
    \caption{\textit{Example corruptions from ImageNet-C~\citep{Hendrycks:2019:BNN}.}}
    \label{fig:imagenet_c}
\end{figure}

\myparagraph{OOD robustness.}
To measure the out-of-domain robustness of a model, we report the geometric mean of the relative accuracy (normalized by $\operatorname{A}$) on five out-of-domain datasets. Specifically, we use ImageNet-R~\citep{Hendrycks:2021:MFR}, ImageNet-Sketch~\citep{Wang:2019:LRG}, as well as Stylized-ImageNet, Edge, and Silhouette from \citet{Geirhos:2019:ITC}. For ImageNet-Sketch and Stylized-ImageNet, we use the versions also used in \citet{Geirhos:2021:PSI}. Please refer to \cref{fig:ood} for example images of the different datasets.

\begin{figure}[t]
    \centering
    \input{figures/quality_dimensions/ood_rob}
    \caption{\textit{Example images from the considered OOD datasets~\citep{Hendrycks:2021:MFR, Wang:2019:LRG, Geirhos:2019:ITC}.}}
    \label{fig:ood}
\end{figure}

\myparagraph{Calibration error.} Calibration means that the output confidence of a model faithfully reflects the probability of the prediction being correct. We use two established metrics for measuring the calibration error ($\operatorname{CE}$). The expected calibration error ($\operatorname{ECE}$)~\citep{Nixon:2019:MCD, Guo:2017:OCM} divides the predictions into $B$ bins $b$ based on the output confidence of the model and compares how well the confidences $\text{conf}(b)$ of the predictions in that bin are aligned with the accuracy $\operatorname{A}_b$ of the predictions in that bin:
\begin{equation}
\operatorname{ECE} = \sum_{b=1}^{B} \frac{n_b}{N} \left| \operatorname{A}_b - \text{conf}(b) \right|\,,
\end{equation}
with $n_b$ denoting the number of predictions in bin $b$. Since a common criticism of the $\operatorname{ECE}$ is the use of a fixed bin range, we additionally report the adaptive calibration error ($\operatorname{ACE}$)~\citep{Nixon:2019:MCD} that measures the discrepancy between $\operatorname{A}_{r, c}$ and $\text{conf}(r, c)$, \textit{i.e.}, the accuracy and confidence of images in the adaptive calibration range $r$ for class label $c$:
\begin{equation}
\operatorname{ACE} = \frac{1}{CR}\sum_{c=1}^{C} \sum_{r=1}^{R} \left| \operatorname{A}_{r, c} - \text{conf}(r, c) \right|.
\end{equation}
As in \citet{Nixon:2019:MCD, Guo:2017:OCM}, we use $15$ bins for both protocols and again report the geometric mean ($\operatorname{GM}$) of both errors, \ie,
\begin{equation}
    \operatorname{CE} = \operatorname{GM}(\operatorname{ECE}, \operatorname{ACE}).
\end{equation}

\myparagraph{Class balance.} We consider a model fair if none of the classes is classified less well than the others~\citep{Benz:2020:RMB}. We evaluate the \updated{class balance of accuracies} ($\operatorname{F}_\text{Acc}$) of a model similar to \citet{Croce:2021:RBS} and subtract the standard deviation of ImageNet-1k class accuracies from 1 (this ensures that higher scores indicate a higher class balance; the standard deviation cannot exceed 1): 
\begin{equation}
\operatorname{F}_\text{Acc} = 1-\sqrt {\frac{1}{C} \sum_{c=1}^{C} (A_c - A)^2}\, ,
\end{equation}
with $A_c$ denoting the accuracy for images of class $c$. %
Intuitively, a high value indicates that the accuracies of each class are similar, and thus, the model behaves fairly, as illustrated in \cref{fig:fairness}. Similar to \citet{Kuzucu:2024:UFM}, we also consider a model fair if the average confidence (target softmax outputs) for each class is balanced. We compute the \updated{class balance of confidences} ($\operatorname{F}_\text{Conf}$) of a model by subtracting the standard deviation of ImageNet-1k average class confidences from 1:
\begin{equation}
\operatorname{F}_\text{Conf} = 1-\sqrt {\frac{1}{C} \sum_{c=1}^{C} (\text{Conf}_c - \text{Conf})^2}\, ,
\end{equation}
with $\text{Conf}_c$ denoting the average confidence for images of class $c$ and $\text{Conf}$ denoting the average confidence for all images. The final class balance score (F) is the geometric mean (GM) of $\operatorname{F}_\text{Acc}$ and $\operatorname{F}_\text{Conf}$, \ie,
\begin{equation}
    \operatorname{F} = \operatorname{GM}(\operatorname{F}_\text{Acc}, \operatorname{F}_\text{Conf}).
\end{equation}

\begin{figure}[t]
    \centering
        \begin{tikzpicture}[every node/.style={font=\sffamily}]
    \sffamily
    \scriptsize
\def\yscale{0.0001}
\def\ticklabely{0.00cm};
\def\setuponey{0.4};
\def\setuptwoy{0.0};

\def\myxmin{0};
\def\myxmax{4};
\def\myymin{0.0};
\def\myymax{1.1};

\def\centergroupone{0.5};
\def\centergrouptwo{1.5};
\def\centergroupthree{2.5};
\def\centergroupfour{3.5};
\def\centergroupfive{4.5};
\def\centergroupsix{5.5};
\def\centergroupseven{6.5};

\def\bardistance{0.3};
\def\barwidth{0.2};
\def\mymarkeroffsety{0.05}
    
\begin{axis}[
    axis lines=middle,
    xmin=\myxmin-0.6, xmax=\myxmax,
    ymin=\myymin-0.2, ymax=\myymax, %
    xtick=\empty, ytick=\empty,
    axis line style={draw=none},
    tick style={draw=none},
    height=5cm, 
    width=9.5cm,
]

\draw[->] (\myxmin,\myymin) -- (\myxmin,\myymax);
\draw[-] (\myxmin,0.5) -- (\myxmin,0.5) node[xshift=-0.9cm, anchor=center, rotate=90] {\scriptsize Accuracy};
\draw[-] (\myxmin,\myymin) -- (\myxmax,\myymin);

\foreach \yValue in {0,0.2, 0.4, 0.6, 0.8,1} {
    \edef\temp{\noexpand\draw [black] (\myxmin,\yValue) -- (\myxmin-0.1,\yValue);}
    \temp
}
\draw[-] (\myxmin,0.0) -- (\myxmin-0.1,0.00) node[left=\ticklabely] {\scriptsize 0};
\draw[-] (\myxmin,0.20) -- (\myxmin-0.1,0.20) node[left=\ticklabely] {\scriptsize 0.2};
\draw[-] (\myxmin,0.40) -- (\myxmin-0.1,0.40) node[left=\ticklabely] {\scriptsize 0.4};
\draw[-] (\myxmin,0.60) -- (\myxmin-0.1,0.60) node[left=\ticklabely] {\scriptsize 0.6};
\draw[-] (\myxmin,0.80) -- (\myxmin-0.1,0.8) node[left=\ticklabely] {\scriptsize 0.8};
\draw[-] (\myxmin,1.0) -- (\myxmin-0.1,1.0) node[left=\ticklabely] {\scriptsize 1};

\draw[-] ([yshift=-0.2cm]+1.0,\myymin) -- ([yshift=0.cm]+1.0,\myymin);
\draw[-] ([yshift=-0.2cm]+2.0,\myymin) -- ([yshift=0.0cm]+2.0,\myymin);
\draw[-] ([yshift=-0.2cm]+3.0,\myymin) -- ([yshift=0.0cm]+3.0,\myymin);

\draw[-] (\centergroupone,\myymin) -- (\centergroupone,\myymin) node[below=\ticklabely] {\scriptsize \makecell{Class 1}};
\draw[-] (\centergrouptwo,\myymin) -- (\centergrouptwo,\myymin) node[below=\ticklabely] {\scriptsize \makecell{Class 2}};
\draw[-] (\centergroupthree,\myymin) -- (\centergroupthree,\myymin) node[below=\ticklabely] {\scriptsize \makecell{Class 3}};
\draw[-] (\centergroupfour,\myymin) -- (\centergroupfour,\myymin) node[below=\ticklabely] {\scriptsize \makecell{Class 4}};

\draw[fill=cnncolor, draw=cnncolor!100] (\centergroupone-0.5*\bardistance-\barwidth/2,\myymin) rectangle (\centergroupone-0.5*\bardistance+\barwidth/2,0.8);

\draw[fill=transformercolor, draw=transformercolor!100] (\centergroupone+0.5*\bardistance-\barwidth/2,\myymin) rectangle (\centergroupone+0.5*\bardistance+\barwidth/2,0.6);

\draw[fill=cnncolor, draw=cnncolor!100] (\centergrouptwo-0.5*\bardistance-\barwidth/2,\myymin) rectangle (\centergrouptwo-0.5*\bardistance+\barwidth/2,0.4);

\draw[fill=transformercolor, draw=transformercolor!100] (\centergrouptwo+0.5*\bardistance-\barwidth/2,\myymin) rectangle (\centergrouptwo+0.5*\bardistance+\barwidth/2,0.6);

\draw[fill=cnncolor, draw=cnncolor!100] (\centergroupthree-0.5*\bardistance-\barwidth/2,\myymin) rectangle (\centergroupthree-0.5*\bardistance+\barwidth/2,0.7);

\draw[fill=transformercolor, draw=transformercolor!100] (\centergroupthree+0.5*\bardistance-\barwidth/2,\myymin) rectangle (\centergroupthree+0.5*\bardistance+\barwidth/2,0.6);

\draw[fill=cnncolor, draw=cnncolor!100] (\centergroupfour-0.5*\bardistance-\barwidth/2,\myymin) rectangle (\centergroupfour-0.5*\bardistance+\barwidth/2,0.5);

\draw[fill=transformercolor, draw=transformercolor!100] (\centergroupfour+0.5*\bardistance-\barwidth/2,\myymin) rectangle (\centergroupfour+0.5*\bardistance+\barwidth/2,0.6);

\draw[fill=transformercolor!20, draw=transformercolor!100] (3,0.8) rectangle (\myxmax,1.1);

\draw[fill=cnncolor, draw=cnncolor!100] (3.1-0.025,1-0.013) rectangle (3.1+0.025,1+0.013);

\node[anchor=west] at (3.1+0.01,1) {\scriptsize Unfair model};

\draw[fill=transformercolor, draw=transformercolor!100] (3.1-0.025,0.9-0.013) rectangle (3.1+0.025,0.9+0.013);

\node[anchor=west] at (3.1+0.01,0.9) {\scriptsize Fair model};

\draw[-] (\myxmin,\myymin) -- (\myxmax,\myymin);

\end{axis}

    \end{tikzpicture}
    \caption{\textit{Visual illustration of the class accuracies of a fair model \vsiccv an unfair one, both with equal average accuracy.}}
    \label{fig:fairness}
\end{figure}

\myparagraph{Object focus.} To compute the object focus ($\operatorname{OF}$), we first compute the background focus $\operatorname{BF} = \operatorname{A}_{\text{\textsc{Mixed-Same}}} - \operatorname{A}_{\text{\textsc{Mixed-Rand}}}$~\citep{Xiao:2021:NSR}. \textsc{Mixed-Rand} is a dataset where image backgrounds are substituted with backgrounds from random classes and, therefore, contain no class information (see \cref{fig:object_focus}). \textsc{Mixed-Same} is a dataset where image backgrounds are substituted with backgrounds from the same class to account for editing artifacts in \textsc{Mixed-Same}. Intuitively, we measure the drop in accuracy when changing the image background with the background from another class to assess if the model focuses on background signals. Next, we compute the inverse of the background focus to obtain the object focus $\operatorname{OF} = 1 - \operatorname{BF}$. 

\begin{figure}[t]
    \centering
    \input{figures/quality_dimensions/object_focus}
    \caption{\textit{Example images from \citet{Xiao:2021:NSR} to estimate the object focus.}}
    \label{fig:object_focus}
\end{figure}

\myparagraph{Shape bias.} \citet{Geirhos:2019:ITC} showed that ImageNet-trained CNNs exhibit a strong texture bias, meaning that decisions are formed on the basis of texture information rather than shape information. As a stronger shape bias is said to be advantageous for robustness and more in line with how humans form decisions, we follow \citet{Geirhos:2019:ITC} and report the shape bias ($\operatorname{SB}$) as follows:
\begin{equation}
\operatorname{SB} = \frac{ \sum_{n=1}^{N} \big[f(\tilde{x}_n) = c_n^{\text{(shape)}} \big]\ }{\sum_{m=1}^{N} \left( \big[f(\tilde{x}_m) = c_m^{\text{(shape)}} \big] + \big[f(\tilde{x}_m) = c_m^{\text{(texture)}} \big]\right)},
\end{equation}
with $\tilde{x}_n$ being synthetically generated images with a texture-shape cue conflict, \textit{i.e.}, where the shape is from one class $c_n^{\text{(shape)}}$ and the texture is from another class $c_n^{\text{(texture)}}$, \eg, as in \cref{fig:shape_bias}.

\begin{figure}[t]
    \centering
    \input{figures/quality_dimensions/shape_bias}
    \caption{\textit{Example images from \citet{Geirhos:2019:ITC} to estimate the shape bias.}}
    \label{fig:shape_bias}
\end{figure}

\myparagraph{Parameters.} %
As memory efficiency and inference time depend highly on the implementation and hardware used, impeding future comparisons, we report the number of parameters as a proxy. %

\section{Interactive scatter plot}\label{sec:scatter_interactive}

\begin{figure*}[t]
\centering
\frame{\includegraphics[width=0.75\linewidth]{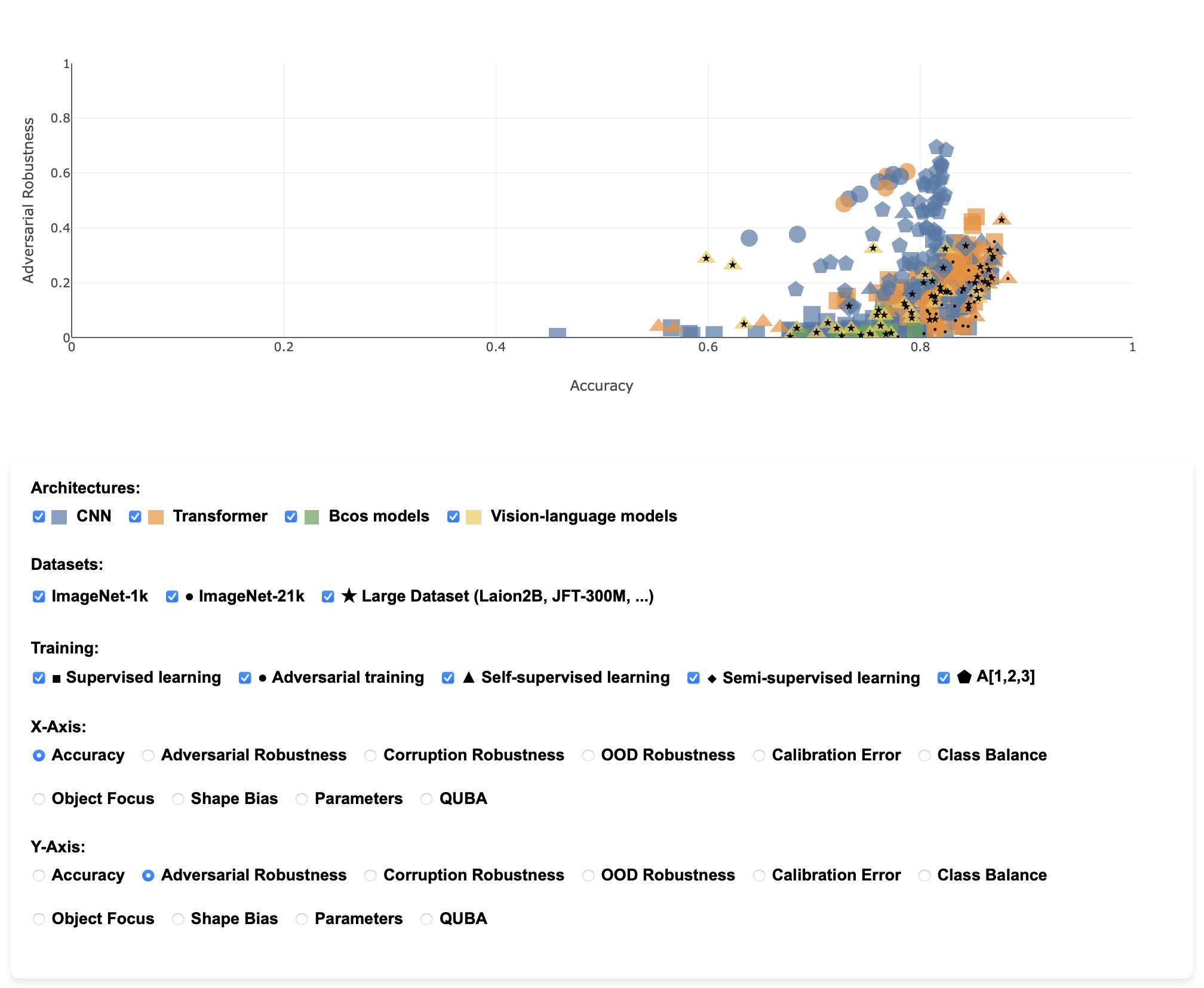}}
\caption{\textit{Preview of our interactive plot.} It can be found in the supplement under \texttt{interactive\_plot.html} or on our project page.}
\label{fig:interactive_plot}
\end{figure*}

In \cref{fig:scatter_dimensions_acc} of the main paper, we only plot a representative subset of models to reduce clutter. In the supplement, we additionally include an interactive plot, called \texttt{interactive\_plot.html}, that can be opened with standard browsers. It includes all \nrmodels models and allows the filtering of the models based on the training dataset, training paradigm, and architecture. Also, different quality dimensions can be chosen for the $x$ and $y$-axis to visualize different relationships. Hovering the cursor over a marker reveals a tooltip displaying the model's name and scores for the considered quality dimensions, offering detailed performance insights at a glance. A preview of the interactive plot is included in \cref{fig:interactive_plot}.  

\section{Additional experiments}\label{sec:add_exp}

\subsection{Relationships in model subgroups}\label{appendix:sec:subgroups}

During our analysis, we noticed that the rank correlation matrices for certain subgroups of the models can change compared to the correlation matrix of all models shown in \cref{fig:correlation_matrix} of the main paper. This has two important implications: First, it is crucial to look at as many models as possible to make general statements, which shows that our confirmation of (or contradiction to) existing results with a large model zoo is a valuable contribution. Second, general statements should be taken cautiously, as they do not necessarily apply to all model groups. We thus continue our analysis by studying the correlation matrices for different model subgroups. In \cref{fig:corr_matrix_cnn_transformer}, we plot the correlation matrices for all CNNs and Transformers, respectively. Overall, both matrices are similar to the matrix for all models (\cref{fig:correlation_matrix} of the main paper). 
However, for CNNs, C-robustness and OOD robustness are negatively correlated with adversarial robustness, while the opposite holds for Transformers. 
We observe that the calibration error is positively correlated with accuracy for CNNs and negatively for Transformers (however, both correlations are not statistically significant). This also aligns with another large-scale study that found that the calibration error is decreasing with more accuracy for recent state-of-the-art Transformers~\citep{Minderer:2021:RCM}.
Notably, accuracy and shape bias are strongly correlated for CNN-based models, while they exhibit only a weak correlation for Transformer models. %
Moreover, for Transformers, an increased number of parameters is quite strongly correlated with desirable properties of all other quality dimensions, which is less pronounced for CNNs. %

\begin{figure*}[t]
\centering
\begin{subfigure}{0.33\linewidth}
    \hspace{0pt}
    \centering
    \begin{tikzpicture}[every node/.style={font=\sffamily}]
\sffamily
\scriptsize

\def\yticklabeloffset{-0.55};
\def\xticklabeloffset{-0.6};
\def\xticklabeloffsetx{-0.1};
\def\xticklabelrotation{60};

\begin{axis}[%
    width=1.00\linewidth,
    axis equal image, %
    axis line style = {line width=.5pt,draw=gray!50},
    scatter, %
    colormap name=correlation_cm, %
    point meta min=-1,
    point meta max=1,
    clip=false,
    grid=minor, %
    minor grid style={line width=.5pt,draw=gray!50},
    minor tick num=1, %
    minor tick length=0pt, 
    tickwidth=0pt, %
    ticks=none, %
    try min ticks=10, %
    y dir=reverse, %
    colorbar style={
        at={(1.05,0.5)},anchor=west,width=0.05*\pgfkeysvalueof{/pgfplots/parent axis width}},
    xticklabel pos=right, %
    enlargelimits={abs=0.5}, %
    scatter/@pre marker code/.append code={%
      \pgfplotstransformcoordinatex{sqrt(abs(\pgfplotspointmeta))}%
      \scope[mark size=\pgfplotsunitxlength*\pgfmathresult/3.7 + \pgfplotsunitxlength*20.7/2, fill=mapped color]
    },
    scatter/@post marker code/.append code={%
      \endscope%
    }
    ]

\addplot +[
    point meta=explicit, %
    only marks, %
    every node near coord/.append style={font=\small, color=white,anchor=center},
    visualization depends on={value \thisrow{label} \as \Label}, %
    visualization depends on={value \thisrow{size} \as \Size}, %
    ] table [
    x expr={int(mod(\coordindex+0.01,9))}, %
    y expr={int((\coordindex+0.01)/9))},
    meta=value,
] {
X   Y   value label size
0 0 1.0 $+$ 0.1em
0 0 0 $$ 0.1em
0 0 0 $$ 0.1em
0 0 0 $$ 0.1em
0 0 0 $$ 0.1em
0 0 0 $$ 0.1em
0 0 0 $$ 0.1em
0 0 0 $$ 0.1em
0 0 0 $$ 0.1em
 
0 0 0.4016 $+$ 0.1em
0 0 1.0 $+$ 0.1em
0 0 0 $$ 0.1em
0 0 0 $$ 0.1em
0 0 0 $$ 0.1em
0 0 0 $$ 0.1em
0 0 0 $$ 0.1em
0 0 0 $$ 0.1em
0 0 0 $$ 0.1em
 
0 0 0.3824 $+$ 0.1em
0 0 -0.3094 $-$ 0.1em
0 0 1.0 $+$ 0.1em
0 0 0 $$ 0.1em
0 0 0 $$ 0.1em
0 0 0 $$ 0.1em
0 0 0 $$ 0.1em
0 0 0 $$ 0.1em
0 0 0 $$ 0.1em
 
0 0 0.2225 $+$ 0.1em
0 0 -0.2831 $-$ 0.1em
0 0 0.8394 $+$ 0.1em
0 0 1.0 $+$ 0.1em
0 0 0 $$ 0.1em
0 0 0 $$ 0.1em
0 0 0 $$ 0.1em
0 0 0 $$ 0.1em
0 0 0 $$ 0.1em
 
0 0 0.1364 $+$ 0.1em
0 0 0.3447 $+$ 0.1em
0 0 0.0343 $+$ 0.1em
0 0 -0.0244 $-$ 0.1em
0 0 1.0 $+$ 0.1em
0 0 0 $$ 0.1em
0 0 0 $$ 0.1em
0 0 0 $$ 0.1em
0 0 0 $$ 0.1em
 
0 0 0.6505 $+$ 0.1em
0 0 0.0331 $+$ 0.1em
0 0 0.6758 $+$ 0.1em
0 0 0.5466 $+$ 0.1em
0 0 0.3998 $+$ 0.1em
0 0 1.0 $+$ 0.1em
0 0 0 $$ 0.1em
0 0 0 $$ 0.1em
0 0 0 $$ 0.1em
 
0 0 0.7389 $+$ 0.1em
0 0 0.4831 $+$ 0.1em
0 0 0.3732 $+$ 0.1em
0 0 0.2248 $+$ 0.1em
0 0 0.0921 $+$ 0.1em
0 0 0.5223 $+$ 0.1em
0 0 1.0 $+$ 0.1em
0 0 0 $$ 0.1em
0 0 0 $$ 0.1em
 
0 0 0.4431 $+$ 0.1em
0 0 0.3072 $+$ 0.1em
0 0 0.5013 $+$ 0.1em
0 0 0.4165 $+$ 0.1em
0 0 0.1177 $+$ 0.1em
0 0 0.4181 $+$ 0.1em
0 0 0.5799 $+$ 0.1em
0 0 1.0 $+$ 0.1em
0 0 0 $$ 0.1em
 
0 0 0.4207 $+$ 0.1em
0 0 0.2732 $+$ 0.1em
0 0 0.3091 $+$ 0.1em
0 0 0.1704 $+$ 0.1em
0 0 -0.0689 $-$ 0.1em
0 0 0.3012 $+$ 0.1em
0 0 0.5135 $+$ 0.1em
0 0 0.2882 $+$ 0.1em
0 0 1.0 $+$ 0.1em
 
};

\addplot[
    mark=x,only marks, mark size=6pt, thick, gray!50,
    point meta =explicit symbolic,
    nodes near coords,
]
table[x=x,y=y, meta=label]{
    x   y   label
    0 4 {} 
    2 4 {} 
    3 4 {} 
    1 5 {} 
    4 6 {} 
    4 7 {} 
    4 8 {}

};

\node[anchor=east] at (axis cs:0+\yticklabeloffset,0) {Accuracy\strut};
\node[anchor=east] at (axis cs:0+\yticklabeloffset,1) {Adv. Rob.\strut};
\node[anchor=east] at (axis cs:0+\yticklabeloffset,2) {C-Rob.\strut};
\node[anchor=east] at (axis cs:0+\yticklabeloffset,3) {OOD Rob.\strut};
\node[anchor=east] at (axis cs:0+\yticklabeloffset,4) {Cal. Error\strut};
\node[anchor=east] at (axis cs:0+\yticklabeloffset,5) {Class Balance\strut};
\node[anchor=east] at (axis cs:0+\yticklabeloffset,6) {Obj. Focus\strut};
\node[anchor=east] at (axis cs:0+\yticklabeloffset,7) {Shape Bias\strut};
\node[anchor=east] at (axis cs:0+\yticklabeloffset,8) {Parameters\strut};

\node[anchor=west, rotate=\xticklabelrotation] at (axis cs:0+\xticklabeloffsetx,0+\xticklabeloffset) {Accuracy\strut};
\node[anchor=west, rotate=\xticklabelrotation] at (axis cs:1+\xticklabeloffsetx,0+\xticklabeloffset) {Adv. Rob. \strut};
\node[anchor=west, rotate=\xticklabelrotation] at (axis cs:2+\xticklabeloffsetx,0+\xticklabeloffset) {C-Rob.\strut};
\node[anchor=west, rotate=\xticklabelrotation] at (axis cs:3+\xticklabeloffsetx,0+\xticklabeloffset) {OOD Rob.\strut};
\node[anchor=west, rotate=\xticklabelrotation] at (axis cs:4+\xticklabeloffsetx,0+\xticklabeloffset) {Cal. Error\strut};
\node[anchor=west, rotate=\xticklabelrotation] at (axis cs:5+\xticklabeloffsetx,0+\xticklabeloffset) {Class Balance\strut};
\node[anchor=west, rotate=\xticklabelrotation] at (axis cs:6+\xticklabeloffsetx,0+\xticklabeloffset) {Obj. Focus\strut};
\node[anchor=west, rotate=\xticklabelrotation] at (axis cs:7+\xticklabeloffsetx,0+\xticklabeloffset) {Shape Bias\strut};
\node[anchor=west, rotate=\xticklabelrotation] at (axis cs:8+\xticklabeloffsetx,0+\xticklabeloffset) {Parameters\strut};

\node[anchor=center] at (axis cs:4,9.5) {\textbf{CNN (n=170)}\strut};

\end{axis}
\end{tikzpicture}%
\end{subfigure}%
\begin{subfigure}{0.33\linewidth}
    \hspace{0pt}
     \centering
     \begin{tikzpicture}[every node/.style={font=\sffamily}]
\sffamily
\scriptsize

\def\yticklabeloffset{-0.55};
\def\xticklabeloffset{-0.6};
\def\xticklabeloffsetx{-0.1};
\def\xticklabelrotation{60};

\begin{axis}[%
    width=1.00\linewidth,
    axis equal image, %
    axis line style = {line width=.5pt,draw=gray!50},
    scatter, %
    colormap name=correlation_cm, %
    point meta min=-1,
    point meta max=1,
    clip=false,
    grid=minor, %
    minor grid style={line width=.5pt,draw=gray!50},
    minor tick num=1, %
    minor tick length=0pt, 
    tickwidth=0pt, %
    ticks=none, %
    try min ticks=10, %
    y dir=reverse, %
    colorbar style={
        at={(1.05,0.5)},anchor=west,width=0.05*\pgfkeysvalueof{/pgfplots/parent axis width}},
    xticklabel pos=right, %
    enlargelimits={abs=0.5}, %
    scatter/@pre marker code/.append code={%
      \pgfplotstransformcoordinatex{sqrt(abs(\pgfplotspointmeta))}%
      \scope[mark size=\pgfplotsunitxlength*\pgfmathresult/3.7 + \pgfplotsunitxlength*20.7/2, fill=mapped color]
    },
    scatter/@post marker code/.append code={%
      \endscope%
    }
    ]

\addplot +[
    point meta=explicit, %
    only marks, %
    every node near coord/.append style={font=\small, color=white,anchor=center},
    visualization depends on={value \thisrow{label} \as \Label}, %
    visualization depends on={value \thisrow{size} \as \Size}, %
    ] table [
    x expr={int(mod(\coordindex+0.01,9))}, %
    y expr={int((\coordindex+0.01)/9))},
    meta=value,
] {
X   Y   value label size
0 0 1.0 $+$ 0.1em
0 0 0 $$ 0.1em
0 0 0 $$ 0.1em
0 0 0 $$ 0.1em
0 0 0 $$ 0.1em
0 0 0 $$ 0.1em
0 0 0 $$ 0.1em
0 0 0 $$ 0.1em
0 0 0 $$ 0.1em
 
0 0 0.5297 $+$ 0.1em
0 0 1.0 $+$ 0.1em
0 0 0 $$ 0.1em
0 0 0 $$ 0.1em
0 0 0 $$ 0.1em
0 0 0 $$ 0.1em
0 0 0 $$ 0.1em
0 0 0 $$ 0.1em
0 0 0 $$ 0.1em
 
0 0 0.1375 $+$ 0.1em
0 0 -0.3312 $-$ 0.1em
0 0 1.0 $+$ 0.1em
0 0 0 $$ 0.1em
0 0 0 $$ 0.1em
0 0 0 $$ 0.1em
0 0 0 $$ 0.1em
0 0 0 $$ 0.1em
0 0 0 $$ 0.1em
 
0 0 0.0439 $+$ 0.1em
0 0 -0.3259 $-$ 0.1em
0 0 0.9051 $+$ 0.1em
0 0 1.0 $+$ 0.1em
0 0 0 $$ 0.1em
0 0 0 $$ 0.1em
0 0 0 $$ 0.1em
0 0 0 $$ 0.1em
0 0 0 $$ 0.1em
 
0 0 0.1913 $+$ 0.1em
0 0 0.3388 $+$ 0.1em
0 0 0.1007 $+$ 0.1em
0 0 0.0904 $+$ 0.1em
0 0 1.0 $+$ 0.1em
0 0 0 $$ 0.1em
0 0 0 $$ 0.1em
0 0 0 $$ 0.1em
0 0 0 $$ 0.1em
 
0 0 0.5233 $+$ 0.1em
0 0 0.0788 $+$ 0.1em
0 0 0.5986 $+$ 0.1em
0 0 0.5272 $+$ 0.1em
0 0 0.448 $+$ 0.1em
0 0 1.0 $+$ 0.1em
0 0 0 $$ 0.1em
0 0 0 $$ 0.1em
0 0 0 $$ 0.1em
 
0 0 0.7142 $+$ 0.1em
0 0 0.5606 $+$ 0.1em
0 0 0.2385 $+$ 0.1em
0 0 0.1566 $+$ 0.1em
0 0 0.1371 $+$ 0.1em
0 0 0.4717 $+$ 0.1em
0 0 1.0 $+$ 0.1em
0 0 0 $$ 0.1em
0 0 0 $$ 0.1em
 
0 0 0.342 $+$ 0.1em
0 0 0.3418 $+$ 0.1em
0 0 0.425 $+$ 0.1em
0 0 0.4525 $+$ 0.1em
0 0 0.0967 $+$ 0.1em
0 0 0.3147 $+$ 0.1em
0 0 0.5119 $+$ 0.1em
0 0 1.0 $+$ 0.1em
0 0 0 $$ 0.1em
 
0 0 0.3932 $+$ 0.1em
0 0 0.2966 $+$ 0.1em
0 0 0.2901 $+$ 0.1em
0 0 0.1653 $+$ 0.1em
0 0 -0.004 $-$ 0.1em
0 0 0.2992 $+$ 0.1em
0 0 0.4646 $+$ 0.1em
0 0 0.2405 $+$ 0.1em
0 0 1.0 $+$ 0.1em
 
};

\addplot[
    mark=x,only marks, mark size=6pt, thick, gray!50,
    point meta =explicit symbolic,
    nodes near coords,
]
table[x=x,y=y, meta=label]{
    x   y   label
    0 2 {} 
    0 3 {} 
    2 4 {} 
    3 4 {} 
    1 5 {} 
    3 6 {} 
    4 6 {} 
    4 7 {} 
    3 8 {} 
    4 8 {}

};

\node[anchor=west, rotate=\xticklabelrotation] at (axis cs:0+\xticklabeloffsetx,0+\xticklabeloffset) {Accuracy\strut};
\node[anchor=west, rotate=\xticklabelrotation] at (axis cs:1+\xticklabeloffsetx,0+\xticklabeloffset) {Adv. Rob. \strut};
\node[anchor=west, rotate=\xticklabelrotation] at (axis cs:2+\xticklabeloffsetx,0+\xticklabeloffset) {C-Rob.\strut};
\node[anchor=west, rotate=\xticklabelrotation] at (axis cs:3+\xticklabeloffsetx,0+\xticklabeloffset) {OOD Rob.\strut};
\node[anchor=west, rotate=\xticklabelrotation] at (axis cs:4+\xticklabeloffsetx,0+\xticklabeloffset) {Cal. Error\strut};
\node[anchor=west, rotate=\xticklabelrotation] at (axis cs:5+\xticklabeloffsetx,0+\xticklabeloffset) {Class Balance\strut};
\node[anchor=west, rotate=\xticklabelrotation] at (axis cs:6+\xticklabeloffsetx,0+\xticklabeloffset) {Obj. Focus\strut};
\node[anchor=west, rotate=\xticklabelrotation] at (axis cs:7+\xticklabeloffsetx,0+\xticklabeloffset) {Shape Bias\strut};
\node[anchor=west, rotate=\xticklabelrotation] at (axis cs:8+\xticklabeloffsetx,0+\xticklabeloffset) {Parameters\strut};

\node[anchor=center] at (axis cs:4,9.5) {\textbf{CNN IN1k (n=140)}\strut};

\end{axis}
\end{tikzpicture}
 \end{subfigure}%
\begin{subfigure}{0.33\linewidth}
    \hspace{0pt}
     \centering
     \hspace{-2em}\begin{tikzpicture}[every node/.style={font=\sffamily}]
\sffamily
\scriptsize

\def\yticklabeloffset{-0.55};
\def\xticklabeloffset{-0.6};
\def\xticklabeloffsetx{-0.1};
\def\xticklabelrotation{60};

\begin{axis}[%
    width=1.00\linewidth,
    axis equal image, %
    axis line style = {line width=.5pt,draw=gray!50},
    scatter, %
    colormap name=correlation_cm, %
    colorbar, %
    point meta min=-1,
    point meta max=1,
    clip=false,
    grid=minor, %
    minor grid style={line width=.5pt,draw=gray!50},
    minor tick num=1, %
    minor tick length=0pt, 
    tickwidth=0pt, %
    ticks=none, %
    try min ticks=10, %
    y dir=reverse, %
    colorbar style={
        at={(1.05,0.5)},anchor=west,width=0.05*\pgfkeysvalueof{/pgfplots/parent axis width},
        ytick={-1,-0.5,0,0.5,1},
        yticklabels={-1, -0.5, 0, 0.5, 1},
        },
    xticklabel pos=right, %
    enlargelimits={abs=0.5}, %
    scatter/@pre marker code/.append code={%
      \pgfplotstransformcoordinatex{sqrt(abs(\pgfplotspointmeta))}%
      \scope[mark size=\pgfplotsunitxlength*\pgfmathresult/3.7 + \pgfplotsunitxlength*20.7/2, fill=mapped color]
    },
    scatter/@post marker code/.append code={%
      \endscope%
    }
    ]

\addplot +[
    point meta=explicit, %
    only marks, %
    every node near coord/.append style={font=\small, color=white,anchor=center},
    visualization depends on={value \thisrow{label} \as \Label}, %
    visualization depends on={value \thisrow{size} \as \Size}, %
    ] table [
    x expr={int(mod(\coordindex+0.01,9))}, %
    y expr={int((\coordindex+0.01)/9))},
    meta=value,
] {
X   Y   value label size
0 0 1.0 $+$ 0.1em
0 0 0 $$ 0.1em
0 0 0 $$ 0.1em
0 0 0 $$ 0.1em
0 0 0 $$ 0.1em
0 0 0 $$ 0.1em
0 0 0 $$ 0.1em
0 0 0 $$ 0.1em
0 0 0 $$ 0.1em
 
0 0 0.8618 $+$ 0.1em
0 0 1.0 $+$ 0.1em
0 0 0 $$ 0.1em
0 0 0 $$ 0.1em
0 0 0 $$ 0.1em
0 0 0 $$ 0.1em
0 0 0 $$ 0.1em
0 0 0 $$ 0.1em
0 0 0 $$ 0.1em
 
0 0 0.8706 $+$ 0.1em
0 0 0.9824 $+$ 0.1em
0 0 1.0 $+$ 0.1em
0 0 0 $$ 0.1em
0 0 0 $$ 0.1em
0 0 0 $$ 0.1em
0 0 0 $$ 0.1em
0 0 0 $$ 0.1em
0 0 0 $$ 0.1em
 
0 0 0.5206 $+$ 0.1em
0 0 0.6147 $+$ 0.1em
0 0 0.6147 $+$ 0.1em
0 0 1.0 $+$ 0.1em
0 0 0 $$ 0.1em
0 0 0 $$ 0.1em
0 0 0 $$ 0.1em
0 0 0 $$ 0.1em
0 0 0 $$ 0.1em
 
0 0 0.1912 $+$ 0.1em
0 0 0.3176 $+$ 0.1em
0 0 0.3118 $+$ 0.1em
0 0 -0.4441 $-$ 0.1em
0 0 1.0 $+$ 0.1em
0 0 0 $$ 0.1em
0 0 0 $$ 0.1em
0 0 0 $$ 0.1em
0 0 0 $$ 0.1em
 
0 0 0.9265 $+$ 0.1em
0 0 0.9059 $+$ 0.1em
0 0 0.9206 $+$ 0.1em
0 0 0.4059 $+$ 0.1em
0 0 0.3794 $+$ 0.1em
0 0 1.0 $+$ 0.1em
0 0 0 $$ 0.1em
0 0 0 $$ 0.1em
0 0 0 $$ 0.1em
 
0 0 0.8706 $+$ 0.1em
0 0 0.9353 $+$ 0.1em
0 0 0.9647 $+$ 0.1em
0 0 0.65 $+$ 0.1em
0 0 0.2 $+$ 0.1em
0 0 0.9059 $+$ 0.1em
0 0 1.0 $+$ 0.1em
0 0 0 $$ 0.1em
0 0 0 $$ 0.1em
 
0 0 0.7941 $+$ 0.1em
0 0 0.8412 $+$ 0.1em
0 0 0.8676 $+$ 0.1em
0 0 0.3 $+$ 0.1em
0 0 0.5618 $+$ 0.1em
0 0 0.8618 $+$ 0.1em
0 0 0.7794 $+$ 0.1em
0 0 1.0 $+$ 0.1em
0 0 0 $$ 0.1em
 
0 0 0.7682 $+$ 0.1em
0 0 0.443 $+$ 0.1em
0 0 0.4503 $+$ 0.1em
0 0 0.443 $+$ 0.1em
0 0 -0.2119 $-$ 0.1em
0 0 0.5607 $+$ 0.1em
0 0 0.5254 $+$ 0.1em
0 0 0.3046 $+$ 0.1em
0 0 1.0 $+$ 0.1em
 
};

\addplot[
    mark=x,only marks, mark size=6pt, thick, gray!50,
    point meta =explicit symbolic,
    nodes near coords,
]
table[x=x,y=y, meta=label]{
    x   y   label
    0 4 {} 
    1 4 {} 
    2 4 {} 
    3 4 {} 
    3 5 {} 
    4 5 {} 
    4 6 {} 
    3 7 {} 
    1 8 {} 
    2 8 {} 
    3 8 {} 
    4 8 {} 
    7 8 {}

};

\node[anchor=west, rotate=\xticklabelrotation] at (axis cs:0+\xticklabeloffsetx,0+\xticklabeloffset) {Accuracy\strut};
\node[anchor=west, rotate=\xticklabelrotation] at (axis cs:1+\xticklabeloffsetx,0+\xticklabeloffset) {Adv. Rob. \strut};
\node[anchor=west, rotate=\xticklabelrotation] at (axis cs:2+\xticklabeloffsetx,0+\xticklabeloffset) {C-Rob.\strut};
\node[anchor=west, rotate=\xticklabelrotation] at (axis cs:3+\xticklabeloffsetx,0+\xticklabeloffset) {OOD Rob.\strut};
\node[anchor=west, rotate=\xticklabelrotation] at (axis cs:4+\xticklabeloffsetx,0+\xticklabeloffset) {Cal. Error\strut};
\node[anchor=west, rotate=\xticklabelrotation] at (axis cs:5+\xticklabeloffsetx,0+\xticklabeloffset) {Class Balance\strut};
\node[anchor=west, rotate=\xticklabelrotation] at (axis cs:6+\xticklabeloffsetx,0+\xticklabeloffset) {Obj. Focus\strut};
\node[anchor=west, rotate=\xticklabelrotation] at (axis cs:7+\xticklabeloffsetx,0+\xticklabeloffset) {Shape Bias\strut};
\node[anchor=west, rotate=\xticklabelrotation] at (axis cs:8+\xticklabeloffsetx,0+\xticklabeloffset) {Parameters\strut};

\node[anchor=center] at (axis cs:4,9.5) {\textbf{CNN IN21k (n=16)}\strut};

\end{axis}
\end{tikzpicture}
 \end{subfigure}\\
\begin{subfigure}{0.33\linewidth}
    \hspace{0pt}
    \centering
    \hspace{-2.1em} \begin{tikzpicture}[every node/.style={font=\sffamily}]
\sffamily
\scriptsize

\def\yticklabeloffset{-0.55};
\def\xticklabeloffset{-0.6};
\def\xticklabeloffsetx{-0.1};
\def\xticklabelrotation{60};

\begin{axis}[
    width=1.00\linewidth,
    axis equal image, %
    axis line style = {line width=.5pt,draw=gray!50},
    scatter, %
    colormap name=correlation_cm, %
    point meta min=-1,
    point meta max=1,
    clip=false,
    grid=minor, %
    minor grid style={line width=.5pt,draw=gray!50},
    minor tick num=1, %
    minor tick length=0pt, 
    tickwidth=0pt, %
    ticks=none, %
    try min ticks=10, %
    y dir=reverse, %
    colorbar style={
        at={(1.05,0.5)},anchor=west,width=0.05*\pgfkeysvalueof{/pgfplots/parent axis width}},
    xticklabel pos=right, %
    enlargelimits={abs=0.5}, %
    scatter/@pre marker code/.append code={%
      \pgfplotstransformcoordinatex{sqrt(abs(\pgfplotspointmeta))}%
      \scope[mark size=\pgfplotsunitxlength*\pgfmathresult/3.7 + \pgfplotsunitxlength*20.7/2, fill=mapped color]
    },
    scatter/@post marker code/.append code={%
      \endscope%
    }
]
\addplot +[
    point meta=explicit, %
    only marks, %
    every node near coord/.append style={font=\small, color=white,anchor=center},
    visualization depends on={value \thisrow{label} \as \Label}, %
    visualization depends on={value \thisrow{size} \as \Size}, %
    ] table [
    x expr={int(mod(\coordindex+0.01,9))}, %
    y expr={int((\coordindex+0.01)/9))},
    meta=value,
] {
X   Y   value label size
0 0 1.0 $+$ 0.1em
0 0 0 $$ 0.1em
0 0 0 $$ 0.1em
0 0 0 $$ 0.1em
0 0 0 $$ 0.1em
0 0 0 $$ 0.1em
0 0 0 $$ 0.1em
0 0 0 $$ 0.1em
0 0 0 $$ 0.1em
 
0 0 0.4412 $+$ 0.1em
0 0 1.0 $+$ 0.1em
0 0 0 $$ 0.1em
0 0 0 $$ 0.1em
0 0 0 $$ 0.1em
0 0 0 $$ 0.1em
0 0 0 $$ 0.1em
0 0 0 $$ 0.1em
0 0 0 $$ 0.1em
 
0 0 0.8605 $+$ 0.1em
0 0 0.4975 $+$ 0.1em
0 0 1.0 $+$ 0.1em
0 0 0 $$ 0.1em
0 0 0 $$ 0.1em
0 0 0 $$ 0.1em
0 0 0 $$ 0.1em
0 0 0 $$ 0.1em
0 0 0 $$ 0.1em
 
0 0 0.6541 $+$ 0.1em
0 0 0.5275 $+$ 0.1em
0 0 0.5957 $+$ 0.1em
0 0 1.0 $+$ 0.1em
0 0 0 $$ 0.1em
0 0 0 $$ 0.1em
0 0 0 $$ 0.1em
0 0 0 $$ 0.1em
0 0 0 $$ 0.1em
 
0 0 -0.1345 $-$ 0.1em
0 0 0.2659 $+$ 0.1em
0 0 -0.1249 $-$ 0.1em
0 0 0.1261 $+$ 0.1em
0 0 1.0 $+$ 0.1em
0 0 0 $$ 0.1em
0 0 0 $$ 0.1em
0 0 0 $$ 0.1em
0 0 0 $$ 0.1em
 
0 0 0.8378 $+$ 0.1em
0 0 0.53 $+$ 0.1em
0 0 0.713 $+$ 0.1em
0 0 0.6474 $+$ 0.1em
0 0 0.317 $+$ 0.1em
0 0 1.0 $+$ 0.1em
0 0 0 $$ 0.1em
0 0 0 $$ 0.1em
0 0 0 $$ 0.1em
 
0 0 0.7343 $+$ 0.1em
0 0 0.633 $+$ 0.1em
0 0 0.7949 $+$ 0.1em
0 0 0.5817 $+$ 0.1em
0 0 0.0473 $+$ 0.1em
0 0 0.6495 $+$ 0.1em
0 0 1.0 $+$ 0.1em
0 0 0 $$ 0.1em
0 0 0 $$ 0.1em
 
0 0 0.1552 $+$ 0.1em
0 0 0.3375 $+$ 0.1em
0 0 0.3787 $+$ 0.1em
0 0 0.2983 $+$ 0.1em
0 0 -0.0327 $-$ 0.1em
0 0 0.0995 $+$ 0.1em
0 0 0.3265 $+$ 0.1em
0 0 1.0 $+$ 0.1em
0 0 0 $$ 0.1em
 
0 0 0.4728 $+$ 0.1em
0 0 0.3936 $+$ 0.1em
0 0 0.5324 $+$ 0.1em
0 0 0.3433 $+$ 0.1em
0 0 -0.1661 $-$ 0.1em
0 0 0.335 $+$ 0.1em
0 0 0.3662 $+$ 0.1em
0 0 0.584 $+$ 0.1em
0 0 1.0 $+$ 0.1em
 
};

\addplot[
    mark=x,only marks, mark size=6pt, thick, gray!50,
    point meta =explicit symbolic,
    nodes near coords,
]
table[x=x,y=y, meta=label]{
    x   y   label
    0 4 {} 
    2 4 {} 
    3 4 {} 
    4 6 {} 
    0 7 {} 
    4 7 {} 
    5 7 {} 
    4 8 {}

};

\node[anchor=east] at (axis cs:0+\yticklabeloffset,0) {Accuracy\strut};
\node[anchor=east] at (axis cs:0+\yticklabeloffset,1) {Adv. Rob.\strut};
\node[anchor=east] at (axis cs:0+\yticklabeloffset,2) {C-Rob.\strut};
\node[anchor=east] at (axis cs:0+\yticklabeloffset,3) {OOD Rob.\strut};
\node[anchor=east] at (axis cs:0+\yticklabeloffset,4) {Cal. Error\strut};
\node[anchor=east] at (axis cs:0+\yticklabeloffset,5) {Class Balance\strut};
\node[anchor=east] at (axis cs:0+\yticklabeloffset,6) {Obj. Focus\strut};
\node[anchor=east] at (axis cs:0+\yticklabeloffset,7) {Shape Bias\strut};
\node[anchor=east] at (axis cs:0+\yticklabeloffset,8) {Parameters\strut};

\node[anchor=center] at (axis cs:4,9.5) {\textbf{Transformer (n=106)}\strut};

\end{axis}
\end{tikzpicture}
\end{subfigure}%
\begin{subfigure}{0.33\linewidth}
    \hspace{0pt}
    \centering
    \hspace{-2.15em} \begin{tikzpicture}[every node/.style={font=\sffamily}]
\sffamily
\scriptsize

\def\yticklabeloffset{-0.55};
\def\xticklabeloffset{-0.6};
\def\xticklabeloffsetx{-0.1};
\def\xticklabelrotation{60};

\begin{axis}[
    width=1.00\linewidth,
    axis equal image, %
    axis line style = {line width=.5pt,draw=gray!50},
    scatter, %
    colormap name=correlation_cm, %
    point meta min=-1,
    point meta max=1,
    clip=false,
    grid=minor, %
    minor grid style={line width=.5pt,draw=gray!50},
    minor tick num=1, %
    minor tick length=0pt, 
    tickwidth=0pt, %
    ticks=none, %
    try min ticks=10, %
    y dir=reverse, %
    colorbar style={
        at={(1.05,0.5)},anchor=west,width=0.05*\pgfkeysvalueof{/pgfplots/parent axis width}},
    xticklabel pos=right, %
    enlargelimits={abs=0.5}, %
    scatter/@pre marker code/.append code={%
      \pgfplotstransformcoordinatex{sqrt(abs(\pgfplotspointmeta))}%
      \scope[mark size=\pgfplotsunitxlength*\pgfmathresult/3.7 + \pgfplotsunitxlength*20.7/2, fill=mapped color]
    },
    scatter/@post marker code/.append code={%
      \endscope%
    }
]
\addplot +[
    point meta=explicit, %
    only marks, %
    every node near coord/.append style={font=\small, color=white,anchor=center},
    visualization depends on={value \thisrow{label} \as \Label}, %
    visualization depends on={value \thisrow{size} \as \Size}, %
    ] table [
    x expr={int(mod(\coordindex+0.01,9))}, %
    y expr={int((\coordindex+0.01)/9))},
    meta=value,
] {
X   Y   value label size
0 0 1.0 $+$ 0.1em
0 0 0 $$ 0.1em
0 0 0 $$ 0.1em
0 0 0 $$ 0.1em
0 0 0 $$ 0.1em
0 0 0 $$ 0.1em
0 0 0 $$ 0.1em
0 0 0 $$ 0.1em
0 0 0 $$ 0.1em
 
0 0 0.5238 $+$ 0.1em
0 0 1.0 $+$ 0.1em
0 0 0 $$ 0.1em
0 0 0 $$ 0.1em
0 0 0 $$ 0.1em
0 0 0 $$ 0.1em
0 0 0 $$ 0.1em
0 0 0 $$ 0.1em
0 0 0 $$ 0.1em
 
0 0 0.8765 $+$ 0.1em
0 0 0.6595 $+$ 0.1em
0 0 1.0 $+$ 0.1em
0 0 0 $$ 0.1em
0 0 0 $$ 0.1em
0 0 0 $$ 0.1em
0 0 0 $$ 0.1em
0 0 0 $$ 0.1em
0 0 0 $$ 0.1em
 
0 0 0.5996 $+$ 0.1em
0 0 0.6437 $+$ 0.1em
0 0 0.657 $+$ 0.1em
0 0 1.0 $+$ 0.1em
0 0 0 $$ 0.1em
0 0 0 $$ 0.1em
0 0 0 $$ 0.1em
0 0 0 $$ 0.1em
0 0 0 $$ 0.1em
 
0 0 -0.0777 $-$ 0.1em
0 0 0.1676 $+$ 0.1em
0 0 -0.0435 $-$ 0.1em
0 0 0.2036 $+$ 0.1em
0 0 1.0 $+$ 0.1em
0 0 0 $$ 0.1em
0 0 0 $$ 0.1em
0 0 0 $$ 0.1em
0 0 0 $$ 0.1em
 
0 0 0.8271 $+$ 0.1em
0 0 0.5321 $+$ 0.1em
0 0 0.7541 $+$ 0.1em
0 0 0.6016 $+$ 0.1em
0 0 0.3787 $+$ 0.1em
0 0 1.0 $+$ 0.1em
0 0 0 $$ 0.1em
0 0 0 $$ 0.1em
0 0 0 $$ 0.1em
 
0 0 0.685 $+$ 0.1em
0 0 0.7829 $+$ 0.1em
0 0 0.8 $+$ 0.1em
0 0 0.5922 $+$ 0.1em
0 0 0.0679 $+$ 0.1em
0 0 0.5998 $+$ 0.1em
0 0 1.0 $+$ 0.1em
0 0 0 $$ 0.1em
0 0 0 $$ 0.1em
 
0 0 -0.009 $-$ 0.1em
0 0 0.5 $+$ 0.1em
0 0 0.2496 $+$ 0.1em
0 0 0.2828 $+$ 0.1em
0 0 0.1608 $+$ 0.1em
0 0 0.037 $+$ 0.1em
0 0 0.3171 $+$ 0.1em
0 0 1.0 $+$ 0.1em
0 0 0 $$ 0.1em
 
0 0 0.4399 $+$ 0.1em
0 0 0.4371 $+$ 0.1em
0 0 0.4169 $+$ 0.1em
0 0 0.3936 $+$ 0.1em
0 0 -0.1409 $-$ 0.1em
0 0 0.3076 $+$ 0.1em
0 0 0.3336 $+$ 0.1em
0 0 0.5331 $+$ 0.1em
0 0 1.0 $+$ 0.1em
 
};

\addplot[
    mark=x,only marks, mark size=6pt, thick, gray!50,
    point meta =explicit symbolic,
    nodes near coords,
]
table[x=x,y=y, meta=label]{
    x   y   label
    0 4 {} 
    1 4 {} 
    2 4 {} 
    3 4 {} 
    4 6 {} 
    0 7 {} 
    4 7 {} 
    5 7 {} 
    4 8 {}

};

\node[anchor=center] at (axis cs:4,9.5) {\textbf{Transformer IN1k (n=84)}\strut};

\end{axis}
\end{tikzpicture}
\end{subfigure}%
\begin{subfigure}{0.33\linewidth}
    \hspace{0pt}
    \centering
    \hspace{-2em}\begin{tikzpicture}[every node/.style={font=\sffamily}]
\sffamily

\scriptsize

\def\yticklabeloffset{-0.55};
\def\xticklabeloffset{-0.6};
\def\xticklabeloffsetx{-0.1};
\def\xticklabelrotation{60};

\begin{axis}[
    width=1.00\linewidth,
    axis equal image, %
    axis line style = {line width=.5pt,draw=gray!50},
    scatter, %
    colormap name=correlation_cm, %
    colorbar, %
    point meta min=-1,
    point meta max=1,
    clip=false,
    grid=minor, %
    minor grid style={line width=.5pt,draw=gray!50},
    minor tick num=1, %
    minor tick length=0pt, 
    tickwidth=0pt, %
    ticks=none, %
    try min ticks=10, %
    y dir=reverse, %
    colorbar style={
        at={(1.05,0.5)},anchor=west,width=0.05*\pgfkeysvalueof{/pgfplots/parent axis width},
        ytick={-1,-0.5,0,0.5,1},
        yticklabels={-1, -0.5, 0, 0.5, 1},
        },
    xticklabel pos=right, %
    enlargelimits={abs=0.5}, %
    scatter/@pre marker code/.append code={%
      \pgfplotstransformcoordinatex{sqrt(abs(\pgfplotspointmeta))}%
      \scope[mark size=\pgfplotsunitxlength*\pgfmathresult/3.7 + \pgfplotsunitxlength*20.7/2, fill=mapped color]
    },
    scatter/@post marker code/.append code={%
      \endscope%
    }
]
\addplot +[
    point meta=explicit, %
    only marks, %
    every node near coord/.append style={font=\small, color=white,anchor=center},
    visualization depends on={value \thisrow{label} \as \Label}, %
    visualization depends on={value \thisrow{size} \as \Size}, %
    ] table [
    x expr={int(mod(\coordindex+0.01,9))}, %
    y expr={int((\coordindex+0.01)/9))},
    meta=value,
] {
X   Y   value label size
0 0 1.0 $+$ 0.1em
0 0 0 $$ 0.1em
0 0 0 $$ 0.1em
0 0 0 $$ 0.1em
0 0 0 $$ 0.1em
0 0 0 $$ 0.1em
0 0 0 $$ 0.1em
0 0 0 $$ 0.1em
0 0 0 $$ 0.1em
 
0 0 0.7228 $+$ 0.1em
0 0 1.0 $+$ 0.1em
0 0 0 $$ 0.1em
0 0 0 $$ 0.1em
0 0 0 $$ 0.1em
0 0 0 $$ 0.1em
0 0 0 $$ 0.1em
0 0 0 $$ 0.1em
0 0 0 $$ 0.1em
 
0 0 0.793 $+$ 0.1em
0 0 0.4123 $+$ 0.1em
0 0 1.0 $+$ 0.1em
0 0 0 $$ 0.1em
0 0 0 $$ 0.1em
0 0 0 $$ 0.1em
0 0 0 $$ 0.1em
0 0 0 $$ 0.1em
0 0 0 $$ 0.1em
 
0 0 0.8632 $+$ 0.1em
0 0 0.6404 $+$ 0.1em
0 0 0.6035 $+$ 0.1em
0 0 1.0 $+$ 0.1em
0 0 0 $$ 0.1em
0 0 0 $$ 0.1em
0 0 0 $$ 0.1em
0 0 0 $$ 0.1em
0 0 0 $$ 0.1em
 
0 0 0.3246 $+$ 0.1em
0 0 0.314 $+$ 0.1em
0 0 0.1088 $+$ 0.1em
0 0 0.5614 $+$ 0.1em
0 0 1.0 $+$ 0.1em
0 0 0 $$ 0.1em
0 0 0 $$ 0.1em
0 0 0 $$ 0.1em
0 0 0 $$ 0.1em
 
0 0 0.9333 $+$ 0.1em
0 0 0.7702 $+$ 0.1em
0 0 0.6175 $+$ 0.1em
0 0 0.8561 $+$ 0.1em
0 0 0.4596 $+$ 0.1em
0 0 1.0 $+$ 0.1em
0 0 0 $$ 0.1em
0 0 0 $$ 0.1em
0 0 0 $$ 0.1em
 
0 0 0.8404 $+$ 0.1em
0 0 0.4632 $+$ 0.1em
0 0 0.8526 $+$ 0.1em
0 0 0.6561 $+$ 0.1em
0 0 0.2456 $+$ 0.1em
0 0 0.7772 $+$ 0.1em
0 0 1.0 $+$ 0.1em
0 0 0 $$ 0.1em
0 0 0 $$ 0.1em
 
0 0 0.3456 $+$ 0.1em
0 0 0.1632 $+$ 0.1em
0 0 0.714 $+$ 0.1em
0 0 0.1368 $+$ 0.1em
0 0 -0.2667 $-$ 0.1em
0 0 0.1526 $+$ 0.1em
0 0 0.4228 $+$ 0.1em
0 0 1.0 $+$ 0.1em
0 0 0 $$ 0.1em
 
0 0 0.5714 $+$ 0.1em
0 0 0.4246 $+$ 0.1em
0 0 0.7692 $+$ 0.1em
0 0 0.3288 $+$ 0.1em
0 0 -0.1952 $-$ 0.1em
0 0 0.3824 $+$ 0.1em
0 0 0.4809 $+$ 0.1em
0 0 0.8185 $+$ 0.1em
0 0 1.0 $+$ 0.1em
 
};

\addplot[
    mark=x,only marks, mark size=6pt, thick, gray!50,
    point meta =explicit symbolic,
    nodes near coords,
]
table[x=x,y=y, meta=label]{
    x   y   label
    1 2 {} 
    0 4 {} 
    1 4 {} 
    2 4 {} 
    4 6 {} 
    0 7 {} 
    1 7 {} 
    3 7 {} 
    4 7 {} 
    5 7 {} 
    6 7 {} 
    1 8 {} 
    3 8 {} 
    4 8 {} 
    5 8 {}

};

\node[anchor=center] at (axis cs:4,9.5) {\textbf{Transformer IN21k (n=19)}\strut};

\end{axis}
\end{tikzpicture}
\end{subfigure}
\caption{\textit{Rank correlation matrices for model subgroups.} We investigate the rank correlations of different quality dimensions for all CNNs \textit{(top left)}, all CNNs trained on ImageNet-1k \textit{(top middle)}, all CNNs trained on ImageNet-21k \textit{(top right)}, all Transformers \textit{(bottom left)}, all Transformers trained on ImageNet-1k \textit{(bottom middle)}, and all Transformers trained on ImageNet-21k \textit{(bottom right)}. Crossed-out entries indicate a $p$-value above 0.05, and thus, are not statistically significant.}
\label{fig:corr_matrix_cnn_transformer}
\end{figure*}

Given that a larger fraction of the newer Transformer models were trained on ImageNet-21k, whereas most CNNs were trained on ImageNet-1k, some of the correlations could also be due to the training dataset size. To account for this, in \cref{fig:corr_matrix_cnn_transformer}, we further plot the correlation matrices for CNNs and Transformers trained exclusively on ImageNet-1k and ImageNet-21k, respectively. %
Interestingly, the negative correlation between adversarial robustness and C-robustness/OOD robustness for CNNs is only apparent when trained on ImageNet-1k but not when trained on ImageNet-21k. Calibration error and OOD robustness have a strong negative correlation for ImageNet-21k CNNs while having almost no correlation for ImageNet-1k CNNs (however, not statistically significant). Generally, the correlations for CNNs are much more pronounced for the models trained on the larger dataset.
For Transformers, the statistically significant correlations are quite similar for models trained on ImageNet-1k and ImageNet-21k.

\subsection{OOD robustness for models trained on large-scale datasets}\label{appendix:sec:ood}

In \cref{sec:experiments_whatmakesbetter} of the main paper, in the paragraph about vision-language (ViL) models, we state that vision-language models outperform other models trained on similarly large datasets when it comes to out-of-domain robustness. In \cref{tab:large_dataset_ood}, we support this statement with numerical results. To this end, we report the ``raw'' OOD accuracy, \ie, the OOD accuracy without normalizing by the clean accuracy, for the 15 models with the highest OOD accuracy. The best ViL model achieves an OOD accuracy of \num{0.87} while the best self-supervised model (pre-)trained on a large-scale dataset only achieves an OOD accuracy of \num{0.79}. Further, vision-language models clearly dominate in the list (14 out of 15). These results suggest that the increased robustness of ViL models is not only due to the increased dataset size but also due to other factors that are likely linked to the language part of the models.

\begin{table}[t]
  \centering
  \tiny
      \caption{\textit{Top 15 models with the highest OOD accuracy.} %
    Vision-language (ViL) models are clearly dominating the list.}
  \begin{tabularx}{\linewidth}{@{}lXS[table-format=1.2]@{}}%
    \toprule
    \textbf{Model} & {\textbf{Configuration}}&  {\makecell{\textbf{OOD} \\ \textbf{Acc.}}$\uparrow$} \\
\midrule
SigLIP2-l/16 ~\citep{Tschannen:2025:SMV}  & ViL, WebLI~\citep{Chen:2023:PAL}, self-SL (E2E) & 0.87\\
MetaCLIP-L/14 ~\citep{Xu:2024:DCD}  & ViL, MetaCLIP-400M~\citep{Xu:2024:DCD}, self-SL & 0.85\\
MobileCLIP-B (LT) ~\citep{Vase:2024:MFI}  & ViL, DataCompDR-1B~\citep{Vase:2024:MFI}, self-SL & 0.85\\
CLIP-L/14-CommPool XL-DFN2B ~\citep{Fang:2024:DFN}  & ViL, DFN2B~\citep{Fang:2024:DFN}, self-SL (E2E) & 0.84\\
SigLIP-l/16 ~\citep{Zhai:2023:SLF}  & ViL, WebLI~\citep{Chen:2023:PAL}, self-SL & 0.83\\
CLIP-L/14-DataCompXL ~\citep{Gadre:2023:DIS}  & ViL, DataCompXL~\citep{Gadre:2023:DIS}, self-SL (E2E) & 0.83\\
MobileCLIP-B ~\citep{Vase:2024:MFI}  & ViL, DataCompDR-1B~\citep{Vase:2024:MFI}, self-SL & 0.83\\
SigLIP2-b/16 ~\citep{Tschannen:2025:SMV}  & ViL, WebLI~\citep{Chen:2023:PAL}, self-SL (E2E) & 0.82\\
CLIP-ConvNeXt-L ~\citep{Schuhmann:2022:AOL, Wortsman:2022:MSA}  & ViL, Laion2B~\citep{Schuhmann:2022:AOL}, self-SL & 0.81\\
CLIP-L/14-Laion2B ~\citep{Schuhmann:2022:AOL,  Wortsman:2022:MSA}  & ViL, Laion2B~\citep{Schuhmann:2022:AOL}, self-SL (E2E) & 0.80\\
SigLIP-b/16 ~\citep{Zhai:2023:SLF}  & ViL, WebLI~\citep{Chen:2023:PAL}, self-SL & 0.80\\
CLIP-ConvNeXt-L-320px ~\citep{Schuhmann:2022:AOL,  Wortsman:2022:MSA}  & ViL, Laion2B~\citep{Schuhmann:2022:AOL}, self-SL & 0.80\\
CLIP-B/16-DataCompXL ~\citep{Gadre:2023:DIS}  & ViL, DataCompXL~\citep{Gadre:2023:DIS}, self-SL (E2E) & 0.79\\
ViT-l/14-DINOv2-reg-LP ~\citep{Darcet:2024:VTN}  & Transformer, LVD142m~\citep{Oquab:2024:DIN}, self-SL (E2E) & 0.79\\
MetaCLIP-B/16 ~\citep{Xu:2024:DCD}  & ViL, MetaCLIP-400M~\citep{Xu:2024:DCD}, self-SL & 0.78\\

    \bottomrule
  \end{tabularx}

  \label{tab:large_dataset_ood}
\end{table}

\subsection{Reproducing conflicting results}\label{appendix:sec:conflicting_results}

We report several results that extend findings from related work using smaller model pools. To verify these results, we reproduce their experiments with similar model pools in \cref{tab:reproducing}. When using comparable models, we successfully replicate most of their findings, indicating that the discrepancies arise from the limited model pools in related work. This underscores the importance of our large model zoo.

\begin{table*}[t]
  \centering
  \tiny
      \caption{\textit{Comparison of our results that contradict/extend findings from related work.} In the main paper, we highlight a few findings that diverge from results in related work. To verify these discrepancies, we reproduce their findings using a similar model pool. In doing so, we find that most of their conclusions hold when considering similar models.}
  \begin{tabularx}{\linewidth}{@{}p{2cm}Xp{3cm}p{2.5cm}p{1.2cm}@{}}%
    \toprule
    \textbf{Finding from related work on their respective models} & \textbf{Used models in related work} & \textbf{Reproduction with similar models} & \textbf{Our finding} & \textbf{Used models}\\
    \midrule
    Adversarial Training improves calibration error~\citep{Grabinski:2022:RBA} & ResNet18, ResNet50~\citep{He:2016:DRL}, WRN-50-2 ~\citep{Zagoruyko:2016:WRN} \newline \vsiccv \newline ResNet18, ResNet50, WRN-50-2 from \citet{Salman:2020:ARI} \newline NOTE: \citet{Grabinski:2022:RBA} conducted additional experiments with non-ImageNet models & \textbf{Average ECE} \newline Standard Models: 0.2722 \newline Robust Models: 0.0876 \newline \textbf{Conclusion}: We can reproduce their finding when using a similar model pool & Adversarial training impairs calibration when considering a broad and diverse set of models & See \cref{tab:details_comparison} \tablesetupat \\
    \midrule
    Longer training worsens ECE~\citep{Minderer:2021:RCM} & Five self-trained versions of BiT-L-R50x1 and BiT-L-R101x3~\citep{Minderer:2021:RCM} \newline NOTE: Since the used or similar checkpoints are not publicly available, we use the A[1,2]~\citep{Wightman:2021:RSB} versions of ResNet18, ResNet34, ResNet50, ResNet101 and ResNet152, which are at least somewhat similar to BiT~\citep{Kolesnikov:2020:BIT}, for reproduction & \textbf{Average ECE} \newline Standard models: 0.0396 \newline A[1] models: 0.0845 \newline A[2] models: 0.1025 \newline \textbf{Conclusion}: We can reproduce their finding when using a similar model pool & Longer training improves calibration error when considering a broad and diverse set of models& See \cref{tab:details_comparison} \tablesetuplong 
    \\

    \midrule
    Accuracy is positively correlated with calibration error~\citep{Guo:2017:OCM} & DenseNet161~\citep{Huang:2017:DCC}, ResNet152~\citep{He:2016:DRL} \newline NOTE: \citet{Guo:2017:OCM} conducted additional experiments with non-ImageNet models & \textbf{Accuracy \& ECE} \newline ResNet152~\citep{He:2016:DRL}: 0.7832 \& 0.05 \newline DenseNet161~\citep{Huang:2017:DCC}: 0.7711 \& 0.06 \newline \textbf{Conclusion}: We can \emph{not} reproduce their finding when using a similar model pool & We found that accuracy is negatively correlated with calibration error (p $<$ 0.05) when considering a broad and diverse set of models & Entire model zoo \\

    \midrule
    Adversarial robustness is positively correlated with C-robustness and OOD robustness for a given architecture~\citep{Liu:2023:CSR} & VGG13~\citep{Simonyan:2015:VDC}, VGG16~\citep{Simonyan:2015:VDC}, VGG19~\citep{Simonyan:2015:VDC}, XciT-S~\citep{Ali:2021:XCC}, XciT-M~\citep{Ali:2021:XCC}, XciT-L~\citep{Ali:2021:XCC}, ResNet50~\citep{He:2016:DRL}, ResNet101~\citep{He:2016:DRL}, ResNet152~\citep{He:2016:DRL}, Wide-ResNet50~\citep{Salman:2020:ARI}, DenseNet121~\citep{Huang:2017:DCC}, DenseNet161~\citep{Huang:2017:DCC}, DenseNet201~\citep{Huang:2017:DCC}, ConvNeXT-S~\citep{Liu:2022:ACF}, ConvNeXT-S (21k) ~\citep{Liu:2022:ACF}, ConvNeXT-B~\citep{Liu:2022:ACF}, ConvNeXT-B (21k)~\citep{Liu:2022:ACF}, ConvNeXT-L~\citep{Liu:2022:ACF}, ConvNeXT-L (21k)~\citep{Liu:2022:ACF}, ViT-s/16~\citep{Dosovitskiy:2021:IWW}, ViT-s/16 (21k)~\citep{Dosovitskiy:2021:IWW}, ViT-b/16~\citep{Dosovitskiy:2021:IWW}, ViT-b/16 (21k)~\citep{Dosovitskiy:2021:IWW}, ViT-b/16 (MAE)~\citep{He:2022:MAA}, ViT-l/16~\citep{Dosovitskiy:2021:IWW}, ViT-l/16 (21k)~\citep{Dosovitskiy:2021:IWW}, ViT-l (MAE)~\citep{He:2022:MAA}, Swin-S~\citep{Liu:2021:STH}, Swin-b~\citep{Liu:2021:STH}, Swin-b (21k)~\citep{Liu:2021:STH}, ResNet50 (Mocov3) ~\citep{Chen:2021:AES}, T2T-14~\citep{Yuan:2021:TTT}, T2T-19~\citep{Yuan:2021:TTT}, T2T-24~\citep{Yuan:2021:TTT}, Swin-L~\citep{Liu:2021:STH} & Correlation between adversarial robustness and c-robustness: 0.6547 (p = 0.0). Correlation between adversarial robustness and OOD robustness: 0.6942 (p=0.0) \newline \textbf{Conclusion}: We can reproduce their finding when using a similar model pool & We found no significant correlations between adversarial robustness and c-robustness (p=0.84) or OOD robustness (p=0.15) when considering a broad and diverse set of models & 
    Entire model zoo \\
    \bottomrule
  \end{tabularx}

  \label{tab:reproducing}
\end{table*}

\subsection{Comparison between our adversarial robustness protocol and AutoAttack}\label{appendix:sec:autoattack}

In the main paper, we employ a combination of FGSM and PGD adversarial attacks to assess adversarial robustness. However, within the community, alternative evaluation standards have emerged, such as \emph{AutoAttack} -- a widely adopted adversarial benchmark introduced in \emph{RobustBench} \citep{Croce:2021:RBS}. It consists of an ensemble of four parameter-free attacks: two PGD variants using cross-entropy and difference-of-logits ratio losses, a targeted FAB attack \citep{Croce:2020:MDA}, and a black-box Square attack \citep{Andriushchenko:2020:SAA}.

We opted for a more straightforward setup, as we found the success rate of \emph{AutoAttack} to be excessively high for our purposes -- nearly all non-adversarially robust models are reduced below 0.1 accuracy, effectively collapsing. Moreover, prior work has criticized \emph{AutoAttack} for producing perturbations that significantly alter the input images, making adversarial examples easily detectable, and for exhibiting sensitivity to image resolution~\citep{Lorenz:2021:IRA}.

Nevertheless, given that \emph{AutoAttack} remains an important benchmark in adversarial robustness research, we include a comparison between the results obtained using \emph{AutoAttack} and those produced by our proposed evaluation protocol in this section. 
First, we evaluate how different training paradigms and architectures (\cf \cref{sec:experiments_whatmakesbetter}) affect adversarial robustness under both our protocol and \emph{AutoAttack}, as shown in \cref{tab:comparisons_autoattack}. For nearly all configurations, the overall conclusions remain consistent. Only in setups (e) and (f) do we observe improvements in adversarial robustness when using our protocol, while robustness under \emph{AutoAttack} remains unchanged (though these differences are not statistically significant).
Next, we compare the correlation matrices among our full model zoo (\cf \cref{sec:experiments_correlations}) with the two different adversarial robustness protocols in \cref{fig:adv_rob_vs_autoattack}. Again, all the statistically significant conclusions remain the same.
Lastly, we compare how model rankings change (\cf \cref{sec:experiments_rankingdetails}) under the two different evaluation protocols for adversarial robustness. While the rank correlation between the two adversarial robustness metrics remains fairly high (0.64), the overall ordering of models changes substantially. This shift occurs because, under \emph{AutoAttack}, nearly the entire model zoo collapses -- reducing the mean/standard deviation of adversarial robustness from 0.2/0.11 to 0.04/0.03. Consequently, the adversarially trained models that still achieve relatively high accuracies of around 0.6–0.7 deviate by many more standard deviations from the mean. As a result, the adversarial robustness score dominates the overall ranking, with only adversarially robust models appearing among the top five (see Table \ref{tab:top5_autoattack}). Although this effect could easily be mitigated by reweighting our proposed QUBA score, as discussed in \cref{sec:experiments_rankingdetails}, we consider this finding an indication that \emph{AutoAttack} is not the most suitable evaluation protocol for adversarial robustness within our benchmark. Nevertheless, aside from the unweighted rankings -- which must be weighted according to user requirements in any case -- almost none of the conclusions in our work would differ under \emph{AutoAttack}.

\begin{figure*}[t]
\centering
\begin{subfigure}{0.4\linewidth}
    \centering
    \begin{tikzpicture}[every node/.style={font=\sffamily}]
\sffamily
\scriptsize

\def\yticklabeloffset{-0.55};
\def\xticklabeloffset{-0.6};
\def\xticklabeloffsetx{-0.1};
\def\xticklabelrotation{60};

\begin{axis}[%
    width=1.00\linewidth,
    axis equal image, %
    axis line style = {line width=.5pt,draw=gray!50},
    scatter, %
    colormap name=correlation_cm, %
    point meta min=-1,
    point meta max=1,
    clip=false,
    grid=minor, %
    minor grid style={line width=.5pt,draw=gray!50},
    minor tick num=1, %
    minor tick length=0pt, 
    tickwidth=0pt, %
    ticks=none, %
    try min ticks=10, %
    y dir=reverse, %
    colorbar style={
        at={(1.05,0.5)},anchor=west,width=0.05*\pgfkeysvalueof{/pgfplots/parent axis width}},
    xticklabel pos=right, %
    enlargelimits={abs=0.5}, %
    scatter/@pre marker code/.append code={%
      \pgfplotstransformcoordinatex{sqrt(abs(\pgfplotspointmeta))}%
      \scope[mark size=\pgfplotsunitxlength*\pgfmathresult/3.7 + \pgfplotsunitxlength*20.7/2, fill=mapped color]
    },
    scatter/@post marker code/.append code={%
      \endscope%
    }
    ]

\addplot +[
    point meta=explicit, %
    only marks, %
    every node near coord/.append style={font=\small, color=white,anchor=center},
    visualization depends on={value \thisrow{label} \as \Label}, %
    visualization depends on={value \thisrow{size} \as \Size}, %
    ] table [
    x expr={int(mod(\coordindex+0.01,9))}, %
    y expr={int((\coordindex+0.01)/9))},
    meta=value,
] {
X   Y   value label size
0 0 1.0 $+$ 0.1em
0 0 0 $$ 0.1em
0 0 0 $$ 0.1em
0 0 0 $$ 0.1em
0 0 0 $$ 0.1em
0 0 0 $$ 0.1em
0 0 0 $$ 0.1em
0 0 0 $$ 0.1em
0 0 0 $$ 0.1em
 
0 0 0.4399 $+$ 0.1em
0 0 1.0 $+$ 0.1em
0 0 0 $$ 0.1em
0 0 0 $$ 0.1em
0 0 0 $$ 0.1em
0 0 0 $$ 0.1em
0 0 0 $$ 0.1em
0 0 0 $$ 0.1em
0 0 0 $$ 0.1em
 
0 0 0.6239 $+$ 0.1em
0 0 0.0116 $+$ 0.1em
0 0 1.0 $+$ 0.1em
0 0 0 $$ 0.1em
0 0 0 $$ 0.1em
0 0 0 $$ 0.1em
0 0 0 $$ 0.1em
0 0 0 $$ 0.1em
0 0 0 $$ 0.1em
 
0 0 0.3523 $+$ 0.1em
0 0 -0.0807 $-$ 0.1em
0 0 0.7957 $+$ 0.1em
0 0 1.0 $+$ 0.1em
0 0 0 $$ 0.1em
0 0 0 $$ 0.1em
0 0 0 $$ 0.1em
0 0 0 $$ 0.1em
0 0 0 $$ 0.1em
 
0 0 -0.1175 $-$ 0.1em
0 0 0.066 $+$ 0.1em
0 0 0.0779 $+$ 0.1em
0 0 0.2519 $+$ 0.1em
0 0 1.0 $+$ 0.1em
0 0 0 $$ 0.1em
0 0 0 $$ 0.1em
0 0 0 $$ 0.1em
0 0 0 $$ 0.1em
 
0 0 0.5338 $+$ 0.1em
0 0 0.0821 $+$ 0.1em
0 0 0.7349 $+$ 0.1em
0 0 0.7625 $+$ 0.1em
0 0 0.4903 $+$ 0.1em
0 0 1.0 $+$ 0.1em
0 0 0 $$ 0.1em
0 0 0 $$ 0.1em
0 0 0 $$ 0.1em
 
0 0 0.7169 $+$ 0.1em
0 0 0.4489 $+$ 0.1em
0 0 0.6397 $+$ 0.1em
0 0 0.4709 $+$ 0.1em
0 0 0.0892 $+$ 0.1em
0 0 0.5691 $+$ 0.1em
0 0 1.0 $+$ 0.1em
0 0 0 $$ 0.1em
0 0 0 $$ 0.1em
 
0 0 0.2612 $+$ 0.1em
0 0 0.1673 $+$ 0.1em
0 0 0.6071 $+$ 0.1em
0 0 0.623 $+$ 0.1em
0 0 0.2409 $+$ 0.1em
0 0 0.5399 $+$ 0.1em
0 0 0.5213 $+$ 0.1em
0 0 1.0 $+$ 0.1em
0 0 0 $$ 0.1em
 
0 0 0.3119 $+$ 0.1em
0 0 0.1781 $+$ 0.1em
0 0 0.4356 $+$ 0.1em
0 0 0.4339 $+$ 0.1em
0 0 0.108 $+$ 0.1em
0 0 0.4748 $+$ 0.1em
0 0 0.4821 $+$ 0.1em
0 0 0.5299 $+$ 0.1em
0 0 1.0 $+$ 0.1em
 
};

\addplot[
    mark=x,only marks, mark size=6pt, thick, gray!50,
    point meta =explicit symbolic,
    nodes near coords,
]
table[x=x,y=y, meta=label]{
    x   y   label
    1 2 {} 
    1 3 {} 
    1 4 {} 
    2 4 {} 
    1 5 {} 
    4 6 {} 
    4 8 {}

};

\node[anchor=east] at (axis cs:0+\yticklabeloffset,0) {Accuracy\strut};
\node[anchor=east] at (axis cs:0+\yticklabeloffset,1) {Adv. Rob.\strut};
\node[anchor=east] at (axis cs:0+\yticklabeloffset,2) {C-Rob.\strut};
\node[anchor=east] at (axis cs:0+\yticklabeloffset,3) {OOD Rob.\strut};
\node[anchor=east] at (axis cs:0+\yticklabeloffset,4) {Cal. Error\strut};
\node[anchor=east] at (axis cs:0+\yticklabeloffset,5) {Class Balance\strut};
\node[anchor=east] at (axis cs:0+\yticklabeloffset,6) {Obj. Focus\strut};
\node[anchor=east] at (axis cs:0+\yticklabeloffset,7) {Shape Bias\strut};
\node[anchor=east] at (axis cs:0+\yticklabeloffset,8) {Parameters\strut};

\node[anchor=west, rotate=\xticklabelrotation] at (axis cs:0+\xticklabeloffsetx,0+\xticklabeloffset) {Accuracy\strut};
\node[anchor=west, rotate=\xticklabelrotation] at (axis cs:1+\xticklabeloffsetx,0+\xticklabeloffset) {Adv. Rob. \strut};
\node[anchor=west, rotate=\xticklabelrotation] at (axis cs:2+\xticklabeloffsetx,0+\xticklabeloffset) {C-Rob.\strut};
\node[anchor=west, rotate=\xticklabelrotation] at (axis cs:3+\xticklabeloffsetx,0+\xticklabeloffset) {OOD Rob.\strut};
\node[anchor=west, rotate=\xticklabelrotation] at (axis cs:4+\xticklabeloffsetx,0+\xticklabeloffset) {Cal. Error\strut};
\node[anchor=west, rotate=\xticklabelrotation] at (axis cs:5+\xticklabeloffsetx,0+\xticklabeloffset) {Class Balance\strut};
\node[anchor=west, rotate=\xticklabelrotation] at (axis cs:6+\xticklabeloffsetx,0+\xticklabeloffset) {Obj. Focus\strut};
\node[anchor=west, rotate=\xticklabelrotation] at (axis cs:7+\xticklabeloffsetx,0+\xticklabeloffset) {Shape Bias\strut};
\node[anchor=west, rotate=\xticklabelrotation] at (axis cs:8+\xticklabeloffsetx,0+\xticklabeloffset) {Parameters\strut};

\end{axis}
\end{tikzpicture}\vspace{+1.7em}
\end{subfigure}%
\begin{subfigure}{0.4\linewidth}
     \centering
     \begin{tikzpicture}[every node/.style={font=\sffamily}]
\sffamily
\scriptsize

\def\yticklabeloffset{-0.55};
\def\xticklabeloffset{-0.6};
\def\xticklabeloffsetx{-0.1};
\def\xticklabelrotation{60};

\begin{axis}[%
    width=1.00\linewidth,
    axis equal image, %
    axis line style = {line width=.5pt,draw=gray!50},
    scatter, %
    colormap name=correlation_cm, %
    colorbar, %
    point meta min=-1,
    point meta max=1,
    clip=false,
    grid=minor, %
    minor grid style={line width=.5pt,draw=gray!50},
    minor tick num=1, %
    minor tick length=0pt, 
    tickwidth=0pt, %
    ticks=none, %
    try min ticks=10, %
    y dir=reverse, %
    colorbar style={
        at={(1.05,0.5)},anchor=west,width=0.05*\pgfkeysvalueof{/pgfplots/parent axis width},
        ytick={-1,-0.5,0,0.5,1},
        yticklabels={-1, -0.5, 0, 0.5, 1},
        },
    xticklabel pos=right, %
    enlargelimits={abs=0.5}, %
    scatter/@pre marker code/.append code={%
      \pgfplotstransformcoordinatex{sqrt(abs(\pgfplotspointmeta))}%
      \scope[mark size=\pgfplotsunitxlength*\pgfmathresult/3.7 + \pgfplotsunitxlength*20.7/2, fill=mapped color]
    },
    scatter/@post marker code/.append code={%
      \endscope%
    }
    ]

\addplot +[
    point meta=explicit, %
    only marks, %
    every node near coord/.append style={font=\small, color=white,anchor=center},
    visualization depends on={value \thisrow{label} \as \Label}, %
    visualization depends on={value \thisrow{size} \as \Size}, %
    ] table [
    x expr={int(mod(\coordindex+0.01,9))}, %
    y expr={int((\coordindex+0.01)/9))},
    meta=value,
] {
X   Y   value label size
0 0 1.0 $+$ 0.1em
0 0 0 $$ 0.1em
0 0 0 $$ 0.1em
0 0 0 $$ 0.1em
0 0 0 $$ 0.1em
0 0 0 $$ 0.1em
0 0 0 $$ 0.1em
0 0 0 $$ 0.1em
0 0 0 $$ 0.1em
 
0 0 0.4041 $+$ 0.1em
0 0 1.0 $+$ 0.1em
0 0 0 $$ 0.1em
0 0 0 $$ 0.1em
0 0 0 $$ 0.1em
0 0 0 $$ 0.1em
0 0 0 $$ 0.1em
0 0 0 $$ 0.1em
0 0 0 $$ 0.1em
 
0 0 0.6239 $+$ 0.1em
0 0 0.262 $+$ 0.1em
0 0 1.0 $+$ 0.1em
0 0 0 $$ 0.1em
0 0 0 $$ 0.1em
0 0 0 $$ 0.1em
0 0 0 $$ 0.1em
0 0 0 $$ 0.1em
0 0 0 $$ 0.1em
 
0 0 0.3523 $+$ 0.1em
0 0 0.1829 $+$ 0.1em
0 0 0.7957 $+$ 0.1em
0 0 1.0 $+$ 0.1em
0 0 0 $$ 0.1em
0 0 0 $$ 0.1em
0 0 0 $$ 0.1em
0 0 0 $$ 0.1em
0 0 0 $$ 0.1em
 
0 0 -0.1175 $-$ 0.1em
0 0 -0.0817 $-$ 0.1em
0 0 0.0779 $+$ 0.1em
0 0 0.2519 $+$ 0.1em
0 0 1.0 $+$ 0.1em
0 0 0 $$ 0.1em
0 0 0 $$ 0.1em
0 0 0 $$ 0.1em
0 0 0 $$ 0.1em
 
0 0 0.5338 $+$ 0.1em
0 0 0.1315 $+$ 0.1em
0 0 0.7349 $+$ 0.1em
0 0 0.7625 $+$ 0.1em
0 0 0.4903 $+$ 0.1em
0 0 1.0 $+$ 0.1em
0 0 0 $$ 0.1em
0 0 0 $$ 0.1em
0 0 0 $$ 0.1em
 
0 0 0.7169 $+$ 0.1em
0 0 0.3563 $+$ 0.1em
0 0 0.6397 $+$ 0.1em
0 0 0.4709 $+$ 0.1em
0 0 0.0892 $+$ 0.1em
0 0 0.5691 $+$ 0.1em
0 0 1.0 $+$ 0.1em
0 0 0 $$ 0.1em
0 0 0 $$ 0.1em
 
0 0 0.2612 $+$ 0.1em
0 0 0.275 $+$ 0.1em
0 0 0.6071 $+$ 0.1em
0 0 0.623 $+$ 0.1em
0 0 0.2409 $+$ 0.1em
0 0 0.5399 $+$ 0.1em
0 0 0.5213 $+$ 0.1em
0 0 1.0 $+$ 0.1em
0 0 0 $$ 0.1em
 
0 0 0.3119 $+$ 0.1em
0 0 0.0473 $+$ 0.1em
0 0 0.4356 $+$ 0.1em
0 0 0.4339 $+$ 0.1em
0 0 0.108 $+$ 0.1em
0 0 0.4748 $+$ 0.1em
0 0 0.4821 $+$ 0.1em
0 0 0.5299 $+$ 0.1em
0 0 1.0 $+$ 0.1em
 
};

        \addplot[
            mark=x,only marks, mark size=6pt, thick, gray!50,
            point meta =explicit symbolic,
            nodes near coords,
        ]
        table[x=x,y=y, meta=label]{
            x   y   label
            1 4 {} 
2 4 {} 
4 6 {} 
1 8 {} 
4 8 {}

        };

\node[anchor=east] at (axis cs:0+\yticklabeloffset,0) {Accuracy\strut};
\node[anchor=east] at (axis cs:0+\yticklabeloffset,1) {AutoAttack  \strut};
\node[anchor=east] at (axis cs:0+\yticklabeloffset,2) {C-Rob.\strut};
\node[anchor=east] at (axis cs:0+\yticklabeloffset,3) {OOD Rob.\strut};
\node[anchor=east] at (axis cs:0+\yticklabeloffset,4) {Cali. Error\strut};
\node[anchor=east] at (axis cs:0+\yticklabeloffset,5) {Fairness\strut};
\node[anchor=east] at (axis cs:0+\yticklabeloffset,6) {Obj. Focus\strut};
\node[anchor=east] at (axis cs:0+\yticklabeloffset,7) {Shape Bias\strut};
\node[anchor=east] at (axis cs:0+\yticklabeloffset,8) {Parameters\strut};

\node[anchor=west, rotate=\xticklabelrotation] at (axis cs:0+\xticklabeloffsetx,0+\xticklabeloffset) {Accuracy\strut};
\node[anchor=west, rotate=\xticklabelrotation] at (axis cs:1+\xticklabeloffsetx,0+\xticklabeloffset) {AutoAttack \strut};
\node[anchor=west, rotate=\xticklabelrotation] at (axis cs:2+\xticklabeloffsetx,0+\xticklabeloffset) {C-Rob.\strut};
\node[anchor=west, rotate=\xticklabelrotation] at (axis cs:3+\xticklabeloffsetx,0+\xticklabeloffset) {OOD Rob.\strut};
\node[anchor=west, rotate=\xticklabelrotation] at (axis cs:4+\xticklabeloffsetx,0+\xticklabeloffset) {Cal. Error\strut};
\node[anchor=west, rotate=\xticklabelrotation] at (axis cs:5+\xticklabeloffsetx,0+\xticklabeloffset) {Class Balance\strut};
\node[anchor=west, rotate=\xticklabelrotation] at (axis cs:6+\xticklabeloffsetx,0+\xticklabeloffset) {Obj. Focus\strut};
\node[anchor=west, rotate=\xticklabelrotation] at (axis cs:7+\xticklabeloffsetx,0+\xticklabeloffset) {Shape Bias\strut};
\node[anchor=west, rotate=\xticklabelrotation] at (axis cs:8+\xticklabeloffsetx,0+\xticklabeloffset) {Parameters\strut};

\end{axis}
\end{tikzpicture}
 \end{subfigure}\\
\caption{\textit{Rank correlation matrix for the considered quality dimensions among our full model zoo, comparing our adversarial robustness protocol to \emph{AutoAttack}.} We investigate differences of the rank correlations among all our models with our adversarial robustness metric \textit{(left)} and AutoAttack \textit{(right)}. Crossed-out entries indicate a $p$-value above 0.05, and thus, are not statistically significant. All the statistically significant conclusions remain unchanged.}
\label{fig:adv_rob_vs_autoattack}
\end{figure*}

\begin{table*}[t!]
  \centering
  \tiny
    \caption{\textit{Average adversarial robustness and AutoAttack for models with different configurations.} We evaluate various setups, focusing on different training strategies (a--g) and different architectural choices (h--j). In each configuration, we report the average score for both metrics across the models associated with that configuration.
  As different models are available for different setups, each setup considers a distinct selection of models being compared. As a result, both the models and their total number (indicated by the number beside each setup) vary across setups to maintain a fair basis for comparison.
  The number of asterisks represents the statistical significance of differences in the average scores of a quality dimension across configurations within each setup:~$^{***}$~for~$p < 0.05$,~$^{**}$~for $p < 0.1$, and $^{*}$ for $p < 0.2$, based on $t$-test results.
  }

  \begin{tabularx}{\linewidth}{
  @{}
  p{0.1cm}
  p{0.1cm}
  X
  S[table-format=1.2, table-space-text-post ={*}, table-space-text-pre ={+}]
  S[table-format=1.2, table-space-text-post ={*}, table-space-text-pre ={+}]
  }%
    \toprule
    \textbf{Setup} & 
    &
    \textbf{Configuration}  & {\tablearrowcenter\makecell{\textbf{Adv.} \\ \textbf{Rob.}}$\uparrow$} & {\tablearrowcenter\makecell{\textbf{Auto-} \\ \textbf{Attack}}$\uparrow$}\\
\midrule
\tablesetupcnndataset\ & 14 & CNNs (IN-1k)& 0.14 & 0.03 \\
&& CNNs (IN-21k)& \bfseries 0.15 & \bfseries 0.04\\
\midrule
\tablesetuptransdataset\ & 16 & Transformers (IN-1k)& \bfseries 0.21& \bfseries 0.09\\
&& Transformers (IN-21k)& 0.17$^{*}$ & 0.05$^{**}$ \\
\midrule
\tablesetupat\ & 11 & Supervised models& 0.12& 0.04 \\
&& Adversarially trained models& \bfseries 0.52$^{***}$  & \bfseries 0.68$^{***}$\\
\midrule
\tablesetupselfsllp\ & 13 & Supervised models& \bfseries 0.20& \bfseries 0.08\\
&& Self-supervised models (LP)& 0.10$^{***}$& 0.06 \\
\midrule
\tablesetupselfsletoe\ & 25 & Supervised models& 0.16& 0.07\\
&& Self-supervised models (E2E)& \bfseries 0.24$^{***}$ & 0.07\\
\midrule
\tablesetupsemisl\ & 13 & Supervised models& 0.13& 0.04\\
&& Semi-supervised models& \bfseries 0.20$^{***}$  & 0.04\\
\midrule
\tablesetuplong\ & 19 & Supervised models& 0.12& 0.03\\
& & A1 supervised models (600 epochs)& \bfseries 0.47$^{***}$ & \bfseries 0.05$^{***}$\\
& & A2 supervised models (300 epochs)& 0.41$^{***}$& \bfseries 0.05$^{***}$\\
& & A3 supervised models (100 epochs)& 0.32$^{***}$ & \bfseries 0.05$^{***}$\\
\midrule
\tablesetupcnnvstrans\ & 46 & CNNs& 0.11 & 0.11\\
& & Transformers& \bfseries 0.20& \bfseries 0.20\\
\midrule
\tablesetupbcos\ & 12 & Standard models& \bfseries 0.07& \bfseries 0.03\\
&& B-cos models& 0.02$^{***}$  & 0.00$^{***}$\\
\midrule
\tablesetupzeroshot\ & 24 & Standard models& \bfseries 0.18& \bfseries 0.09\\
&& Vision-language models& 0.10$^{*}$  & 0.01$^{***}$\\

    \bottomrule
  \end{tabularx}

  \label{tab:comparisons_autoattack}
\end{table*}

\begin{table*}[t]
  \centering
  \tiny
      \caption{\textit{QUBA score and quality dimensions for the top five performing models using AutoAttack as metric for adversarial robustness.} The configuration lists the architecture, training dataset, and training paradigm. \textsuperscript{\textdagger} indicates models trained with knowledge distillation.}
  \setlength{\tabcolsep}{4.5pt}
  \setlength\extrarowheight{5pt}
  \begin{tabularx}{\textwidth}{@{}lp{2.55cm}S[table-format=1.2]S[table-format=1.2]S[table-format=1.2]S[table-format=1.2]S[table-format=1.2]S[table-format=1.4]S[table-format=1.2]S[table-format=1.2]S[table-format=1.2]S[table-format=2]@{}}%
    \toprule
    \makecell[l]{\textbf{Model}} & {\makecell[l]{\textbf{Configuration}}} & {\makecell{\textbf{QUBA} \\ \textbf{Score}}$\uparrow$} & {\textbf{Acc.}$\uparrow$} & {\makecell{\textbf{Adv.} \\ \textbf{Rob.}}$\uparrow$} & {\textbf{C-Rob.}$\uparrow$} & {\makecell{\textbf{OOD} \\ \textbf{Rob.}}$\uparrow$} & {\makecell{\textbf{Cal.} \\ \textbf{Error}}$\downarrow$} & {\makecell{\textbf{Class} \\ \textbf{Balance}}$\uparrow$} & {\makecell{\textbf{Obj.} \\ \textbf{Focus}}$\uparrow$} & {\makecell{\textbf{Shape} \\ \textbf{Bias}}$\uparrow$} & {\makecell{\textbf{Params.} \\ \textbf{in Mil.}}$\downarrow$} \\
\midrule
\makecell[l]{ViT-S \\ \citep{Singh:2023:RAT}} & \makecell[l]{Transformer, IN1k, AT}& 1.27& 0.73& 0.66& 0.63& 0.72& 0.0081& 0.78& 0.94& 0.72& \bfseries 22\\
\makecell[l]{Swin-B \\ \citep{Liu:2023:CSR}} & \makecell[l]{Transformer, IN1k, AT} & 1.39& 0.77& \bfseries 0.74& 0.65& 0.74& 0.0083& 0.79& 0.94& 0.73& 87\\
\makecell[l]{ViT-B \\ \cite{Singh:2023:RAT}} & \makecell[l]{Transformer, IN1k, AT} & 1.44& 0.77& 0.71& \bfseries 0.69& 0.50& 0.0072& 0.79& \bfseries 0.95& \bfseries 0.77& 87\\
\makecell[l]{ConvNeXt-B \\ \cite{Liu:2023:CSR}} & \makecell[l]{CNN, IN1k, AT} & 1.45& \bfseries 0.77& 0.73& 0.64& 0.62& 0.0075& \bfseries 0.79& 0.95& 0.73& 88\\
\makecell[l]{ConNeXt-B \\ \cite{Singh:2023:RAT}} & \makecell[l]{CNN, IN1k, AT} & \bfseries 1.45& 0.76& 0.72& 0.64& \bfseries 0.80& \bfseries 0.0069& 0.78& 0.95& 0.74& 88\\

    \bottomrule
  \end{tabularx}

  \label{tab:top5_autoattack}
\end{table*}

\section{Experimental details}\label{sec:experimental_details}

The code provided in the supplemental material gives detailed instructions on how to use our proposed benchmark and how to reproduce our main results. To include new models, one only needs to add the respective model file and weights. 

To ensure comprehensive comparisons in future work, we publish the code and results of our model zoo (\url{https://visinf.github.io/beyond-accuracy}). This will allow practitioners to evaluate and compare the considered quality dimensions easily for their model. In the next subsections, we will describe additional experimental details.

\subsection{Comparisons}\label{appendix:sec:details_comparison}

In \cref{tab:comparisons} of the main paper, we compare the average of each quality dimension for different setups. 
We report the models that have been used for each comparison in \cref{tab:details_comparison}.

\onecolumn
{\tiny
\begin{longtable}{@{}lllp{10cm}@{}}
\caption{\textit{Models used for each comparison in \cref{tab:comparisons} of the main paper.} In some setups, the same model appears multiple times within a configuration because the corresponding configuration contains multiple models that need to be compared to that single model (\eg, there might be two different ViL models using the same backbone). This duplication ensures that the final average accounts for these cases correctly.
} \label{tab:details_comparison} \\
\toprule
\textbf{Setup} & \textbf{\# of Models} & \textbf{Configuration} &  \textbf{Models}\\
\midrule
\endfirsthead

\tablesetupcnndataset & 14 & CNNs (IN-1k)& MobileNetV3-l~\citep{Howard:2019:SFM}, ResNet50~\citep{He:2016:DRL}, ResNet101~\citep{He:2016:DRL}, EfficientNet-v2-S~\citep{Tan:2021:ESM}, EfficientNet-v2-M~\citep{Tan:2021:ESM}, EfficientNet-v2-L~\citep{Tan:2021:ESM}, ConvNeXt-T~\citep{Liu:2022:ACF}, ConvNeXt-S~\citep{Liu:2022:ACF}, ConvNeXt-B~\citep{Liu:2022:ACF}, ConvNeXt-L~\citep{Liu:2022:ACF}, ConvNeXtV2-N~\citep{Woo:2023:CCS}, ConvNeXtV2-T~\citep{Woo:2023:CCS}, ConvNeXtV2-B~\citep{Woo:2023:CCS}, ConvNeXtV2-L~\citep{Woo:2023:CCS}\\
&& CNNs (IN-21k)& MobileNetV3-l~\citep{Howard:2019:SFM}, BiTM-ResNet50x1~\citep{Kolesnikov:2020:BIT}, BiTM-ResNet101x1~\citep{Kolesnikov:2020:BIT}, EfficientNet-v2-S~\citep{Tan:2021:ESM}, EfficientNet-v2-M~\citep{Tan:2021:ESM}, EfficientNet-v2-L~\citep{Tan:2021:ESM}, ConvNeXt-T~\citep{Liu:2022:ACF}, ConvNeXt-S~\citep{Liu:2022:ACF}, ConvNeXt-B~\citep{Liu:2022:ACF}, ConvNeXt-L~\citep{Liu:2022:ACF}, ConvNeXtV2-N~\citep{Woo:2023:CCS}, ConvNeXtV2-T~\citep{Woo:2023:CCS}, ConvNeXtV2-B~\citep{Woo:2023:CCS}, ConvNeXtV2-L~\citep{Woo:2023:CCS}\\
\midrule
\tablesetuptransdataset\ & 16 & Transformers (IN-1k) & DeiT3-s~\citep{Touvron:2022:DRO}, DeiT3-m~\citep{Touvron:2022:DRO}, DeiT3-b~\citep{Touvron:2022:DRO}, DeiT3-l~\citep{Touvron:2022:DRO}, SwinV2-b/16~\citep{Liu:2022:STS}, TinyViT-5M/16~\citep{Wu:2022:TFP}, TinyViT-11M/16~\citep{Wu:2022:TFP}, TinyViT-21M/16~\citep{Wu:2022:TFP}, DeiT-s~\citep{Touvron:2021:TDE}, ViT-b/16~\citep{Dosovitskiy:2021:IWW}, ViT-b/32~\citep{Dosovitskiy:2021:IWW}, ViT-l/16~\citep{Dosovitskiy:2021:IWW}, ViT-b/32~\citep{Dosovitskiy:2021:IWW}, ViT-b/16~\citep{Dosovitskiy:2021:IWW}, ViT-b/16~\citep{Dosovitskiy:2021:IWW}, DeiT-s~\citep{Touvron:2021:TDE}\\
&& Transformers (IN-21k)& DeiT3-s~\citep{Touvron:2022:DRO}, DeiT3-m~\citep{Touvron:2022:DRO}, DeiT3-b~\citep{Touvron:2022:DRO}, DeiT3-l~\citep{Touvron:2022:DRO}, SwinV2-b/12to16~\citep{Liu:2022:STS}, TinyViT-5M/16~\citep{Wu:2022:TFP}, TinyViT-11M/16~\citep{Wu:2022:TFP}, TinyViT-21M/16~\citep{Wu:2022:TFP}, ViT-s/16~\citep{Steiner:2022:HTT}, ViT-b/16~\citep{Steiner:2022:HTT}, ViT-l/16~\citep{Steiner:2022:HTT}, ViT-b/32~\citep{Steiner:2022:HTT}, ViT-l/16~\citep{Steiner:2022:HTT}, BeiT-b~\citep{Bao:2022:BBP}, EVA02-B/14~\citep{Fang:2023:EVA}, EVA02-S/14~\citep{Fang:2023:EVA}\\
\midrule
\tablesetupat\ & 11 & Supervised models& WRN-50-2~\citep{Zagoruyko:2016:WRN}, ResNet50~\citep{He:2016:DRL}, Swin-b~\citep{Liu:2021:STH}, ConvNeXt-B~\citep{Liu:2022:ACF}, ConvNeXt-L~\citep{Liu:2022:ACF}, ConvNeXt-T~\citep{Liu:2022:ACF}, ConvNeXt-S~\citep{Liu:2022:ACF}, ConvNeXt-B~\citep{Liu:2022:ACF}, ConvNeXt-L~\citep{Liu:2022:ACF}, ViT-b/16~\citep{Dosovitskiy:2021:IWW}, DeiT-s~\citep{Touvron:2021:TDE}\\
&& Adversarially trained models& WRN-50-2~\citep{Salman:2020:ARI}, ResNet50~\citep{Salman:2020:ARI}, Swin-b~\citep{Liu:2023:CSR}, ConvNeXt-B~\citep{Liu:2023:CSR}, ConvNeXt-L~\citep{Liu:2023:CSR}, ConvNeXt-T~\citep{Singh:2023:RAT}, ConvNeXt-S~\citep{Singh:2023:RAT}, ConNeXt-B~\citep{Singh:2023:RAT}, ConvNeXt-L~\citep{Singh:2023:RAT}, ViT-b/16~\citep{Singh:2023:RAT}, ViT-s/16~\citep{Singh:2023:RAT}\\
\midrule
\tablesetupselfsllp\ & 13 & Supervised models& MViTv2-b~\citep{Li:2022:MIM}, ViT-b/16~\citep{Dosovitskiy:2021:IWW}, ResNet50~\citep{He:2016:DRL}, ViT-b/16~\citep{Dosovitskiy:2021:IWW}, ViT-l/16~\citep{Dosovitskiy:2021:IWW}, ViT-b/16~\citep{Dosovitskiy:2021:IWW}, DeiT-s~\citep{Touvron:2021:TDE}, DeiT-s~\citep{Touvron:2021:TDE}, Hiera-T~\citep{Ryali:2023:HHV}, Hiera-S~\citep{Ryali:2023:HHV}, ViT-l/16~\citep{Dosovitskiy:2021:IWW}, ViT-b/16~\citep{Dosovitskiy:2021:IWW}, DeiT-s~\citep{Touvron:2021:TDE}\\
&& Self-supervised models (LP)& Hiera-B~\citep{Ryali:2023:HHV}, ViT-b/16-MAE~\citep{He:2022:MAA}, ResNet50-DINO~\citep{Caron:2021:EPS}, ViT-b/16-DINO~\citep{Caron:2021:EPS}, ViT-l/14-DINOv2-LP~\citep{Oquab:2024:DIN}, ViT-b/14-DINOv2-LP~\citep{Oquab:2024:DIN}, ViT-s/14-DINOv2-LP~\citep{Oquab:2024:DIN}, ViT-s/16-DINO~\citep{Caron:2021:EPS}, Hiera-T~\citep{Ryali:2023:HHV}, Hiera-S~\citep{Ryali:2023:HHV}, ViT-l/14-DINOv2-reg-LP~\citep{Darcet:2024:VTN}, ViT-b/14-DINOv2-reg-LP~\citep{Darcet:2024:VTN}, ViT-s/14-DINOv2-reg-LP~\citep{Darcet:2024:VTN}\\
\midrule
\tablesetupselfsletoe\ & 25 & Supervised models& MViTv2-b~\citep{Li:2022:MIM}, ViT-b/16~\citep{Dosovitskiy:2021:IWW}, ResNet50~\citep{He:2016:DRL}, ViT-b/16~\citep{Dosovitskiy:2021:IWW}, ConvNeXt-T~\citep{Liu:2022:ACF}, ConvNeXt-B~\citep{Liu:2022:ACF}, ConvNeXt-L~\citep{Liu:2022:ACF}, ConvNeXt-T~\citep{Liu:2022:ACF}, ConvNeXt-B~\citep{Liu:2022:ACF}, ConvNeXt-L~\citep{Liu:2022:ACF}, ViT-b/16~\citep{Steiner:2022:HTT}, ViT-b/16~\citep{Dosovitskiy:2021:IWW}, ViT-b/16~\citep{Steiner:2022:HTT}, ViT-s/16~\citep{Steiner:2022:HTT}, ViT-t/16~\citep{Steiner:2022:HTT}, ViT-l/16~\citep{Dosovitskiy:2021:IWW}, ViT-b/16~\citep{Dosovitskiy:2021:IWW}, DeiT-s~\citep{Touvron:2021:TDE}, ViT-l/16~\citep{Dosovitskiy:2021:IWW}, ViT-b/16~\citep{Dosovitskiy:2021:IWW}, DeiT-s~\citep{Touvron:2021:TDE}, ViT-b/16~\citep{Dosovitskiy:2021:IWW}, ViT-b/16~\citep{Dosovitskiy:2021:IWW}, ViT-b/32~\citep{Dosovitskiy:2021:IWW}, ViT-b/32~\citep{Dosovitskiy:2021:IWW}\\
&& Self-supervised models (E2E)& Hiera-B~\citep{Ryali:2023:HHV}, ViT-b/16-MAE~\citep{He:2022:MAA}, ResNet50-DINO~\citep{Caron:2021:EPS}, ViT-b/16-DINO~\citep{Caron:2021:EPS}, ConvNeXtV2-T~\citep{Woo:2023:CCS}, ConvNeXtV2-B~\citep{Woo:2023:CCS}, ConvNeXtV2-L~\citep{Woo:2023:CCS}, ConvNeXtV2-T~\citep{Woo:2023:CCS}, ConvNeXtV2-B~\citep{Woo:2023:CCS}, ConvNeXtV2-L~\citep{Woo:2023:CCS}, BeiT-b~\citep{Bao:2022:BBP}, BeiTV2-b~\citep{Peng:2022:BMI}, EVA02-S/14~\citep{Fang:2023:EVA}, EVA02-B/14~\citep{Fang:2023:EVA}, EVA02-T/14~\citep{Fang:2023:EVA}, ViT-l/14-DINOv2-FT~\citep{Oquab:2024:DIN}, ViT-b/14-DINOv2-FT~\citep{Oquab:2024:DIN}, ViT-s/14-DINOv2-FT~\citep{Oquab:2024:DIN}, ViT-l/14-DINOv2-reg-LP~\citep{Darcet:2024:VTN}, ViT-b/14-DINOv2-reg-LP~\citep{Darcet:2024:VTN}, ViT-s/14-DINOv2-reg-LP~\citep{Darcet:2024:VTN}, CLIP-B/16-OpenAI-FT-Vision-Encoder~\citep{Cherti:2023:RSL}, CLIP-B/16-Laion2B-FT-Vision-Encoder~\citep{Cherti:2023:RSL}, CLIP-B/32-OpenAI-FT-Vision-Encoder~\citep{Cherti:2023:RSL}, CLIP-B/32-Laion2B-FT-Vision-Encoder~\citep{Cherti:2023:RSL}\\
\midrule
\tablesetupsemisl\ & 13 & Supervised models& EfficientNet-B0~\citep{Tan:2019:ERM}, EfficientNet-B1~\citep{Tan:2019:ERM}, EfficientNet-B2~\citep{Tan:2019:ERM}, EfficientNet-B3~\citep{Tan:2019:ERM}, EfficientNet-B4~\citep{Tan:2019:ERM}, EfficientNet-B5~\citep{Tan:2019:ERM}, EfficientNet-B6~\citep{Tan:2019:ERM}, EfficientNet-B7~\citep{Tan:2019:ERM}, ResNet50~\citep{He:2016:DRL}, ResNeXt50-32x4d~\citep{Xie:2017:ART}, ResNet18~\citep{He:2016:DRL}, ResNeXt101-32x8d~\citep{Xie:2017:ART}, ResNeXt50-32x4d~\citep{Xie:2017:ART}\\
&& Semi-supervised models& EfficientNet-B0~\citep{Xie:2020:STW}, EfficientNet-B1~\citep{Xie:2020:STW}, EfficientNet-B2~\citep{Xie:2020:STW}, EfficientNet-B3~\citep{Xie:2020:STW}, EfficientNet-B4~\citep{Xie:2020:STW}, EfficientNet-B5~\citep{Xie:2020:STW}, EfficientNet-B6~\citep{Xie:2020:STW}, EfficientNet-B7~\citep{Xie:2020:STW}, ResNet50~\citep{Yalniz:2019:BSS}, ResNeXt50-32x4d~\citep{Yalniz:2019:BSS}, ResNet18~\citep{Yalniz:2019:BSS}, ResNeXt101-32x8d~\citep{Yalniz:2019:BSS}, ResNeXt50-32x4d~\citep{Yalniz:2019:BSS}\\
\midrule
\newpage
\midrule
\tablesetuplong\ & 19 & Supervised models& EfficientNet-v2-M~\citep{Tan:2021:ESM}, EfficientNet-v2-S~\citep{Tan:2021:ESM}, SeNet154~\citep{Hu:2018:SEN}, RegNet-y-8gf~\citep{Radosavovic:2020:DND}, RegNet-y-4gf~\citep{Radosavovic:2020:DND}, RegNet-y-16gf~\citep{Radosavovic:2020:DND}, RegNet-y-32gf~\citep{Radosavovic:2020:DND}, ResNet101~\citep{He:2016:DRL}, ResNet18~\citep{He:2016:DRL}, ResNet152~\citep{He:2016:DRL}, ResNet34~\citep{He:2016:DRL}, ResNet50d~\citep{He:2019:BOT}, ResNet50~\citep{He:2016:DRL}, ResNeXt50-32x4d~\citep{Xie:2017:ART}, EfficientNet-B0~\citep{Tan:2019:ERM}, EfficientNet-B1~\citep{Tan:2019:ERM}, EfficientNet-B2~\citep{Tan:2019:ERM}, EfficientNet-B3~\citep{Tan:2019:ERM}, EfficientNet-B4~\citep{Tan:2019:ERM}\\
& & A1 supervised models& EfficientNet-v2-M~\citep{Wightman:2021:RSB}, EfficientNet-v2-S~\citep{Wightman:2021:RSB}, SeNet154~\citep{Wightman:2021:RSB}, RegNet-y-4gf~\citep{Wightman:2021:RSB}, RegNet-y-8gf~\citep{Wightman:2021:RSB}, RegNet-y-16gf~\citep{Wightman:2021:RSB}, RegNet-y-32gf~\citep{Wightman:2021:RSB}, ResNet101~\citep{Wightman:2021:RSB}, ResNet18~\citep{Wightman:2021:RSB}, ResNet152~\citep{Wightman:2021:RSB}, ResNet34~\citep{Wightman:2021:RSB}, ResNet50d~\citep{Wightman:2021:RSB}, ResNet50~\citep{Wightman:2021:RSB}, ResNeXt50-32x4d~\citep{Wightman:2021:RSB}, EfficientNet-B0~\citep{Wightman:2021:RSB}, EfficientNet-B1~\citep{Wightman:2021:RSB}, EfficientNet-B2~\citep{Wightman:2021:RSB}, EfficientNet-B3~\citep{Wightman:2021:RSB}, EfficientNet-B4~\citep{Wightman:2021:RSB}\\
& & A2 supervised models& EfficientNet-v2-M~\citep{Wightman:2021:RSB}, EfficientNet-v2-S~\citep{Wightman:2021:RSB}, SeNet154~\citep{Wightman:2021:RSB}, RegNet-y-4gf~\citep{Wightman:2021:RSB}, RegNet-y-8gf~\citep{Wightman:2021:RSB}, RegNet-y-16gf~\citep{Wightman:2021:RSB}, RegNet-y-32gf~\citep{Wightman:2021:RSB}, ResNet101~\citep{Wightman:2021:RSB}, ResNet18~\citep{Wightman:2021:RSB}, ResNet152~\citep{Wightman:2021:RSB}, ResNet34~\citep{Wightman:2021:RSB}, ResNet50d~\citep{Wightman:2021:RSB}, ResNet50~\citep{Wightman:2021:RSB}, ResNeXt50-32x4d~\citep{Wightman:2021:RSB}, EfficientNet-B0~\citep{Wightman:2021:RSB}, EfficientNet-B1~\citep{Wightman:2021:RSB}, EfficientNet-B2~\citep{Wightman:2021:RSB}, EfficientNet-B3~\citep{Wightman:2021:RSB}, EfficientNet-B4~\citep{Wightman:2021:RSB}\\
& & A3 supervised models& EfficientNet-v2-M~\citep{Wightman:2021:RSB}, EfficientNet-v2-S~\citep{Wightman:2021:RSB}, SeNet154~\citep{Wightman:2021:RSB}, RegNet-y-4gf~\citep{Wightman:2021:RSB}, RegNet-y-8gf~\citep{Wightman:2021:RSB}, RegNet-y-16gf~\citep{Wightman:2021:RSB}, RegNet-y-32gf~\citep{Wightman:2021:RSB}, ResNet101~\citep{Wightman:2021:RSB}, ResNet18~\citep{Wightman:2021:RSB}, ResNet152~\citep{Wightman:2021:RSB}, ResNet34~\citep{Wightman:2021:RSB}, ResNet50d~\citep{Wightman:2021:RSB}, ResNet50~\citep{Wightman:2021:RSB}, ResNeXt50-32x4d~\citep{Wightman:2021:RSB}, EfficientNet-B0~\citep{Wightman:2021:RSB}, EfficientNet-B1~\citep{Wightman:2021:RSB}, EfficientNet-B2~\citep{Wightman:2021:RSB}, EfficientNet-B3~\citep{Wightman:2021:RSB}, EfficientNet-B4~\citep{Wightman:2021:RSB}\\
\midrule
\tablesetupcnnvstrans\ & 46 & CNNs& RegNet-y-400mf~\citep{Radosavovic:2020:DND}, RegNet-y-800mf~\citep{Radosavovic:2020:DND}, RegNet-y-1-6gf~\citep{Radosavovic:2020:DND}, RegNet-y-3-2gf~\citep{Radosavovic:2020:DND}, RegNet-y-8gf~\citep{Radosavovic:2020:DND}, RegNet-y-16gf~\citep{Radosavovic:2020:DND}, EfficientNet-v2-S~\citep{Tan:2021:ESM}, EfficientNet-v2-L~\citep{Tan:2021:ESM}, ConvNeXt-T~\citep{Liu:2022:ACF}, ConvNeXt-S~\citep{Liu:2022:ACF}, ConvNeXt-B~\citep{Liu:2022:ACF}, RegNet-y-4gf~\citep{Radosavovic:2020:DND}, ConvNeXt-B~\citep{Liu:2022:ACF}, RegNet-y-800mf~\citep{Radosavovic:2020:DND}, ConvNeXt-T~\citep{Liu:2022:ACF}, ConvNeXt-T~\citep{Liu:2022:ACF}, RegNet-y-1-6gf~\citep{Radosavovic:2020:DND}, RegNet-y-800mf~\citep{Radosavovic:2020:DND}, RegNet-y-1-6gf~\citep{Radosavovic:2020:DND}, EfficientNet-v2-S~\citep{Tan:2021:ESM}, RegNet-y-8gf~\citep{Radosavovic:2020:DND}, ConvNeXt-T~\citep{Liu:2022:ACF}, ConvNeXt-S~\citep{Liu:2022:ACF}, ConvNeXt-B~\citep{Liu:2022:ACF}, ConvNeXt-T~\citep{Liu:2022:ACF}, ConvNeXt-S~\citep{Liu:2022:ACF}, ConvNeXt-B~\citep{Liu:2022:ACF}, RegNet-y-800mf~\citep{Radosavovic:2020:DND}, RegNet-y-1-6gf~\citep{Radosavovic:2020:DND}, EfficientNet-v2-S~\citep{Tan:2021:ESM}, EfficientNet-v2-S~\citep{Tan:2021:ESM}, ConvNeXt-T~\citep{Liu:2022:ACF}, ConvNeXt-S~\citep{Liu:2022:ACF}, ConvNeXt-B~\citep{Liu:2022:ACF}, ConvNeXt-T~\citep{Liu:2022:ACF}, ConvNeXt-S~\citep{Liu:2022:ACF}, RegNet-y-1-6gf~\citep{Radosavovic:2020:DND}, EfficientNet-v2-S~\citep{Tan:2021:ESM}, EfficientNet-v2-S~\citep{Tan:2021:ESM}, ConvNeXt-B~\citep{Liu:2022:ACF}, EfficientNet-v2-S~\citep{Tan:2021:ESM}, EfficientNet-v2-S~\citep{Tan:2021:ESM}, ConvNeXt-B~\citep{Liu:2022:ACF}, ConvNeXt-B~\citep{Liu:2023:CSR}, ConNeXt-B~\citep{Singh:2023:RAT}, ConvNeXtV2-T~\citep{Woo:2023:CCS}\\
& & Transformers& PiT-t~\citep{Heo:2021:RSD}, DeiT-t~\citep{Touvron:2021:TDE}, CaiT-xxs24~\citep{Touvron:2021:GDW}, LeViT-256~\citep{Graham:2021:LAV} , LeViT-384~\citep{Graham:2021:LAV} , XCiT-m24-16~\citep{Ali:2021:XCC}, DeiT-s~\citep{Touvron:2021:TDE}, MaxViT-b~\citep{Tu:2022:MAV}, Swin-t~\citep{Liu:2021:STH}, Swin-s~\citep{Liu:2021:STH}, ViT-b/32~\citep{Dosovitskiy:2021:IWW}, CoaT-s-lite~\citep{Xu:2021:CSC}, Swin-b~\citep{Liu:2021:STH}, ConViT-t~\citep{Ascoli:2021:CIV}, ConViT-s~\citep{Ascoli:2021:CIV}, CrossViT-15\textdagger ~\citep{Chen:2021:CCA}, PiT-xs~\citep{Heo:2021:RSD}, CoaT-t-lite~\citep{Xu:2021:CSC}, CoaT-mi-lite~\citep{Xu:2021:CSC}, DeiT3-s~\citep{Touvron:2022:DRO}, DeiT3-m~\citep{Touvron:2022:DRO}, SwinV2-t/8~\citep{Liu:2022:STS}, SwinV2-s/8~\citep{Liu:2022:STS}, SwinV2-b/8~\citep{Liu:2022:STS}, SwinV2-t/16~\citep{Liu:2022:STS}, SwinV2-s/16~\citep{Liu:2022:STS}, SwinV2-b/16~\citep{Liu:2022:STS}, TinyViT-5M/16~\citep{Wu:2022:TFP}, TinyViT-11M/16~\citep{Wu:2022:TFP}, TinyViT-21M/16~\citep{Wu:2022:TFP}, ViT-s/16~\citep{Steiner:2022:HTT}, DaViT-t~\citep{Ding:2022:DAV}, DaViT-s~\citep{Ding:2022:DAV}, DaViT-b~\citep{Ding:2022:DAV}, InceptionNext-t~\citep{Yu:2023:IWI}, InceptionNext-s~\citep{Yu:2023:IWI}, FastViT-sa12~\citep{Vasu:2023:FAF}, FastViT-sa24~\citep{Vasu:2023:FAF}, DeiT3-s~\citep{Touvron:2022:DRO}, SwinV2-b/12to16~\citep{Liu:2022:STS}, TinyViT-21M/16~\citep{Wu:2022:TFP}, ViT-s/16~\citep{Steiner:2022:HTT}, ViT-l/16~\citep{Steiner:2022:HTT}, Swin-b~\citep{Liu:2023:CSR}, Swin-b~\citep{Liu:2023:CSR}, Hiera-T~\citep{Ryali:2023:HHV}\\
\midrule
\tablesetupbcos\ & 12 & Standard models& ResNet18~\citep{He:2016:DRL}, ResNet34~\citep{He:2016:DRL}, ResNet50~\citep{He:2016:DRL}, ResNet152~\citep{He:2016:DRL}, ResNet101~\citep{He:2016:DRL}, DenseNet121~\citep{Huang:2017:DCC}, DenseNet161~\citep{Huang:2017:DCC}, DenseNet169~\citep{Huang:2017:DCC}, DenseNet201~\citep{Huang:2017:DCC}, ViT-b/16~\citep{Dosovitskiy:2021:IWW}, ConvNeXt-T~\citep{Liu:2022:ACF}, ConvNeXt-B~\citep{Liu:2022:ACF}\\
&& B-cos models& ResNet18~\citep{Boehle:2022:BCN}, ResNet34~\citep{Boehle:2022:BCN}, ResNet50~\citep{Boehle:2022:BCN}, ResNet101~\citep{Boehle:2022:BCN}, ResNet152~\citep{Boehle:2022:BCN}, DenseNet121~\citep{Boehle:2022:BCN}, DenseNet161~\citep{Boehle:2022:BCN}, DenseNet169~\citep{Boehle:2022:BCN}, DenseNet201~\citep{Boehle:2022:BCN}, ViT-b/16~\citep{Boehle:2022:BCN}, ConNeXt-B~\citep{Boehle:2022:BCN}, ConvNeXt-T~\citep{Boehle:2022:BCN}\\
\midrule
\tablesetupzeroshot\ & 24 & Standard models& ResNet50~\citep{He:2016:DRL}, ResNet101~\citep{He:2016:DRL}, ViT-b/16~\citep{Dosovitskiy:2021:IWW}, ViT-b/32~\citep{Dosovitskiy:2021:IWW}, FastViT-sa12~\citep{Vasu:2023:FAF}, FastViT-sa24~\citep{Vasu:2023:FAF}, FastViT-sa36~\citep{Vasu:2023:FAF}, ViT-b/16~\citep{Dosovitskiy:2021:IWW}, ViT-b/16~\citep{Dosovitskiy:2021:IWW}, ConvNeXt-L~\citep{Liu:2022:ACF}, ViT-l/16~\citep{Dosovitskiy:2021:IWW}, ViT-l/16~\citep{Dosovitskiy:2021:IWW}, ConvNeXt-B~\citep{Liu:2022:ACF}, ConvNeXt-L~\citep{Liu:2022:ACF}, ConvNeXt-L~\citep{Liu:2022:ACF}, ViT-b/16~\citep{Dosovitskiy:2021:IWW}, ViT-b/16~\citep{Dosovitskiy:2021:IWW}, ViT-b/16~\citep{Dosovitskiy:2021:IWW}, ViT-l/16~\citep{Dosovitskiy:2021:IWW}, ViT-l/16~\citep{Dosovitskiy:2021:IWW}, ViT-l/16~\citep{Dosovitskiy:2021:IWW}, ViT-l/16~\citep{Dosovitskiy:2021:IWW}, ViT-b/16~\citep{Dosovitskiy:2021:IWW}, ViT-l/16~\citep{Dosovitskiy:2021:IWW}\\
&& Vision-language models& CLIP-ResNet50~\citep{Radford:2021:LTV}, CLIP-ResNet101~\citep{Radford:2021:LTV}, CLIP-B/16~\citep{Radford:2021:LTV}, CLIP-B/32~\citep{Radford:2021:LTV}, MobileCLIP-S0~\citep{Vase:2024:MFI}, MobileCLIP-S1~\citep{Vase:2024:MFI}, MobileCLIP-S2~\citep{Vase:2024:MFI}, MobileCLIP-B~\citep{Vase:2024:MFI}, MobileCLIP-B (LT)~\citep{Vase:2024:MFI}, SigLIP-l/16~\citep{Zhai:2023:SLF}, MetaCLIP-B/16~\citep{Xu:2024:DCD}, MetaCLIP-L/14~\citep{Xu:2024:DCD}, CLIP-ConvNeXt-B-320px~\citep{Schuhmann:2022:AOL, Wortsman:2022:MSA}, CLIP-ConvNeXt-L-320px~\citep{Schuhmann:2022:AOL,  Wortsman:2022:MSA}, CLIP-ConvNeXt-L~\citep{Schuhmann:2022:AOL, Wortsman:2022:MSA}, CLIP-B/16-DataCompXL~\citep{Gadre:2023:DIS}, CLIP-B/16-Laion2B~\citep{Schuhmann:2022:AOL,  Wortsman:2022:MSA}, CLIP-B/16-CommonPool-XL-DFN2B~\citep{Fang:2024:DFN}, CLIP-L/14-OpenAI~\citep{Radford:2021:LTV}, CLIP-L/14-DataCompXL~\citep{Gadre:2023:DIS}, CLIP-L/14-Laion2B~\citep{Schuhmann:2022:AOL,  Wortsman:2022:MSA}, CLIP-L/14-CommPool XL-DFN2B~\citep{Fang:2024:DFN}, SigLIP2-b/16~\citep{Tschannen:2025:SMV}, SigLIP2-l/16~\citep{Tschannen:2025:SMV}\\

    \bottomrule
  
\end{longtable}
}

\subsection{Training self-supervised models}\label{sec:details_selfsl}

To increase the number of models we can use to compare self-supervised and supervised models, we trained additional models for the end-to-end fine-tuning setup (E2E) and the linear probing setup (LP). For the E2E setup, we fine-tune the official pre-trained DINO~\citep{Caron:2021:EPS} checkpoints for ResNet50~\citep{He:2016:DRL} and ViT-b/16~\citep{Dosovitskiy:2021:IWW}. For both models, we follow the training procedure of \citet{He:2022:MAA}, as also done in \citet{Goldblum:2023:BBL}. Specifically, we train the ResNet50 for \num{100} epochs with an AdamW \citep{Loshchilov:2017:FWD} optimizer using a batch size of \num{128}, a weight decay of \num{0.05}, and a learning rate of \num{0.001} with a cosine scheduler. We use \num{5} warm-up epochs with a learning rate of \num{0.0001}. For the ViT-b/16, we follow the same procedure but with a learning rate of \num{0.008}. Furthermore, we also fine-tuned the official pre-trained DINOv2 \citep{Oquab:2024:DIN} checkpoints for ViT-s/14, ViT-b/14, and ViT-l/14 as well as their respective version with register tokens \citep{Darcet:2024:VTN}. For training all DINOv2 models, we follow \citet{Touvron:2022:DRO} as done in \citet{Oquab:2024:DIN}. To be more precise, we trained each model for \num{50} epochs with a batch size of \num{128} for the small variant and a batch size of \num{64} for the base and large variants. We used an AdamW \citep{Loshchilov:2017:FWD} optimizer, a weight decay of \num{0.02}, and a learning rate of \num{0.0003}. Additionally, we use \num{5} warm-up epochs with a learning rate of $10^{-6}$. We also use the ThreeAugment augmentation method that was introduced in \citep{Touvron:2022:DRO} together with color jitter and CutMix \citep{Yun:2019:CRS}.

For the LP setup, we train four additional models, ViT-b/16 (with Masked Autoencoder~\citep{He:2022:MAA} pre-training) and Hiera-tiny/small/base~\citep{Ryali:2023:HHV}. We only train the classification head of each model and follow the training of \citet{He:2022:MAA}, consistent with \citet{Ryali:2023:HHV}. We train a linear layer for \num{90} epochs with a LARS~\citep{You:2017:LBT} optimizer using a learning rate of \num{0.2} and a batch size of \num{512}.

\section{Numerical results}\label{sec:numerical_results}

\subsection{Correlation matrix in the main paper}\label{sec:num_results_corr_matrix}

As our visualization of the correlation matrix in \cref{fig:correlation_matrix} of the main paper summarizes much information and partly relies on color vision, we additionally report the numerical results in \cref{tab:correlation_p_all}.

\begin{table}
\tiny
  \caption{\textit{Correlation matrix with corresponding p-values (in parentheses) for all models (numerical results for \cref{fig:correlation_matrix} in the main paper).}}
  \centering
  \addtolength{\tabcolsep}{-0.3em}
  \setlength\extrarowheight{5pt}
  \begin{tabularx}{\linewidth}{@{}Xccccccccc@{}}
    \toprule
     & Acc. & \makecell{Adv. \\ Rob.} & C-Rob. & \makecell{OOD \\ Rob} & \makecell{Cal. \\ Error} & \makecell{Class \\ Balance} & \makecell{Obj. \\ Focus} & \makecell{Shape \\ Bias} & \makecell{Parameters} \\
    \midrule
Acc& $\makecell{1.00 \\ (.00)}$&&&&&&&&\\
Adv. Rob.& $\makecell{.44 \\ (.00)}$& $\makecell{1.00 \\ (.00)}$&&&&&&&\\
C-Rob.& $\makecell{.62 \\ (.00)}$& $\makecell{.01 \\ (.84)}$& $\makecell{1.00 \\ (.00)}$&&&&&&\\
OOD Rob.& $\makecell{.35 \\ (.00)}$& $\makecell{-0.08 \\ (.15)}$& $\makecell{.80 \\ (.00)}$& $\makecell{1.00 \\ (.00)}$&&&&&\\
Cal. Err.& $\makecell{-0.12 \\ (.03)}$& $\makecell{.07 \\ (.23)}$& $\makecell{.08 \\ (.16)}$& $\makecell{.25 \\ (.00)}$& $\makecell{1.00 \\ (.00)}$&&&&\\
Class Balance & $\makecell{.53 \\ (.00)}$& $\makecell{.08 \\ (.14)}$& $\makecell{.73 \\ (.00)}$& $\makecell{.76 \\ (.00)}$& $\makecell{.49 \\ (.00)}$& $\makecell{1.00 \\ (.00)}$&&&\\
Obj. Foc.& $\makecell{.72 \\ (.00)}$& $\makecell{.45 \\ (.00)}$& $\makecell{.64 \\ (.00)}$& $\makecell{.47 \\ (.00)}$& $\makecell{.09 \\ (.11)}$& $\makecell{.57 \\ (.00)}$& $\makecell{1.00 \\ (.00)}$&&\\
Shape Bias& $\makecell{.26 \\ (.00)}$& $\makecell{.17 \\ (.00)}$& $\makecell{.61 \\ (.00)}$& $\makecell{.62 \\ (.00)}$& $\makecell{.24 \\ (.00)}$& $\makecell{.54 \\ (.00)}$& $\makecell{.52 \\ (.00)}$& $\makecell{1.00 \\ (.00)}$&\\
Parameters& $\makecell{.31 \\ (.00)}$& $\makecell{.18 \\ (.00)}$& $\makecell{.44 \\ (.00)}$& $\makecell{.43 \\ (.00)}$& $\makecell{.11 \\ (.05)}$& $\makecell{.47 \\ (.00)}$& $\makecell{.48 \\ (.00)}$& $\makecell{.53 \\ (.00)}$& $\makecell{1.00 \\ (.00)}$\\

    \bottomrule
  \end{tabularx}
  \label{tab:correlation_p_all}
  \vspace{-0.5em}
\end{table}

\subsection{Mean and standard deviation of each quality dimension}\label{sec:mean_std_quality_dimension}

For our proposed QUBA score described in \cref{sec:experiments_rankingdetails} of the main paper, we compute a representative mean and standard deviation for each quality dimension. In \cref{tab:means_stds}, we report these values.

\begin{table}
\scriptsize
  \caption{\textit{Mean and standard deviation for each quality dimension as described in \cref{sec:experiments_rankingdetails} of the main paper.}}
  \centering
  \addtolength{\tabcolsep}{-0.3em}
  \begin{tabularx}{\linewidth}{@{}Xcc@{}}
    \toprule
     {\textbf{Quality Dimension}} & {\textbf{Mean}} & {\textbf{Standard Deviation}}\\
    \midrule
    {Accuracy} & 0.80 & 0.03\\
    {Adversarial Robustness} & 0.19 & 0.11\\
    {C-Robustness} & 0.53 & 0.23\\
    {OOD Robustness} & 0.57 & 0.15\\
    {Calibration Error} & 0.0045 & 0.0027\\
    {Class Balance} & 0.78 & 0.02\\
    {Object Focus} & 0.93 & 0.02\\
    {Shape Bias} & 0.31 & 0.08\\
    {Parameters in Mil.} & 55 & 43\\
    \bottomrule
  \end{tabularx}

  \label{tab:means_stds}
  \vspace{-0.5em}
\end{table}

\subsection{Model zoo}\label{sec:model_zoo}

We use \nrmodels models %
in our large-scale study. For a comprehensive overview of all models, their configuration (architecture, training dataset, and training paradigm), their QUBA score, and their scores for each quality dimension, please refer to Tab.\ \ref{tab:model_zoo}, where models are listed in the order of increasing QUBA score. 
Each model is implemented in PyTorch~\citep{Paszke:2019:PAI}; we use most models as they are, and do not change them in any way (except for training some of the self-supervised models to perform ImageNet-1k classification, \cf \cref{sec:details_selfsl}). 
The selected models were chosen based on several criteria. We only considered models that are publicly accessible and free to use for academic research. We aimed to include the most popular models and models achieving high ImageNet-1k accuracies.
We additionally included some particularly interesting models with designs that differ substantially from the more widely established models. 

\newpage
\onecolumn
{\tiny
\setlength{\LTcapwidth}{\textwidth} 
\setlength{\LTleft}{0pt}
\setlength{\LTright}{0pt} 
\setlength\extrarowheight{5pt}
\addtolength{\tabcolsep}{-1pt}
\begin{longtable}{@{}p{2cm}p{2.2cm}S[table-format=2.2]S[table-format=1.2]S[table-format=1.2]S[table-format=1.2]S[table-format=1.2]S[table-format=1.4]S[table-format=1.2]S[table-format=1.2]S[table-format=1.2]S[table-format=3.0]@{}}
\captionsetup{justification=justified,singlelinecheck=false,margin={0pt, 0pt}}

    \caption{\textit{Overview of our model zoo and the corresponding numerical results.} The configuration of each model lists the architecture, the training dataset, and the training paradigm. ``SL'' refers to supervised learning and ``AT'' to adversarial training. ``LP'' and ``E2E'' indicate if a self-supervised model has been adjusted for ImageNet-1k classification by only training a linear layer (linear probing, LP) or by fine-tuning the full network on the ImageNet-1k dataset (E2E).
Models are sorted according to their QUBA score in increasing order.} \\
    \toprule
     {\textbf{Model}} & {\textbf{Configuration}} & {\tablearrowcenter\makecell{\textbf{QUBA} \\ \textbf{Score}}$\uparrow$} & {\tablearrowcenter\textbf{Acc.}$\uparrow$} & {\tablearrowcenter\makecell{\textbf{Adv.} \\ \textbf{Rob.}}$\uparrow$} & {\tablearrowcenter\textbf{C-Rob.}$\uparrow$} & {\tablearrowcenter\makecell{\textbf{OOD} \\ \textbf{Rob.}}$\uparrow$} & {\tablearrowcenter\makecell{\textbf{Cal.} \\ \textbf{Error}}$\downarrow$} & {\tablearrowcenter\makecell{\textbf{Class} \\ \textbf{Balance}}$\uparrow$} & {\tablearrowcenter\makecell{\textbf{Obj.} \\ \textbf{Focus}}$\uparrow$} & {\tablearrowcenter\makecell{\textbf{Shape} \\ \textbf{Bias}}$\uparrow$} & {\hspace{-0.0em}\tablearrowcenter\makecell{\textbf{Params.} \\ \textbf{in Mil.}}$\downarrow$}\\
    \midrule 
    \endfirsthead

    \toprule
     {\textbf{Model}} & {\textbf{Configuration}} & {\tablearrowcenter\makecell{\textbf{QUBA} \\ \textbf{Score}}$\uparrow$} & {\tablearrowcenter\textbf{Acc.}$\uparrow$} & {\tablearrowcenter\makecell{\textbf{Adv.} \\ \textbf{Rob.}}$\uparrow$} & {\tablearrowcenter\textbf{C-Rob.}$\uparrow$} & {\tablearrowcenter\makecell{\textbf{OOD} \\ \textbf{Rob.}}$\uparrow$} & {\tablearrowcenter\makecell{\textbf{Cal.} \\ \textbf{Error}}$\downarrow$} & {\tablearrowcenter\makecell{\textbf{Class} \\ \textbf{Balance}}$\uparrow$} & {\tablearrowcenter\makecell{\textbf{Obj.} \\ \textbf{Focus}}$\uparrow$} & {\tablearrowcenter\makecell{\textbf{Shape} \\ \textbf{Bias}}$\uparrow$} & {\hspace{-0.0em}\tablearrowcenter\makecell{\textbf{Params.} \\ \textbf{in Mil.}}$\downarrow$}\\%\endhead
    \midrule \endhead
SigLIP2-l/16 \citep{Tschannen:2025:SMV}& ViL, WebLI\citep{Chen:2023:PAL}, self-SL (E2E) & -3.45 & 0.82 & 0.32 & 0.80 & 1.06 & 0.0389 & 0.91 & 0.98 & 0.69 & 882\\
BagNet9 \citep{Brendel:2019:ACB}& CNN, IN1k\citep{Russakovsky:2015:ILS} , SL & -3.01 & 0.46 & 0.00 & 0.15 & 0.25 & 0.0052 & 0.74 & 0.81 & 0.04 & 16\\
SigLIP-l/16 \citep{Zhai:2023:SLF}& ViL, WebLI\citep{Chen:2023:PAL}, self-SL & -2.82 & 0.80 & 0.23 & 0.72 & 1.04 & 0.0384 & 0.90 & 0.97 & 0.63 & 652\\
BagNet17 \citep{Brendel:2019:ACB}& CNN, IN1k\citep{Russakovsky:2015:ILS} , SL & -2.51 & 0.58 & 0.01 & 0.22 & 0.28 & 0.0042 & 0.72 & 0.79 & 0.03 & 16\\
CLIP-L14-Laion2B \citep{Schuhmann:2022:AOL,  Wortsman:2022:MSA}& ViL, Laion2B\citep{Schuhmann:2022:AOL}, self-SL (E2E) & -2.43 & 0.75 & 0.01 & 0.71 & 1.07 & 0.0370 & 0.89 & 0.95 & 0.54 & 428\\
CLIP-L14-DataCompXL \citep{Gadre:2023:DIS}& ViL, DataCompXL\citep{Gadre:2023:DIS}, self-SL (E2E) & -2.10 & 0.79 & 0.07 & 0.78 & 1.05 & 0.0380 & 0.90 & 0.96 & 0.61 & 428\\
AlexNet \citep{Krizhevsky:2012:INC}& CNN, IN1k\citep{Russakovsky:2015:ILS} , SL & -2.09 & 0.57 & 0.04 & 0.34 & 0.57 & 0.0017 & 0.73 & 0.80 & 0.26 & 61\\
CLIP-b/32 \citep{Radford:2021:LTV}& ViL, WIT400M\citep{Radford:2021:LTV}, self-SL & -2.06 & 0.63 & 0.05 & 0.64 & 1.12 & 0.0318 & 0.88 & 0.86 & 0.58 & 151\\
\end{longtable}
}

\twocolumn

\newpage

\clearpage

\end{document}